Brandenburg University of Technology
Faculty I – Mathematics, Computer
Science, Physics and Electrical Engineering
Chair of IT Security
Prof. Dr.-Ing. Andriy Panchenko

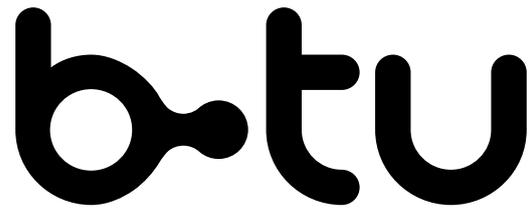

# Representation Learning for Content-Sensitive Anomaly Detection in Industrial Networks

## Fabian Malte Kopp

Master thesis submitted on the *13th of November 2021*

*Advisor*

|  | M.Sc Franka Schuster | Chair of IT Security |
| --- | --- | --- |

*Examination Board*

|  | Prof. Dr.-Ing. Andriy Panchenko | Chair of IT Security |
| --- | --- | --- |
|  | Prof. Dr. rer. nat. Oliver Hohlfeld | Chair of Computer Networks |

# Abstract

Using a *convGRU*-based autoencoder, this thesis proposes a framework to learn spatial-temporal aspects of raw network traffic in an unsupervised and protocol-agnostic manner. The learned representations are used to measure the effect on the results of a subsequent anomaly detection and are compared to the application without the extracted features. The evaluation showed, that the anomaly detection could not effectively be enhanced when applied on compressed traffic fragments for the context of network intrusion detection. Yet, the trained autoencoder successfully generates a compressed representation (code) of the network traffic, which hold spatial and temporal information. Based on the models residual loss, the autoencoder is also capable of detecting anomalies by itself. Lastly, an approach for a kind of model interpretability ($LRP$) was investigated in order to identify relevant areas within the raw input data, which is used to enrich alerts generated by an anomaly detection method.

*Danke für dein Treue Unterstützung Sarah.*

*In Erinnerung an Helga Kopp & Christel Momsen.*

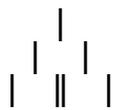

# Table of Contents

# 1 Introduction

Cybersecurity is an ever advancing field of research. One the one side, threat actors try to find vulnerabilities to exploit, while researchers want to develop concepts that are secure. The key to successful IT security is keeping up with the latest tools, strategies, and trends to constantly reconsidering and reevaluating the infrastructure and network. This cat-and-mouse scenario is a billion-dollar industry [1], and has seen rapid advancement in the last 20 years. Not only did more and more *critical* infrastructure, like industrial power plants, banking systems or other vital services providers utilize the cyber-physical domain [2], but hacking activities have also become more sophisticated. Nation-level agencies and independent groups around the globe are actively developing new and unknown attacks, so-called *zero-day* exploits. These exploits are even sold on darknet markets and target specific applications and platforms. Examples of attacks in which zero-day vulnerabilities were exploited are: *Stuxnet*, *WannaCry*, *Industroyer* and *Pegasus* [3].

The protection of critical infrastructure is mandatory by law in many countries. The German '*IT-Sicherheitsgesetz*' for example motivates the regulation with the following statement:

> "The IT Security Act is an expression of the state's responsibility to protect its citizens, the economy and its own institutions and administrations. [...] It also draws the conclusion from experience that a purely voluntary approach to establish IT security has not always led to the necessary commitment in industry and has not been effective across the board or in all security-relevant areas."[1]

Among the techniques, that defenders have to prevent harmful actions in adversarial settings, are *intrusion detection systems* (IDS). Based on their design, these systems can detect zero-day attacks.

To achieve this, the IDS has to learn a model of the underlying *normal* behaviour. Afterwards, a fully trained system can in theory detect any deviations from the established norm and thus alert anomalies. This training process needs sophisticated attributes (feature vectors) which represent the data. Manual extraction with specific domain knowledge or exhausted search methods for good performing feature combinations have been used in the past. This kind of active process is not feasible in ever-changing environments and is also prone to human bias. Advancements in the field of deep learning made it possible to automate the feature extraction process further.

This thesis investigates the use of unsupervised feature learning that is based on so-called *autoencoders*, to capture the essence of network data using different kinds of raw representations. The advantage of deep-learning-based methods will be evaluated during this thesis in the context of anomaly detection. Especially in the field of critical infrastructures like industrial networks, often proprietary protocols are deployed. Since the implementation specifications can change and little public documentation exists, automatic feature extraction is an essential tool to have.

---

[1]Das IT-Sicherheitsgesetz - Broschüre (translated)[4]

The learned representation of the network data retrieved by the autoencoder can then be used to detect anomalies via established one-class classification (OCC) methods like: one-class support vector machine (OCSVM), local outlier factor (LOF), and isolation forest (IF). All forms of intrusion detection do need some kind of numerical input representation. Using the proposed unsupervised framework, which reduces high-dimensional and content-specific network data to a condensed representation, that holds spatial and temporal information, can enable arbitrary OCC algorithms.

## 1.1 Problem Statement

The task of this thesis is to investigate existing methods of representation learning (feature learning) as well as different strategies for the processing of industrial network traffic in order to automatically derive meaningful and superior features for traditional machine learning methods. The aim is to achieve anomaly detection with higher model transparency and interpretability. In the context of the master thesis, the relevance of the new features and the advantages in interpretability and model transparency are to be investigated.
The proposed research concretely investigates:

   I) Are autoencoders able to perform for a spatial-temporal extraction of feature from raw data ?

  II) Can anomaly detection be improved when trained on extracted features ?

 III) What type of input data representations yields better results for different anomalies ?

  IV) How can the semantic gap be closed with the help of explainable models ?

## 1.2 Overview

The preliminary background chapter 2 will further describe concepts of network security, feature extraction, artificial neuronal networks, one-class classification and model interpretability. Afterwards, in chapter 3 the related work in the field of anomaly detection will be presented. For the evaluation of the proposed method, the given data sets will be presented in chapter 4. The *pcapAE* framework is presented in chapter 5, which enables a protocol-agnostic feature extraction of network traffic. Chapter 6 will evaluate the suggested framework against baseline measures for a number of different anomaly detection methods and answers the research question. The conclusion in chapter 7 will sum up the outcome of this thesis and lists recommendations.

# 2 Background

The following section gives an overview of the underlying concepts used in this thesis. After reading the chapter, one can understand what is meant by network anomalies and what different kinds of detection schemes exist as a prerequisite to follow the later argumentation. Also, foundational ideas about dimension reduction (DR) techniques and an introduction to artificial neural networks (ANN) are given. Lastly, necessary background information about one-class classification (OCC) and model interpretability are presented.

## 2.1 Industrial Control Systems

Industrial control systems (ICS) are specialized computer networks. These networks are the backbone of modern manufacturing or power plants. They are not only implemented to monitor the underlying process, but also are used to control physical actions performed inside an environment. Therefore, these systems can be described as so-called cyber-physical systems (CPS). An ICS can be operated from a remote distance over the Internet using a secure connection and thus needs to be protected from unauthorized access. The network topology of these systems range from low-level field devices that are used to operate valves and read values from sensors for monitoring purposes to control servers that issue commands to machines that are operated from control centers and lastly the corporate enterprise network the ICS can be accessed using a virtual private network access.

This rather fixed network environment produces overall predictable traffic patterns, which are very helpful for anomaly-based intrusion detection [5]. Typical patterns that appear are periodic readings from sensors, or cyclic issued commands that control the physical process. Since many low-level devices lack computational power, communication is often unencrypted within the network. Lastly, ICS networks often have a large throughput rate and also have branches of the network that are not based on the Ethernet protocol.

Figure 2.1[1] shows an abstract topology found in ICS networks. The network is typically divided into a number of levels that fulfill separate logical tasks and are secured using various information security concepts (see 2.2). The upper layers handles general tasks and host conventional IT hard- and software infrastructure, while the lower layers host operation technology (OT).

The corporate network handles administrative tasks (enterprise resource planning) and is connected to the industrial site using the Internet. The next level (manufacturing execution) hosts machines that can indirectly interact with the physical process. The plant level, is the only layer that is directly connected to the Internet, and is used to access the OT beneath it. These so-called human machine interfaces (HMI) are used to program specific actions that can be performed or to manually engage with the underling layers. The process control layer (level 2) also contains machines that are conditioned to gather historical data about the process. This data can later be used to analysis the performance.

These so-called distributed control system (DCS) systems are used to issue commands and, as described store metadata of e.g. sensor readings. Sensor readings and issued commands are can be transmitted using Ethernet-based communication. In some cases, low-level devices use serial protocol to exchange information. The level 1 network (instrument control) is connected to the DCS system. The third layer hosts specialized hardware components, the so-called programmable logic controller (PLC) which act as local control units.

---

[1]Illustration adapted from [6].

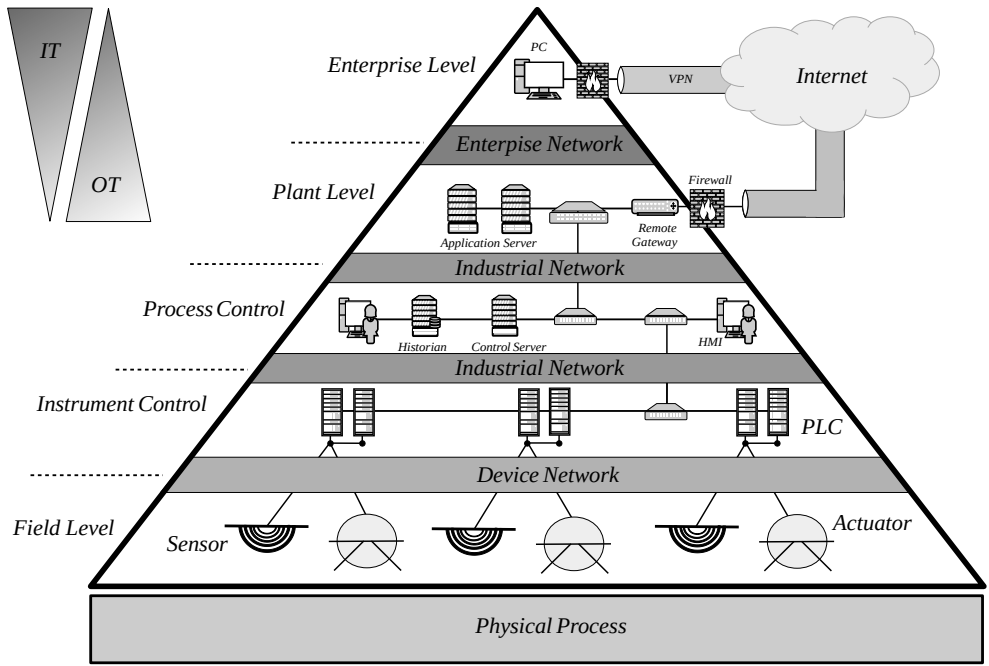

Figure 2.1: Overview of an ICS network hierarchy

These devices receive commands over e.g. Ethernet and translate them into analog signals that are used to initiate actions in the physical system. These programmable device are used interface with low-level hardware. They deliver the core functionality of a DCS system. In order to provide maximum availability, PLCs are often arranged in a redundant manner.

Lastly, hardware in the field layer (level 0) is used to interact with the physical process. Ultimately they form the foundation of the automated production cycle. Sensors are used for supervisory monitoring of real world conditions while programmable automated motor-control allow the actual process. This feedback-loop of sensors and actuators constantly audits its internal state and, if given, regulates itself in order to gain balance again.

### 2.1.1 Threat Model

A general survey about threats that OT-systems have to face is described in [7]. Known attacks against ICS networks are examined in [8]. As ICS networks use special hardware to realize the cyber-physical process, they offer a greater attack surface. A CPS has typically six components [9]:

1. The physical space ($PS$)

2. Raw sensor measurements picked up from the PS ($S_0$)

3. Measurements send to controller ($S_1$)

4. Communication between control unit and DCS System ($C_2$)

5. Issued control commands from controller (PLC) to the actuator ($A_1$)

6. Raw interaction between actuator and the physical process ($A_0$)

The communication between control unit (PLC) and actuators and sensors is denoted as the communication level zero ($C_0$). The content of this $C_0$-communication traffic often contains analog sensor measurement ($S_0$) and control commands ($A_0$).

$C_1$-communication can be threatened when the control unit is targeted by attacks, which can result in spoofed sensor readings and wrongly issued commands for example. The communication between the control unit and the DCS network is defined as $C_2$-communication. Figure 2.2 shows a CPS and its communication layers in detail. While conventional threat vectors like email phishing have to be regarded for the IT-layer in ICS networks, OT-specific threat vectors like firmware exploits or protocols attacks also have to be secured [10].

Attackers may take control over the control devices and issue their own commands without the control server (PLC), they may even try to spoof sensor readings ($C_1$-threat). The physical space and hardware can also be subject to malicious action. Although, this might presuppose local access to the floor level ($C_0$-threat). Lastly, attackers might take control over the ICS infrastructure. A compromised control over the complete automation pyramid, would allow attackers to issue commands directly or disrupt the PS by shutting down the operation ($C_2$-threat).

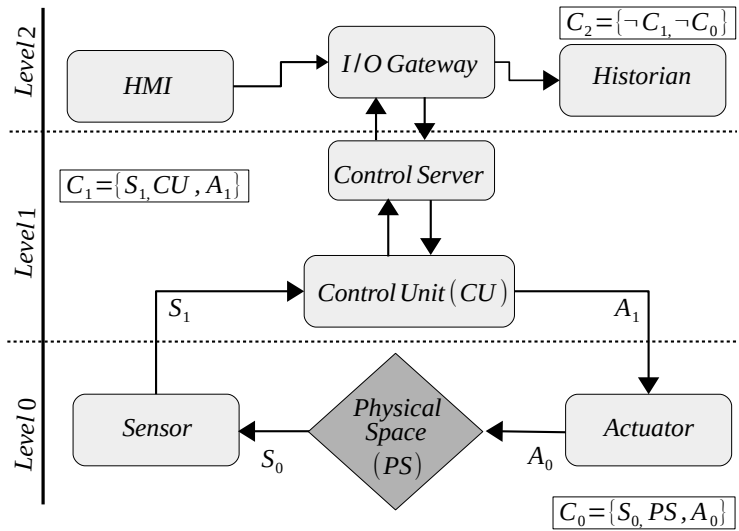

Figure 2.2: Communication layers of a cyber-phsical system

## 2.1.2 Network Traffic

The Open Systems Interconnection model (OSI model) in figure 2.3 shows that transmitted messages contains several independent layers. Network protocols are used to implement the specifications of a given layer. A specific layer can have several defined *header* fields and can contain a variable length *data* (*payload*). As illustrated, the payload data can itself be another protocol layer. The *physical layer* handles how *bits* are transmitted over a given medium. The *data link layer* then organizes these bits into chunks of data referred to as *frames*. The *network layer* (IP layer) handles how *packets* are distributed over connected networks. The *transport layer* is used to deliver *segments* (or *datagram*) of data to the right destination within the target system. The *session layer* provides the mechanism for opening, closing and managing a session between end-user application processes. The *presentation layer* ensures the information that the application layer of one system sends out is readable by the application layer of another system. Lastly, the *application layer* is the interface responsible for communicating *data* with host-based and user-facing applications.

Network traffic is usually captured in the PCAP (**p**acket **cap**ture) file format. The data is made from a collection of *frames* which, depending on the configuration of the network equipment, are fixed to a maximum length of bytes. Frames usually vary in size.

The maximum number of allowed bytes per packet is called maximum transport unit (MTU) and alters in different networks, the average number of bytes per packet in traffic capture is noted as the average transport unit (ATU). If a machine wants to share information that exceeds the *MTU*, the application payload is broken up into a sequence of packets (fragments). Individual packets then only contain a fraction of the complete message and are reassembled at their destination.

Data that is captured in a PCAP file has a chronological order in which the packets were registered by a given network interface. This perceived order is usually not as important as the content of the packets themselves because, depending on the used protocols, packets may be re-transmitted in any given order for the same request. In a PCAP file, it can be observed that packets appear multiple times, with just changing timestamps. This metadata is added by the packet capture itself, packets do not usually contain timestamps. In very homogeneous network environments the observed traffic can be highly regular and the received order of packets may hold more information about the state of the network, than in more open networks.

Lower network protocol layers of a packet contain more general information, while higher layers are specific to individual applications. For some (proprietary) protocols, intermediate layers can be skipped, and application data may be directly encapsulated within Ethernet frames.

Used network protocols are to some degree proprietary for which no public specifications are available. This makes them substantially harder to be inspected on a deep level. Patterns that emerge from network traffic are a product of its underlying network topology and the employed communication protocols. This behaviour is easy to see when data from closed industrial plants is compared to a public Wi-Fi hotspot for example. In the first case, highly regular patterns that follow predictable actions are observed, while open networks seem more chaotic and harder to predict over time.

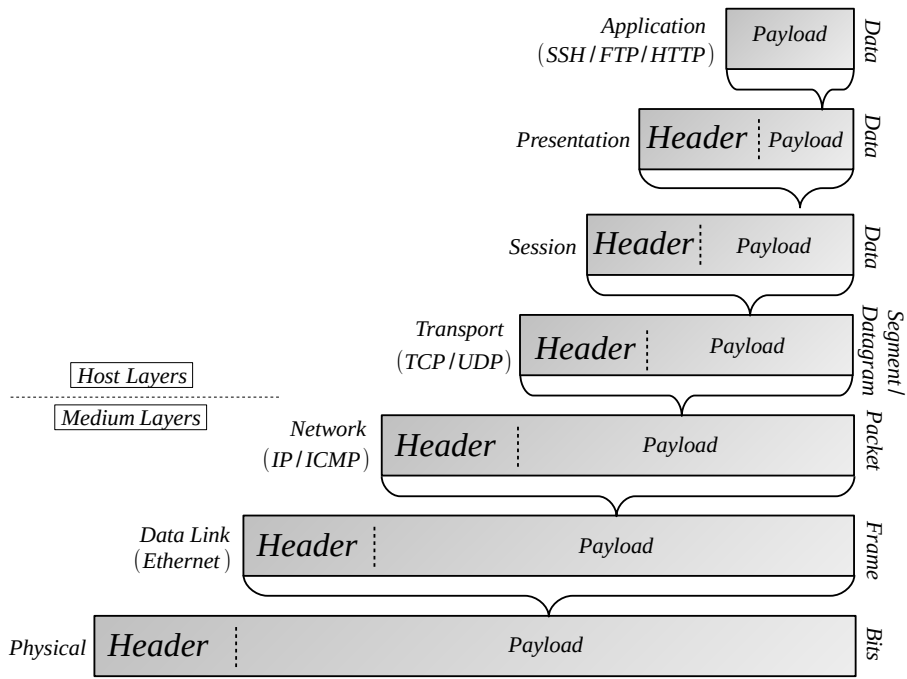

Figure 2.3: Principle of protocol encapsulation

**Feature Representation**

Besides the raw network captures, data from networks can also be further aggregated. This process often focuses on network *flows*. A flow describes a 5-tuple consisting of: source IP, source transport port, destination IP, destination transport port, and the used transport protocol. A flow-based distinction can give the stream of incoming frames a logical order and allows for a communication-oriented analysis. More sophisticated aggregation formats also extract other statistical elements from raw network traffic and use this information to represent network traffic characteristics over time. One advantage these types of data representation have is, that they can express features that are not given during the initial recording.

One can for example look at average bytes in a payload or the percentage of packets received by one address. Particular payload information may be lost in aggregated formats. This can make the detection of content-specific attacks like *server-side template injections* infeasible.

The domain of network security differentiates between methods that either conduct traffic analysis or deep packet inspection (DPI) [11]. Traffic analysis focuses on extracting higher-level aspects of packets and mostly only looks at header information and accumulated features, whereas packet inspection focuses on extracting (specific) features from the payload.

### 2.1.3 Network Anomalies

Anomalies in the context of computer networks are separated into two categories: active attacks or faults. Faults are caused by unmaintained hardware, wrongly configured hardware, or human error. Attacks or exploits on the hard- or software are active and targeted events. Communication protocols, for example, can potentially be abused and the system's architecture might be exploited for malicious intent [12]. After one system is compromised by some *hacker*, they often try to move through the network to infect more devices with malicious software (lateral movement). With modern protocol fuzzing techniques, attackers can also automate the search for protocol vulnerabilities. Fuzzing enables the finding of packet sequences that set devices in an exploitable state [13].

In [14] an ICS related overview about certain attacks and techniques is presented. Attacks on computer networks target the core principles in information security:

- *Confidentiality*

  Confidentiality describes the protection of information from unauthorized access and misuse. Attackers often scan a network to gather information about the environment. The probing of service information can lead to a break in the confidentiality.

- *Integrity*

  Integrity measures the protection of information from unauthorized alteration in any form. These measures provide assurance in the accuracy and completeness of data.

- *Availability*

  Availability is a key to a functional information system. This ensures that authorized users will always have access to the system, regardless of the time or circumstances.

To further explain the presented concepts, three attacks are will be highlighted in this section. As information systems evolve over time, so did attacks against them. The introduced principles help to classify certain attacks into categories. For sophisticated attacks, these categories often overlap as they target a variety of attack vectors which are combined to reach a certain goal.

In so-called Man-in-the-Middle (MitM) attacks, attackers break the confidentiality by re-routing communication between devices (hosts) such that the traffic passes through their control. Victims can potentially not notice the breach, as the attacker does not interrupt the communication. The unauthorized disclosure of information performed by the threat actor harms the confidentiality directly. Additional encryption schemes or *certificate pinning* can be used to protect against MitM attacks.

Related to attacks against confidentiality, are attacks that target the integrity of a system. If attackers can abuse some access control scheme and alter the content of a file, such as a service configuration, the integrity of said system is violated. With most cryptographic hash-functions one can easily validate that a given file was not altered, without actually parsing its content.

Lastly, on Denial of Service (DoS) attacks software components are overwhelmed with more information than they were designed to handle. This can have a negative impact on the system's availability and potentially deny real users access to the service. Network-based DoS attacks are best mitigated through a minimization of the attack surface.

Given the context of computer networks and the nature of traffic captures, the payload from a single application layer protocol may be broken up into multiple Ethernet frames, depending on the network's *MTU* (c.f. 2.1.2). With this regard to packet-chunking and reassembly and packet re-transmission, two notions arise: *intra-* and *inter*-packet anomalies [10].

1. *Intra*-packet anomalies occur when specific bytes within a single packet are anomalous. These *micro* patterns can be the result of carefully crafted payload that exploit some protocol or more general network scans.

2. *Inter*-packet anomalies are unwanted patterns of network traffic, where the context of examined packets before and after play a role to the attack vector. The *macroscopic* context of communication patterns such as time and order of frames is a crucial factor. MitM attacks are also a typical inter-packet anomaly as the communication is not affect directly, but is routed in an unusual manner.

## 2.2 Intrusion Detection Systems

intrusion detection system (IDS) are systems that are designed to protect IT infrastructures. By carefully examining and observing internal states or fingerprints, attacks or anomalies can be alerted using different concepts.

Network-based intrusion detection in particular is used to alarm network operators about attacks on their underlying computer environment. An IDS can be extended to also block communication, this is known as an intrusion prevention system (IPS).

An IDS can be implemented in two different settings. The first type focuses on individual machines and is called host-based IDS (HIDS). A host-based intrusion detection uses system logs, device messages and stack traces of observed machines and may raise an alert if unwanted patterns are seen in those collected messages.

The second context in which an IDS can be operated in is within a computer network. In this setting, the network-based IDS (NIDS) passively collects network packets that pass by the system and tries to determine if they are mischievous or not. This approach is able to monitor many devices at once.

One naive advantage network-based approaches have, is that almost all attacks against an environment are executed over the network, and attackers often use lateral movement to gain deeper access in their victims' systems. But this of course does not rule out host-based systems. However, if machines are attacked locally, log files and other sources are one way to verify the integrity of the machine. In practice, it is best to settle for a hybrid approach in order to cover all attack surfaces. Intrusion detection can help to ensure integrity and availability for the IT system, confidentiality can only be provided when most systems themselves block malicious actions.[2]

The fundamental basis any IDS uses to detect attacks for either described setting is split into two paradigms: signature detection and anomaly detection (AD). Table 2.1 shows the trade-offs between the two detection types.

|  | **Anomaly Detection** | **Signature Detection** |
|---|---|---|
| *PRO* | • Effective in detecting new attacks <br> • Low maintenance after training | • Actionable alerts <br> • Low false-positive rate |
| *CON* | • High false-positive rate <br> • Semantic gap <br> • Hyperparameters tuning | • Constant need to be updated <br> • Unable to detect unknown attacks |

Table 2.1: Intrusion Detection System Overview

### 2.2.1 Signature-based Detection

A signature-based intrusion detection uses a database of *fingerprints* or indicators of compromise (IOC) from before-seen attacks to compare new patterns against. These specific characteristics of observed exploits are a valuable hint for future attacks.

This concept can reliably and descriptively determine if malicious events occurred within the observed systems. The obvious drawback of this approach is that these signatures have to be collected beforehand and updated constantly to work effectively and provide protection.

---

[2]In real-time systems, the prevention of communication cannot be implemented since an IPS my yield false alarms.

### 2.2.2 Anomaly-based Detection

An intrusion detection based on anomaly detection follows the paradigm that a system tries to detect deviations from the before established norm. The norm is learned from available data that is assumed to be non-malicious. Anomalies are either attacks or faulty performance, section **??** describes some examples in detail. In order to detect certain attacks, anomaly-based intrusion detection often focuses on different aspects of the underlying data.

An anomaly-based IDS does not rely on up-to-date vulnerability databases and thus can in theory also detect zero-day attacks. In practices, they often suffer from a high false-alarm rate which limits their usability.

Detection systems work best when they can raise alerts in near real time. But in the case of anomaly-based methods, the data may be aggregated into a higher-level representation that contains information like specific protocol related rates of payload fields. This aggregation can limit its real-time applicability.

Three assumptions are usually made when anomaly-based intrusion detection is considered:

1. *Class Imbalance*

   Most of the observed data (network traffic) is normal [15]. Mostly, less then 1% of the data ever to be observed is abnormal. This results in an inherent imbalance between classes.

2. *Statistical Characteristics*

   Attacks have express characteristic that differentiates them from normal traffic which are used for the detection process [16].

3. *Training Data*

   The provided data to train the initial anomaly detection model does not contain anomalies.

#### Demands on Anomaly Detection

Anomaly-based intrusion detection should try to fulfill the following key aspects that especially arise for the domain of network security. The seminal work done by [17] expands further on the presented points. More pitfalls, like data collection and labelling, system design and problems within the evaluation are examined in [18]. They highlight challenges of e.g. inaccurate labels that researchers face when they conduct experiments in the field of anomaly-based detection systems.

1. *Usability*

   In order for an IDS to work effective, it has to have a manageable rate in which false alarms are generated. If the system throws to many (false) alarms, it cannot reasonably be used [19].
   The cost for misclassification is very high. Network operators have to investigate alerts generated by the system which can be very time intensive in order to determine if a real attack occurred within the network. Likewise, the cost for an unseen attack is very high, since integrity and availability are at stake.

2. *Explainability*

   In complex networks, it tremendously helps if the IDS can give a reason why a specific packet or sequence of packets is alerted. A score can also help to assess the event. In case of signature-based systems, raised alerts are easy to explain as the used IOC describes them. But in anomaly-based intrusion detection, a *semantic gap* [19] is created because systems often fail to indicate why they have chosen to mark parts of the traffic as anomalous. Explainable artificial intelligence is the key aspect for actionable alerts, since a system can tell its user what part and why the analysed data is abnormal.

3. *Robustness*

   The naive assumption is made that the provided training data does not contain any malicious or abnormal traffic, only normal patterns. This assumption may not always be justifiable, therefore, methods that are resistant against some form of training data contamination are sometimes favourable.

4. *Timing*

   Especially for the case of critical infrastructure, malicious events should be identified as fast as possible without any delay. The later an alarm is presented to experts the more damage can be done in the meantime.

5. *Complexity*

   The complexity of a used detection scheme should try to aim for a low computational complexity while ensuring high overall accuracy. This does not only lower the cost efficiency and $CO_2$-emissions, but is also better for given hardware resources on site.

6. *Scalability*

   A system is also arguably better if it can adapt to changes inside the network topology. This also has to apply to any IDS system, because the underlying hardware and software stack is stable but overtime may get extended. If the IDS is not able to adapt to those changes, it has no long-term value.

The proposed representation learning approach and therefore the AD based intrusion detection was developed and analysed in this thesis will be designed and evaluated with respect to its usability, explainability and complexity. Which are considered as crucial even for the initial design. Remaining aspects become only import in potential later and improved versions.

## 2.3 Feature Extraction

The extraction of relevant features from a high-dimensional data set is no trivial task [20]. Yet it is often a mandatory step in order to effectively work with the given data set, because the complete data set $D \in \mathbb{R}^{n \times m}$ may be to *big* to be processed further[3] or the raw data requires some form of *cleaning*. Cleaning a data set describes the act of removing of noise. Noise is subjective to the type of data. A single feature $m_k \in \mathbb{R}^n$ is a one-dimensional column, whereas a data point (or sample) $n_j \in \mathbb{R}^m$ is represented by a row or $m$-dimensional feature vector of the given matrix $D$. The *complexity* of a data set can be viewed as the number of features $|D| = m$. The total number of rows $n$ is usually much larger than the complexity.

Besides the computational performance improvement a reduced and clean feature space can offer to a classification algorithm, a study has shown that a carefully chosen feature space can also increase the accuracy of the class predictor. It also helps to prevent the model to *over-fit* to training samples. Over-fitting occurs when a model does not have the ability to generalize, which results in poorly predictions for new examples [21].

The careful removal of irrelevant or unstructured information in data can intuitively help any data-driven algorithm to make better decisions, because it has less uncertainty in its input signals.

Data sets can be obtained in various forms such as structured data points that have a clear chronological order, multivariate time series or mostly independent data points with no obvious correlation. Machine learning also differentiates between labelled data sets or *supervised* problems, where each data point has a tag that encodes some information about the given sample. This label can then be used in a supervised learning scheme to train a given algorithm. Unlabelled data sets are on the other hand so-called *unsupervised* problems, where a learning algorithm has no extra information to train a model on.

Feature extraction plainly describes the act where input vectors from a data set are transformed into feature vectors that are the basis for a classification or regression task. The resulting vectors are a product of the source data and the given extraction process. Especially for non deep-learning-based algorithms, the provided feature vectors are essential for the performance of the trained model. Deep learning-based models have shown capabilities to extract meaningful features in a supervised manner, and thus do not need handcrafted extraction to achieve state-of-the-art results [22].

The concept of dimension reduction (DR) is related to feature extraction. Given some data, a DR method tries to find a representation of the data with less complexity. This reduction is not arbitrary and theoretically should contain roughly the same amount of information. Of course, it is impossible to reduce the number of dimensions while keeping the exact same information, thus DR always introduces some form of information loss. This inherent loss of information and expressiveness of the data may affect the explainability of the transformed data set. Logically one can only expect a fraction of the patterns that were present in the original data set.

DR methods can transform the data in a linear and non-linear manner, depending on the underlying algorithm. The values of a given data set are embedded in some manifold. This has strong implications for the result of the applied DR method. The ability for methods to extract relevant patterns from unseen data points that were not given during the initial phase is an advantageous feature. These methods are described as being able to process out-of-sample (OOS) inputs. Systems in this sense have an internal knowledge about the provided data and are able to also extract relevant feature vectors.

---

[3]Curse of dimensionality - high-dimensional data is exponentially harder to process.

### 2.3.1 Feature Generation

When unstructured data $U$ is provided, one first need to define or generate some kind of feature space such that further algorithms can digest them. When working with textual data for example, this could be the use of *n-grams* techniques over the text corpus, which creates a set of fixed-length character sequences. Network data in the form of PCAPs can be used to generate a variety of statistical features, or the individual packets can be parsed to extract information from header and payload fields.

$$\nu \colon \mathbb{R}^{\omega} \to \mathbb{R}^{n \times m}, \ \nu(U) = D$$

### 2.3.2 Feature Selection

The process of selecting a smaller subset of the feature space is called feature selection. A naive approach would consist of doing an exhausted search, where a subset of features is chosen and then evaluated against the performance metrics of a classification algorithm and then compared to all other possible subsets of features. This will result in an optimal set of features for the given training data, but might fail to generalize which introduces *over-fitting*. Let $M = \{m_i \in \mathbb{R}^n\}$ be the set of features in $D$ and $\mathbb{P}(M)$ be the power set of $M$. Selecting only certain features $s \in \mathbb{P}(M)$ is noted with $D_s$, where $s \subseteq M$ and $|D_s| << |D|$. Another possible strategy is to find out which features contribute the most to the final result, and only then to select the top $x$ percent of features. Features are therefore more relevant when they have a larger influence on the downstream task.

### 2.3.3 Feature Construction

Feature construction describes a facet of feature extraction which can also reduce the number of dimensions when applied. The idea is that one creates or generates a new space of features from the original one. A simple example of this kind of approach is presented with the function $\tau$. A set of functions can be used to concatenate different features into new ones or to numerically combine features in a *meaningful* combined representation.

One can also generate new features on a row-level basis, where features from different consecutive rows are combined to create a new feature for the data set. The transformation achieved with this function $\phi$, could for example be a sliding window of size 2, which calculates the difference in a feature to create a new attribute that indicates the amount of change that occurred between the two samples.

$$\tau \colon \mathbb{R}^{n \times m} \to \mathbb{R}^{n \times (m-k)} \qquad\qquad \phi \colon \mathbb{R}^{n \times m} \to \mathbb{R}^{n \times (m+k)}$$
$$g, h \colon \mathbb{R}^{m^+} \to \mathbb{R}^m \qquad\qquad t, z \colon \mathbb{N} \times \mathbb{R}^m \to \mathbb{R}^m$$

$$\tau(D) = \begin{cases} \dots \\ g(m_i) = \hat{m}_j \\ h(m_j, m_k) = \hat{m}_{jk} \\ \dots \end{cases} = D_\tau \qquad\qquad \phi(D) = \begin{cases} \dots \\ t(ws_1, m_i) = m_{\delta i} \\ z(ws_2, m_j) = m_{\delta j} \\ \dots \end{cases} = D_\phi$$

### 2.3.4 Feature Engineering

One can also consult domain experts of a given problem space and rely on their knowledge to decide which features are meaningful and should be used to represent high-dimensional data sets. This is often a labour-intensive and static task and thus hard to accomplish on a large scale. Manual engineering of features was utilized in the past with great success. One potential drawback is the bias experts introduce to the extraction process. In fields like network security this is also not applicable since the infrastructure and its protocols change over time, which would need continuous attention and is not future-proof.

### 2.3.5 Representation Learning

The idea behind feature learning or representation learning (RL) is that an automated system derives features from a given data set. The derived set of features can then be used for arbitrary downstream tasks like classification, regression or visualization [23]. It is typical for representation learning to also reduce a data set's complexity.

A suitable function $\lambda$ should reduce the amount of features in a data set while still keeping the intrinsic characteristics of the original data. The compressed data set, just like other dimension reduction methods, can only express a fraction of its original patterns. Therefore, some information about the data may get lost in the process. RL is thus a mostly unsupervised combination of dimension reduction and feature extraction [23].

One possible strategy is to define some measurement function $H$ which can be used to compare different possible transformed data sets.

$$\underset{\lambda \in \Lambda}{\mathrm{argmax}}\{H(\lambda(D))\}$$

where

$$\lambda \colon \mathbb{R}^{n \times m} \to \mathbb{R}^{n \times w}, \ w < m, \ H \colon \mathbb{R}^{n \times w} \to \mathbb{R}$$

Since there are countable infinite ways to choose a function $\lambda$, the metrics used to evaluate the transformed data set are very important in order to determine how *good* the model is. In [24], the Author compares dimension reduction measures and evaluate them using different data sources. He also proposed a new metric which is suitable for a variety of data sets.

A second method one can use to evaluate a DR technique is to apply downstream task on the reduced representation. The task can for example be a classification of the data using an already established model which is trained on the new data. This has the benefit that the evaluation can use metrics of the classification task.

### 2.3.6 Feature Transformation

Closely related to feature extraction is the transformation of identified attributes. Many algorithms that make data-driven decisions converge substantial faster when the provided features are optional: normalized, standardized and also scaled. For all presented preprocessing types of transformations, many heuristics can be picked. While these transformations help to unify a given feature, it is also possible that the transformation skews the underlying geometric manifold.

A transformation is the application of a mathematical function. Every point in a data set $x_i$ is replaced by $f(x_i)$. The function that is used to transform the data is usually invertible, and generally is continuous.

- *Normalization*

  Normalization of a feature vector $n_i \in \mathbb{R}^n$ ensures that the numeric values of a given feature $i$ are bound to an fixed interval of e.g. $[0, 1]$.

$$n_{norm} = \frac{n_i - min(n_i)}{max(n_i) - min(n_i)} \in \mathbb{R}^n$$

- *Standardization*

  A feature standardization transforms a data set such that each feature $m_k \in \mathbb{R}^m$ has a unit *mean* and a *standard deviation $\sigma$* of 1 for example.

$$m_{stand} = \frac{m - \bar{m}}{\sigma} \in \mathbb{R}^n$$

  where

$$\bar{m} = mean(m), \ \sigma = \sqrt{\frac{1}{N-1} \sum_{i=1}^{k} (m_i - \bar{m})^2}$$

- *Scaling*

  When a dimension in a data set is scaled, individual feature vectors $n_i \in \mathbb{R}^n$ are resized to a common length.

$$n_{scaled} = \frac{n_i}{|n_i|} \in \mathbb{R}^m$$

  and where $|n_i|$ defines a measure of *length*.

## 2.4 Artificial Neural Networks

An artificial neural network (ANN), as the name suggests, is vaguely inspired by the biological network of neurons in brains. ANNs were initially designed to simulate associative memory [25].[4] Like biological neurons, artificial neurons [26] also get excited by some input signal, and depending on the non-linear *activation function*, propagate the signal in the network.

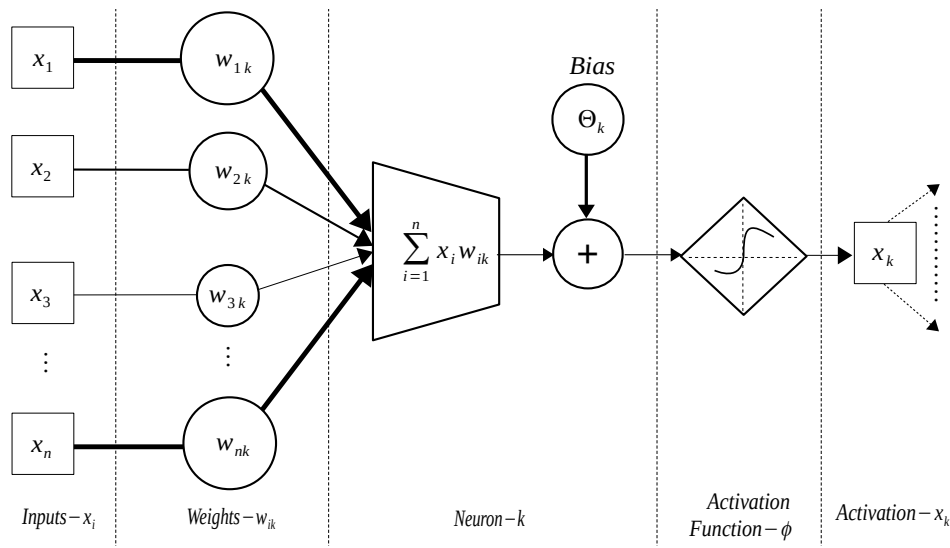

Figure 2.4: Schematic overview of a linear artificial neuron

Each neuron as depicted in figure 2.4 has a number of input signals $X_i$ and an output signal which is connected to a number of other neurons in the network. The output $X_k$ of a single neuron $k$ is computed as the weighted sum of all input connections, where the individual weights to each incoming connection changes during the network's training process. A neuron can also add an internal bias term $\theta_k$ to the sum, which is also adjusted during the training loop. The neuron's output value is finally dictated by its non-linear activation function $\phi \colon \mathbb{R} \to \mathbb{R}$. A popular choice for a non-linear function is the rectified linear unit (ReLU) [27] which is defined as $\phi(x) = max(0, x)$.[5] In the simplest form, each neuron $x_k$ thus computes the following value:

$$\phi((\sum_{i=0}^{n} x_i w_{ik}) + \theta_k) = X_k$$

A collection of neurons can be configured in any shape and form, but is usually arranged in forms of layers of neurons. The first batch of neurons is called the *input layer*, and the last layer of neurons is called the *output layer*, all neurons in between these are called *hidden units*. For every dimension of the input data, the input layer has a corresponding neuron. Networks are so-called *deep* when they consist of many hidden layers. Neurons of one layer are often only connect to other neurons in following or preceding layers. In a recurrent architecture, neurons are arranged in an interconnected manner.

---

[4] Memory that learns and remembers relationships between 'unrelated' items.
[5] Using non-linear activation functions enables an ANN to learn non-linear functions.

The simplest type of architecture is a multi-layer perceptron (MLP). The MLP is a network consisting of just one hidden layer and a fully connected neuron arrangement.

A deep neural network is theoretically a universal function approximator. This means that it can, with enough hidden layers, sufficient data and adequate training time, model any arbitrary function [28]. Specific finite network architectures have the power of universal *Turing machines* [29] when rational weight values are used.

### 2.4.1 Training in Neural Networks

Neural network-based computing systems *learn* on an example basis, where the internal knowledge of the network is represented by the ensemble of weights and biases from the collection of neurons. When training an ANN, it is important that the network does not simply memorize the provided data, this phenomenon is called *over-fitting* in machine learning. The goal is to adjust all weights of the network, such that when the training is finished, the model is also capable to meaningfully process data that it has never seen before.

The learning process in ANN is quite similar among the neural network types. The main difference is often the used *loss function*[6] and the strategy in which the weights connecting the layers of neuron are updated.[7]

The loss function $E$ measures how *wrong* the prediction the network made was and thus is a direct criterion to evaluate the model. A typical example for a criterion function is the mean square error (MSE). This error is a function of the input $X \in \mathbb{R}$ and the quadratic difference between to the prediction $\hat{X} \in \mathbb{R}$ summed for each dimension. This punishes mistakes (high error term), but also rewards the network for its correct predictions.

$$MSE(X, \hat{X}) = \frac{1}{n} \sum_{i=1}^{n} (x_i - \hat{x_i})^2$$

where $n$ is the number of dimensions of a given sample. The computed loss term is propagated back through the network to adjust each weight $w_{ij}$ individually in such a way, that the prediction would be *more* correct. This process is called error *backpropagation* [30] and is a fundamental concept in deep learning. The error amount is divided among the neurons' connections, by computing the gradient (partial derivative) of the loss function with respect to a given weight. As the name backpropagation suggests, this process is applied from back to front of the network. First, the output layer's partial gradients are calculated and used to notch their weights slightly in the direction of the steepest decent. This iterative process is called *gradient descent* [31] and is applied for each input example one at the time.

Figure 2.5[8] visualizes this *loss landscape* for a minimal two-dimensional example. The error function describes some unknown manifold, where weights are pushed in the direction with the steepest decent. Weights tend to become stuck in local valleys if the gradients are very small. This is known as the vanishing gradient problem [33]. The loss within an ANN obviously decreases over time, since the model learns patterns of the provided data. A decreased loss thus indicates that the model is learning. This fact makes the training process an optimization problem, were one tries to adjust weights in order to minimize the loss. When more than one sample is presented to the network before backpropagation is applied, the network can accumulate the error term over all examples of the given *batch*. Batch-oriented learning brings faster and more stable descent to a local minimum on average. Since each update is performed in the direction of the average error term of a batch, this can result in a poor generalization of the model.

---

[6]Sometimes referred as a *criterion* or *cost function*.
[7]When looking at a network architecture with fixed number of units and activation functions.
[8]Illustration adapted from [32].

A popular framework for a gradient descent strategy is *ADAM* [34]. The authors propose a method that offers an intuitive hyper-parameter space with little parameter tuning. Many flavours of ADAM were introduced in the past, a weighted variant of ADAM [35] showed faster convergence on different data sets and will later be used for the training process.

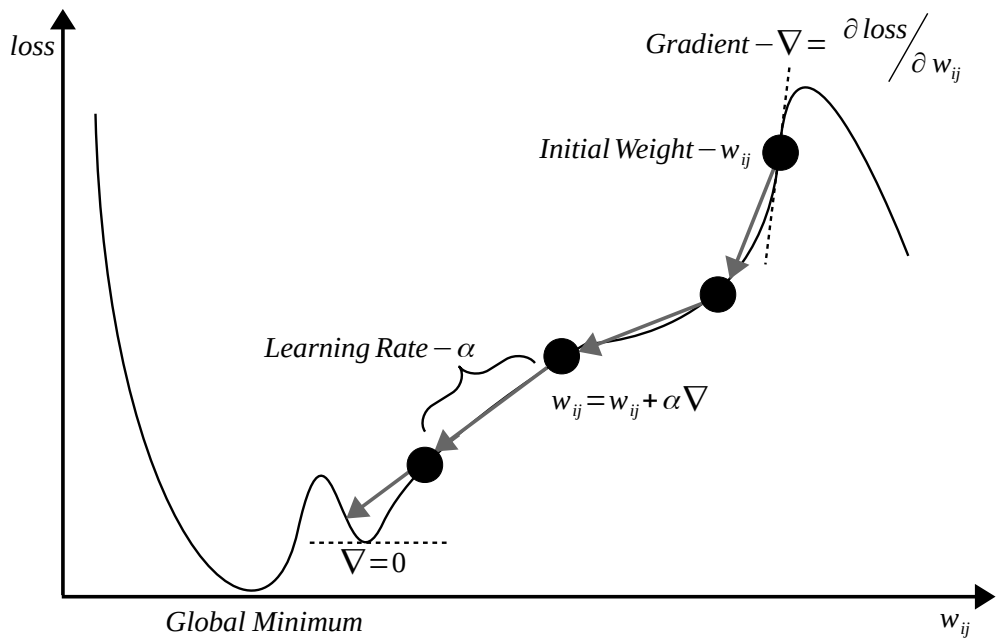

Figure 2.5: Example of gradient descent in two dimensions

Analog to Hebbian learning [36], a learning rate (LR) $\alpha$ dictates the distance or amount a back-propagation step moves the weights in a direction. Choosing a large LR will generally result in fast convergences in a local optimum, while small rates can find a better optimum but can take a more time. Per iteration of backpropagation the rate any given weight changes is fixed by $\Delta w_{ij} = -\alpha \frac{\partial loss}{\partial w_{ij}}$.

One strategy that is deployed in ANN is to adjust the learning rate dynamically. This can help the model to jump out of plateaus and sometimes allows reaching better local optima. In [37] the authors investigate a novel strategy to update the learning rate on a cyclic manner.

Hyperparameters in the context of ANN are: the number of hidden layers ($\#h$), the number of neurons per layer ($\{\#h_i | i \in [0, \#h]\}$), the numbers of examples before the weights get updated (*batch size*), the number of times the learning algorithm will process the all batches (*epochs*), the rate weights are adjusted in each backpropagation pass ($\alpha$) and the used loss function $E$ itself.

The weights of an ANN are more or less initialized randomly, and depending on this starting configuration may take more or less time to find a local optimum.[9] When dealing with deep modular networks, one can also try to pre-train individual sub-networks first, to push their initial weights in a favourable position.

After this pre-training phase, the network is trained as a whole, to eventually find a suitable local optimum. This process can converge significantly faster than when the complete network is trained at once from start.

As stated above, the training of ANNs should lead to generalization, not to memorization of the training data. It is more favorable if the resulting model has learned a general representation of the data. This allows the model to process unseen data correctly.

---

[9]Different weight initialization can yield in different optima.

### 2.4.2 Neurons

The following incomplete list of neurons shall give a high-level overview over some prominently used layers that can be found in many types of ANN. The deployed neurons dictate the net's functionality. Besides the basic *linear* neurons which only calculate the weighted sum as explained above (c.f. section 2.4), one can also use neurons which use more complex transformations. These *advanced* neurons may be more useful at learning complicated patterns. For spatial feature extraction often convolution architectures are used, while sequential data is prominently handled with recurrent layers.

Besides these layers, there are also auxiliary layers like: *GroupNorm* and *Dropout*. GroupNorm layers ensure that numerical values which are propagated throughout the network do not *explode* to infinity and help the network to converge faster. By normalizing the output values of neurons, layers deeper inside the network can work with predictable values. One common strategy against the problem of over-fitting is called *Dropout* [38]. This is used to force an ANN to generalize to the data set by randomly disable neurons during the training phase. Since the network of neurons has uncertainty if a given neuron is activated or not, each neuron automatically learns to encode information about a variety of samples. The proposed framework in chapter 5 will be built using various layers and configurations of the above presented neurons (c.f. appendix C.1).

#### Convolutional Neurons

Calculating the convolution of two functions $f$ and $g$ is a mathematical process that produces a third function ($c = f \star g$). For example, if one function is convoluted by another, the result will be a third function that expresses how the shape of function $g$ will be change based on function $f$.
This idea is used in convolutional neural networks (CNN) to learn so-called filters (kernels or feature-maps) that activate when a certain feature is detected at a specific spatial position in the input. Traditional algorithms rely on engineers who design and build these filters. This is a time-consuming and expensive process. In CNN networks, however, the filters are learned through automated learning. This means that it is less costly to create these filters because they can be created without knowledge of the features beforehand.
A CNN consists of many filters, which have a small receptive field (input map) but can be applied to larger multi-dimensional input matrices (tensors). During the forward-pass, each filter is applied to the input and creates a feature map. The hierarchical network architecture uses the convoluted output (feature maps) and repeats the process in a hierarchical manner. This results in highly specific filters that get excited when, a specific complex pattern is presented to the input layer.
In a CNN, the input is a tensor with a shape: *samples* x *height* x *width* x *channels*. After passing through a convolutional layer, the tensor becomes abstracted into a feature map with shape: *samples* x *feature-map height* x *feature-map width* x *feature-map channels*.
Figure 2.6[10] shows how a feature-map is generated in a simple two-dimensional example. By moving the kernel over the input matrix, the feature map is build up in four steps. The convolution uses an adjustable kernel, represented with dotted areas, which is subject to backpropagation within the CNN. The produced feature map is the input signal for following hidden layers.

#### Recurrent Neurons

A recurrent neural network (RNN) is a type of ANN that is capable of processing variable-length sequences of inputs. RNNs were first introduced in 1982 [40]. Through their recurrent design, they are capable of learning sequential dependencies with in the provided input data. This has the implication that a neuron's output at step $t$ is used at step $t + 1$ and feeds back into the same neuron, along with the next sample of the sequence. In addition, these kinds of neuron's can use their internal state (gate) to process inputs.

---

[10]Illustration adapted from [39].

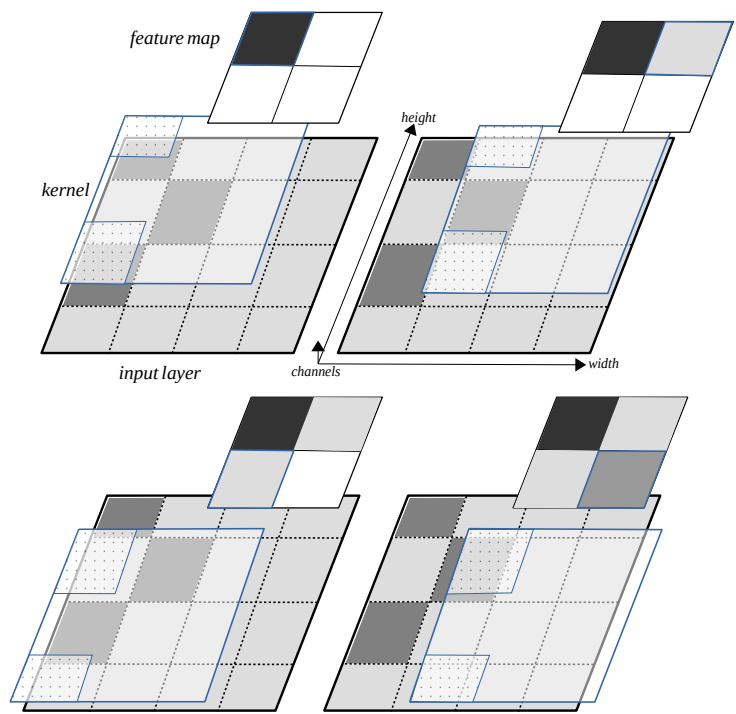

Figure 2.6: Convolution of an input layer using a kernel resulting in a feature-map

In fully recurrent neural networks all outputs are connected to all inputs. This is the most general network topology because anything can be simulated by disabling certain connections. A recurrent neural network can have one or more additional stored states (gates). The storage can be controlled by the neuron itself, and it can also be replaced by another neural network. A gate may include an additional feedback loop or time delays.

A gated recurrent unit (GRU) are one type of RNN and were introduced in 2014 [41]. Figure 2.7[11] shows a schematic overview of a GRU cell. One can see that two input signals are fed into the unit: $x_t$, the current step, and also the last hidden state $h_{t-1}$. The cell outputs a single value ($\hat{x_t}$) which is equal to $h_t$. GRU cells consist of two gates, where each gate can be considered as an individual neuron that is trained during the training loop. First, the reset gate determines how big the impact of the last time step $h_{t-1}$ will be for the current input $x_t$. The internal update gate $r_t$ (reset gate vector) learns how much the current signal will influence the last hidden state. Lastly, the next hidden state ($\hat{h_t}$) will also be multiplied by the current update value $z_t$ (update gate vector) and added to the overall transformed hidden state. Formally the GRU cell can be written as:

$$r_t = \sigma_r(W_r[h_{t-1}; x_t] + b_r)$$
$$z_t = \sigma_z(W_z[h_{t-1}; x_t] + b_z)$$
$$\hat{h}_t = \phi_h(W_h[x_t; r \otimes h_{t-1}] + b_h)$$
$$h_t = (1 - z_t) \otimes h_{t-1} \oplus z_t \otimes \hat{h}_t$$

where the collection of weights $W_r, W_z, W_h$ for each gate and their biases $b_r, b_z, b_h$ are subject of backpropagation and are adjusted over time. Brackets denote a vector concatenation. $\sigma$ is a sigmoidal activation function whereas $\phi$ is a hyperbolic activation function.

---

[11]Illustration adapted from [42].

The core concept of gated RNN is that the cell state, in theory, can carry relevant information throughout the processing of the sequence. Even information from earlier time steps can make their way into later time steps (long short-term memory). The gates can learn what information is relevant to keep or forget during training.

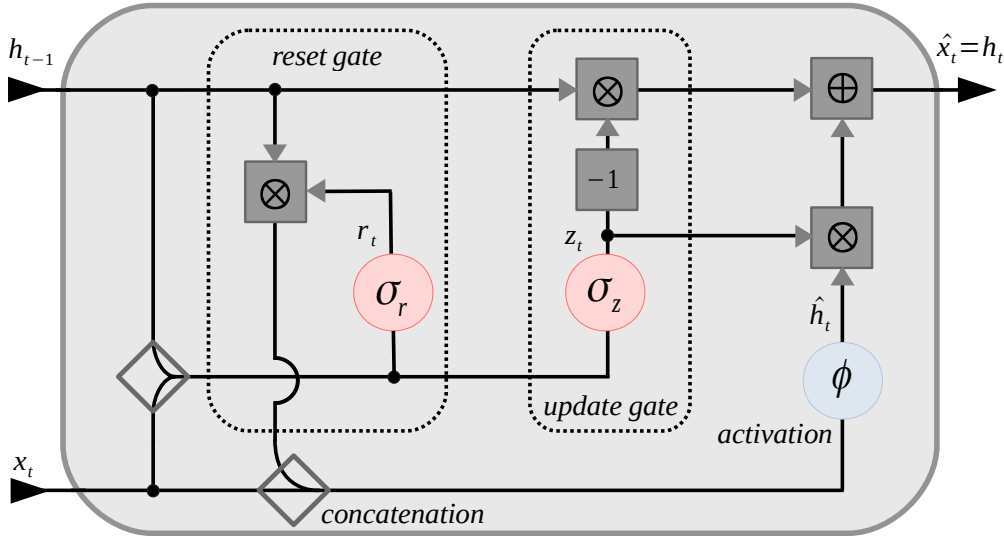

Figure 2.7: Schematic overview of a gated recurrent unit

### 2.4.3 Autoencoder

The autoencoder (AE) is a type of deep neural network which was first published in 1986 [43] and is utilized in various kinds of tasks like: image in-painting, super image resolution, object detection, music generation, and more [44].

An autoencoder is a deep architecture for *unsupervised* learning. As described in 2.3.5, this method can be used to generalize a data set. The keyword *unsupervised* implicates that the method does not rely on some form of labels for the training process. While training, the model will treat the input samples as some form of label itself. This makes the concept *self-supervised*.

An AE has the *trivial* task to learn the identity function $ID(x) = x$. To restrict the network, an information bottleneck is used to handicap the net's ability to just memorize the training data set, and to force the autoencoder to generalize the training data. The information bottleneck is simply created by limiting the number of neurons per hidden layer. The encoder network thus only has a diminishing number of weights available to represent some knowledge. The decoder network on the opposite side contains more and more neurons per hidden layer that help to recreate the original sample. A schematic depiction is shown in figure 2.8.

The network consists of two sub-networks (funnels), the *encoder* network ($e\colon \mathbb{R}^m \to \mathbb{R}^d, d << m$) which creates a compressed representation. The compressed representation generated by the encoder is referred to as the *code* or latent variable $l = e(W_c \times x + b_c) \in \mathbb{R}^d$. Using the latent variable $l$ the *decoder* network ($d\colon \mathbb{R}^d \to \mathbb{R}^m$) is conditioned to recreate the original data from a produced code. This compression performed by autoencoders is not lossless and the decoder network may introduce some artefacts in the reconstruction $d(W_o \times l + b_o) = \hat{X} \approx X$. The weight and bias parameters $W_c, b_c$ and $W_o, b_o$ are changed during the training process of the neural network (see section 2.4.1).

This concept used in AEs exhibits key aspects that are found in feature extraction (c.f. section 2.3.5). By using a form of function approximation, it is possible to model any provided data set to a given degree.

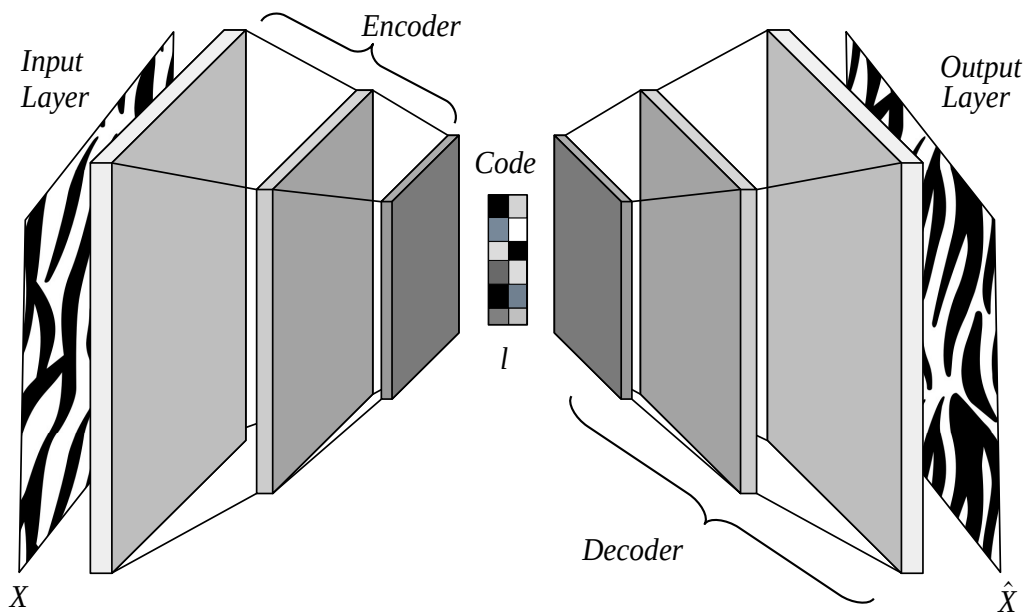

Figure 2.8: Autoencoder composition

The derived code $l$ from a fully trained AE of a given sample $X$ is a compressed representation and can be used to represent the sample itself, using only a fraction of its complexity.

If a steep bottleneck is chosen, the network has fewer neurons to encode a model of the provided data set. This results in larger reconstruction errors. On the other hand, a flat bottleneck will produce fewer artefacts in its reconstruction but often fails to generalize to the data.

The hidden layers in each sub-network can also be made of smaller en- or decoders, this configuration is called a *stacked* autoencoder and provides a smother bottleneck.

Authors in [45] proof that a single-layer autoencoder is related to linear dimension reduction via principal component analysis. When more than one hidden layer is used, the network is also able to perform non-linear transformations.

Some used loss functions can enable an AE to learn to model an underlying distribution of the data set. This flavour of autoencoder is called *variational* [46], it is often used for generative purposes. When artificial noise is added to the input data before it is passed into the autoencoder, one refers to a *de-noising* autoencoder. This type of system has the goal to eliminate the introduced noise in samples, which helps to make it robust for out-of-sample inputs. The autoencoder is trained to remove the noise from the *corrupted* data by using the original input data as a comparison for the optimal target.

Since AEs (encoder networks) have proven to offer a powerful for learning an efficient representation of data [23], this thesis will be evaluated together with other dimension reduction methods, such as principal component analysiss (PCAs) regarding its suitability for the given problem of network traffic feature extraction.

# 2.5 One-Class Classification

In *supervised* tasks, there exist samples from all possible classes and an underlying model can use the class labels to build a model to distinguish between them. The distinct characteristic of a one-class classification (OCC) problem is that one has only one class or category of data available to train a model on. This fact makes OCC an *unsupervised* problem.

Section 2.2.2 describes the problem of anomaly detection and one can see that it is de facto an OCC problem. Given a set of assumed normal data to train on, an AD model is intended to afterwards identify anomalous data it has never seen before. Within normal data some points may occur less frequent than others or can deviate from the mean of the distribution of data points. These events are called *outliers* and are equally important as the average data point when modelling the patterns of the normal data. Besides for intrusion detection, OCC is also applied in the field of fraud detection and others [47].

A trivial extension of OCC is that a model also tries to classify or cluster found anomalies further into specific bins that are relevant to a given knowledge domain. This process can help to increase the explainability of the underlying model.

The Difference between approaches is the mathematical foundation they use for their detection mechanism. Ranging from distance- or cluster-density-based methods, to information theoretic approaches, a summary in [48] surveyed popular frameworks on different data sets.

The selection of features provided for an algorithm are highly relevant for its performance and need to be chosen with great care. In mixed data sources, textual or categorical features are also subject to some degree of feature extraction (c.f. section 2.3) to provide a uniform data set. The type of feature representation also changes the outcome of the model, this highlights the importance of the preprocessing of the used input features.

The following section focuses on three established and non deep-learning-based methods (shallow learning) for OCC, and describes their methodology. The evaluation in chapter 6 will use these algorithms to benchmark the proposed framework.

## 2.5.1 One-Class Support Vector Machine

The support vector machine (SVM) was introduced by [49] as a tool for supervised classification. Utilizing kernel-based transformations, which perform projections into higher dimensional spaces, an SVM is able to separate classes of data in a non-linear fashion. Schölkopf et al. [50] extended the idea to also solve unsupervised problems and called it one-class support vector machine (OCSVM). This classificator learns on examples of a particular class (e.g. normal instances) and later on is able to identify outliers (samples that are not confirmed with the learned data). The OCSVM achieves this by constructing a hypersphere in a transformed feature space. The spherical separation in a high-dimensional representation translates to a non-linear separation of samples in the original feature space. To map samples $x \in X$ into dimensional higher representations, a number of different *kernels* $K : X \times X \to \mathbb{R}$ can be chosen. For each pair $x_i, x_j \in X$ the inner product $\langle \phi(x_i), \phi(x_j) \rangle$ is calculated where $\phi : X \to F$ defines some arbitrary *feature-map*.

An illustration of the concept is shown in figure 2.9.[12] In the simple example, points are mapped to a higher dimension. A typical choice for a kernel is the *radial basis function* since it preserves local properties of the input space in higher dimensions. Other kernels are: polynomial-, sigmoidal- and Gaussian-based functions.

After the OCSVM algorithm has projected data points into a sufficiently higher-dimensional space, all samples can be enclosed by the smallest hypersphere containing most of the points. For Euclidean distances this is equivalent to establish a hyperplane with a maximum margin from the origin [50]. The hypersphere acts as a decision boundary to classify points into binary categories. Normal points that are tangent to the decision boundary are called *essential support vectors.* They are prototypical examples of the learned behaviour and are used to classify new examples. The support vectors are later used to construct the decision boundary.

The parameter $\nu$ is an upper bounded of the fraction of allowed outliers in the training data and is also a lower bounded for the fraction of support vectors in the training data. If a value of $\nu = 0.1$ is considered, at most 10% of the training samples are allowed to be wrongly classified or can be considered as outliers by the decision boundary. And at least 10% of the training samples will act as support vectors.

After an adequate training on only normal data, samples that correspond to points outside the space enclosed by the hypersphere are classified as anomaly as they lie outside the expected samples.

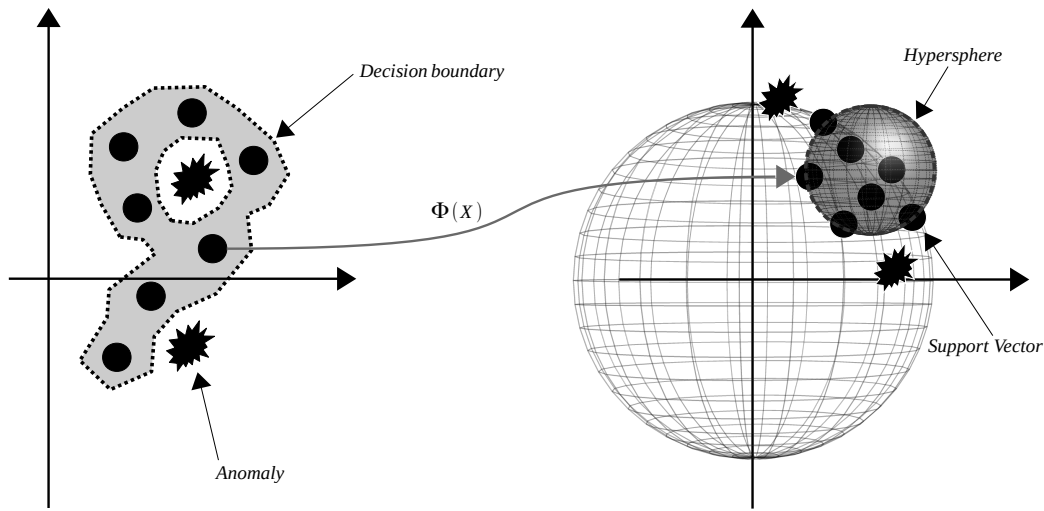

Figure 2.9: Visualization of the kernel trick

### 2.5.2 Isolation Forest

The isolation forest (IF) algorithm [52] is a framework for the identification of outliers in data sets. The method is based on the assumption that anomalies occur less often than normal data and that some features found in anomalies are statistically different.

Through the idea of repeated *isolation*, the model forms a baseline behaviour of normal and acceptable patterns in the training data. The principle of isolation is illustrated in figure 2.10 and shows that the input space is divided repeatedly into distinct chunks. Using the number of separations needed to isolate an individual point gives an estimate how unique it is. The geometric separation is logically equal to a binary decision tree, as demonstrated.

---

[12]Illustration adapted from [51].

To grow an isolation tree, one randomly takes a feature $F_i \in F$, also randomly chooses a boundary value $t \in [min(F_i), max(F_i)]$ for the feature and then divides all samples accordingly. The construction terminates if a leaf in the tree only contains one sample or all left samples are equal regarding the chosen feature.

The tree height is thus equal to the number of isolation iterations and will later be used as decision boundary. Individual trees can be constructed unfavourably, so an ensemble of decision trees (a forest) is evaluated when samples are classified. Thus, the method is named isolation forest and provides a way to implement the procedure.

New samples are said to be abnormal if the average height, i.e. the number of steps it takes to separate them from all other points per tree, is less than the established median height of trees for normal points.

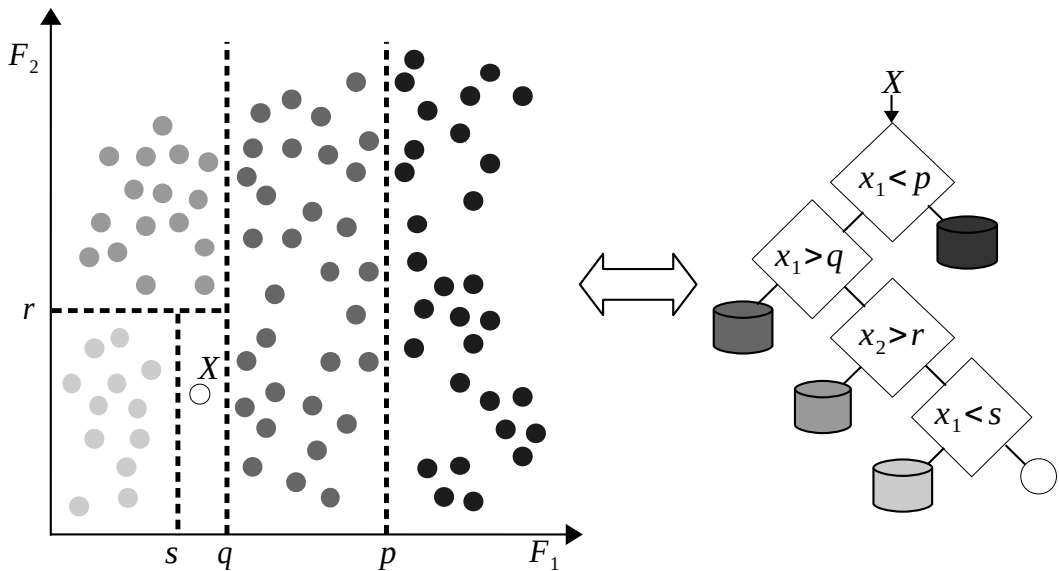

Figure 2.10: Principle of *isolation* and correlation to decision trees

### 2.5.3  Local Outlier Factor

The local outlier factor (LOF) is a statistical method for detecting outliers. It was proposed in 2000 [53] and is based on the idea that an outlier is an observation which lies outside a *neighbourhood* of nearby observations. The local outlier factor method can be used in many types of data sets including multivariate or time-series data, since no limitations on dimensions are given.

The *local outlier factor* is a statistical measure that determines whether a data point $P \in U$ can be considered anomalous when examined in light of its surrounding context. This score compares a data point's deviation within its local area to its global (or overall) deviation. The local density of each point is estimated by comparing it to its $k$-nearest neighbours, and the local density of all the points are compared to their respective $k$-nearest neighbours. For a point $X$, this density is computed using the following formula. The reachabilty factor between to points is defined as $reach_k(A, B) := max\{dst_k(B), \ d(A, B)\}$ where $d$ is a distance metric and $dst_k(B)$ is the distance of point B to its $k$-nearest neighbour. The $k$-neighbourhood $(N_k(B))$ denotes the set of points $N_k(B) = \{b \in U\{B\} \mid d(p, q) \leq dst_k(B)\}$ that are within the local neighbourhood of $B$. If multiple points have equal distances for a $k$-neighbourhood, then the following statement is true: $|N_k(B)| \geq k$.

The *local reachability density* ($lrd$) is calculated using $lrd_k(X) := 1/\left(\dfrac{\sum\limits_{B \in N_k(X)} reach_k(X,B)}{|N_k(X)|}\right)$.

Finally, $LOF(X)_k$ is equal to:

$$\frac{\sum\limits_{B \in N_k(X)} lrd(B)}{|N_k(X)| * lrd_k(X)}$$

A LOF value significantly greater than 1 indicates evidence of an outlier. Samples with a low-density zone surrounded by high-density zones for example produce larger LOF values. Figure 2.11[13] shows an example of the reachability function, which is used to determine the *local reachability density*. A LOF based AD model can implement a decision function by comparing samples to the average LOF scores of the learned normal distribution of samples or by using a threshold value.

The biggest disadvantage of LOF is that the resulting values are difficult to interpret. There is no clear cut-off between *insignificant* and *significant* outlier points; the significance of any point is highly dependent on the distribution of the data set, which can be both difficult to identify and understand.

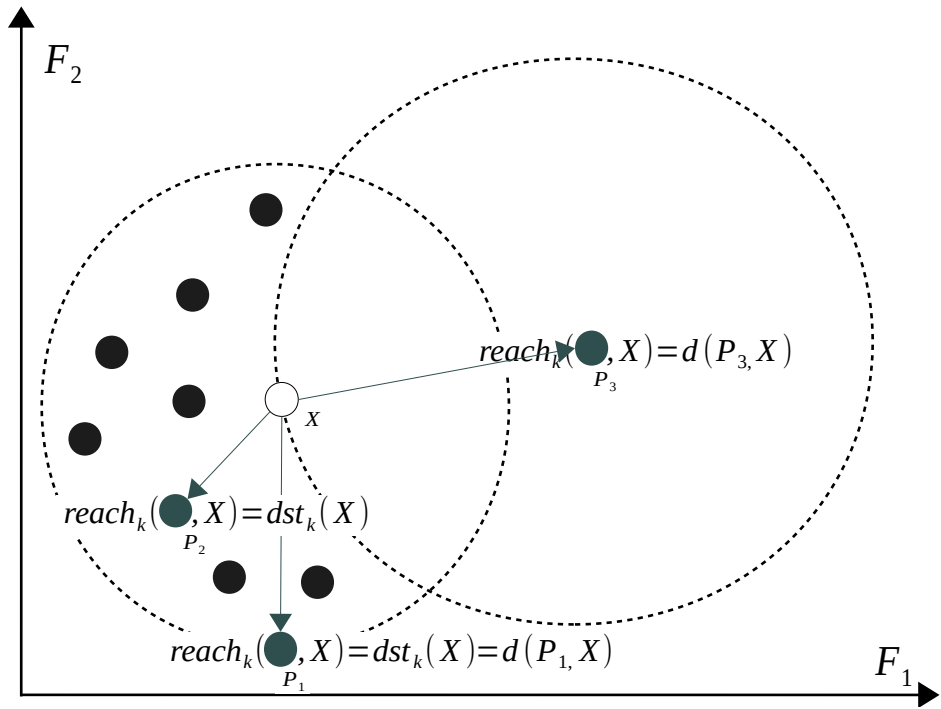

Figure 2.11: Reachability calculation for different points for $k = 9$

---

[13]Illustration adapted from [53].

### 2.5.4 Metrics

To compare the performance of a one-class classification algorithm, two different types of errors that can occur are determined.

Normal data points are called *negatives* while abnormal ones are called *positives* in the following sections. Figure 2.12 shows the *confusion matrix* for a binary classification problem and relations between all possible outcomes in a Venn diagram. The intersection of predicted and conditional classes are used to build four disjoint sets.

- True Negatives (TN)

  If no normal data (negatives) is predicted as abnormal by the OCC method, no error was made. The classificator correctly made the prediction that negative traffic is truly negative.

- True Positives (TP)

  If the prediction was made that a sample is positive when it is indeed abnormal, a true positive occurred. These data points are typically the most interesting ones, since these are the events targeted by design.

- False Positives (FP)

  A *type I* classification error appears if the prediction was made that a sample is positive while in fact it is conditional normal. If the model has over-fitted to its training examples, the amount of false positives tends to be larger.

- False Negatives (FN)

  If abnormal samples (positives) are classified as normal by the model, a *type II* error occurred. The better a model has trained, the fewer false negatives appear, since deviation from fitted data are identified.

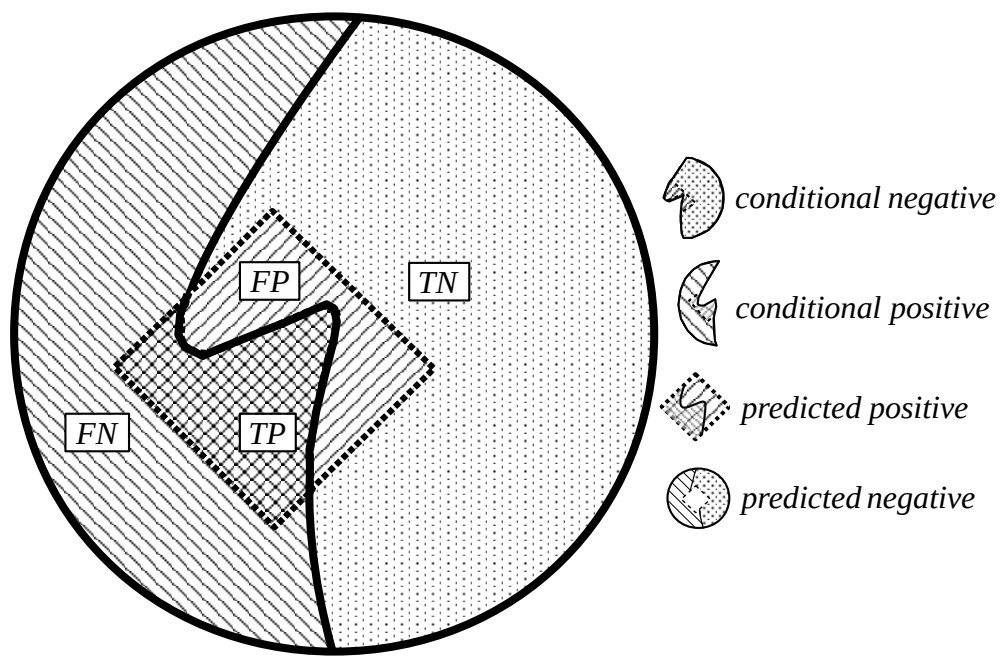

Figure 2.12: Venn diagram of a binary classification outcome

If these four disjoint outcomes can be counted, one can calculate different measures that indicate the performance of a particular method. For the context of intrusion detection and its evaluation, the following four metrics are commonly used, because they judge some demands listed in section 2.2.2. While a *type II* error can have heavy implications in applications like intrusion detection, for productive purposes a *type I* error is even costlier, since raised alerts often have to be investigated which involves human labour.

- Recall (RC)

  *Recall* measures the fraction of correctly selected anomalies regarding all conditional positive. This rate describes how many anomalies are correctly identified by the model. Recall is at 100% if a trivial model marks all samples as abnormal or no positive data is among the tested samples ($|TP| + |FN| = 0 \Rightarrow RC = 100\%$).

$$RC = \frac{|TP|}{|TP| \ + \ |FN|} \times 100\%$$

- Precision (PR)

  *Precision* describes the fraction between the amount of true positive samples among all alerted samples by the OCC algorithm. A model can achieve a perfect precision if either all selected samples are anomalies or if trivially no positive sample is selected ($|TP| + |FP| = 0 \Rightarrow PR = 100\%$).

$$PR = \frac{|TP|}{|TP| \ + \ |FP|} \times 100\%$$

- $F_1$-score (F1)

  *F1* measures the harmonic mean between recall and precision, and gives a single value to compare different classifiers. The trivial approaches described above would fail to produce an acceptable $F_1-$score, since they counter each other out. Since $F_1 \in [0, 1]$, a value of 1 denotes a perfect classification method.

$$F_1 = 2 \times \frac{PR \times RC}{PR \ + \ RC}$$

- False Positive Rate (FPR)

  The *false positive rate* measures the rate in which conditional negative samples are predicted as positives (false positives). This fraction can give a rough estimate on how usable a detector is. A low $FPR$ indicates a rather usable model. When no positive sample is selected, the rate is trivially 0.

$$FPR = \frac{|FP|}{|FP| \ + \ |TN|}$$

For other classification-based evaluations, *accuracy*[14] is often used to specify a model's performance. Given the imbalance of classes within intrusion-detection-based problems, accuracy is too coarse for a metric since the true negative ($TN$) will always overshadow true positives ($TP$). Accuracy thus skews the performance evaluation of OCC methods and will be disregarded in this thesis.

---

[14]$AC = \dfrac{|TP| \ + \ |TN|}{|TP| \ + \ |TN| \ + \ |FP| \ + \ |FN|}$

## 2.6 Transparency & Interpretability

The problem of transparency and interpretability of a trained model is concerned on how well statistical information and evaluation metrics can be inferred. Some models tend to be less transparent, depending on their design [54]. An overview about the interpretability of different data-driven models is explained in [55].

### 2.6.1 Approaches

Commonly, researchers have hypothesized that complex models are harder to interpret.[15] This trade-off thus also effects the classification performance, since complex models tend to outperform simpler heuristics.

In its most basic form one can think of a model interpreter as a method that maps concepts to features. For example, an image-based classification model can highlight features that contributed the most to its decision. The model interpreter for example can highlight pixels around the face or tail of a cat (the concept). A human operator can then, to some degree, understand why the image was classified as a cat [57].

Explainable artificial intelligence (XAI) has gained more and more relevance as machine learning systems impacted more real-life decisions. Especially for sensitive or critical infrastructures, a better understanding of a used model is necessary to have a chance to be applies in practice. A general overview about different topics in the field of XAI can be found in [58]. Explainability is linked directly to model interpretability and describes to what degree interpretable features can be transmitted for a given prediction. Explaining machine learning models can be done using two understandings [59]:

- *Mechanistic Understanding*

  Understanding what mechanism the model uses to solve a problem or implement a function. This investigation may involve an analysis of the underlying mathematical gadgets that were involved.

  When looking at nearest-neighbour-based models for example, it is clear that the approach uses defined geometric metrics as a density measure. This measure is internally used to correlate data points. The chosen density measure thus has a direct implication on the model's outcome.

- *Functional Understanding*

  The functional understanding describes how a model relates input vectors to the output variables. The functional understanding aims to help humans to interpret how a complex model forms a relationship between input and output signals. Specific decisions can be communicated by a visualization for example or a textual description.

One clear benefit is that XAI can deliver a second channel of information. The supplementary data can for example be used to bridge the semantic gap (c.f. section 2.2.2) which, among others, is a problem in anomaly detection. A second use case lies in the application of feature extraction methods. Since those systems automatically derive attributes about some data, XAI can help to indicate to humans on what basis some features were selected. Finally, also rising ethical considerations lead to a demand to question and influence the decisions of intelligent systems in order to avoid a biased and discriminatory classification. Deep-learning-based models encode their learned knowledge in terms of weights and biases of each neuron. It is difficult for humans to know why a specific neuron was activated and what the activation actually states with respect to the rest of the ANN. In non dead-learning-based learning schemes, like decision trees, one can comparatively easily comprehend why some action lead to a specific classification. For example when looking at an isolation forest model (c.f. section 2.5.2), one can query trees and look at what type of attributes *isolated* a sample from the distribution the quickest.

---

[15]Hypothesis: "The more complex and accurate a model is, the less it is interpretable."[56]

Frequent features thus indicate an explanation why a sample might be atypical. This assumes that the input features themselves are descriptive enough to offer a meaningful value. Since a key focus of this thesis lies within the topic of unsupervised feature extraction from data, the later proposed framework for representation learning will use methods from the domain of XAI to highlight what type of input features mattered for the feature learning algorithm the most. The subsequent anomaly detection mechanisms will not be investigated furthers, since all AD methods act on the same input, but use vastly different strategies in their detection mechanism.

Explanations can be evaluated in two ways: (1) How understandable is the explanation and (2) how complete the explanation is [54]. An explanation is more complete when one can anticipate how a model would react to a certain signal.

The ultimate goal, as stated, is to help human operators to better recognize why a certain decision was concluded by a model. Thus, the evaluation of interpretability is tied to the knowledge and cognition of the user. This fact introduces a heavy bias into the evaluation process. Lastly, XAI is divided into two problems [58]:

- *Model Analysis*

  "*What* does something predicted as a cat typically look like?"
  Possible approaches that answer such questions are to build prototypes of *typical* examples of a certain class. The model-specific prototypes then can provide insights on how a model *thinks* about a particular class.

- *Decision Analysis*

  "*Why* is a given image classified as a cat?"
  To answer this question one has to identify which input variables contribute to the prediction. Solving this problem often leads to actionable results that help to close the semantic gap.

Solving the presented problems can help to identify if a given model has trained in a correct manner. When finding prototypical classes for example, one can potentially see a variety of prototypes. These indicate that a model works as indented and learned different concepts.

The decision analysis can reveal if a model is for example over-fitted to the training examples. By evaluating what kind of features were used in a classification process, one can get a glimpse what the model thought. Thus, these methods are a valuable tool, which can be used to evaluate a model without the need to calculate metrics for a problem domain.

### 2.6.2 Layer-wise Relevance Propagation

For a period of time it was believed, that so-called *black box* approaches cannot be interpreted by human operators [60]. Several researchers tried to contradict this conjecture by developing methods that tackle this problem. One particular method will be highlighted during this thesis.

The layer-wise relevance propagation (LRP) [59, 61, 62] mechanism has been applied to several problem domains. The approach focuses on making deep neural networks more explainable. The LRP method tries to answer the question that the decision analysis problem poses: "What input vectors contributed the most to its decision ?". LRP thus also helps humans in factional understanding about the model. Besides LRP, other techniques were published that focus on the explainability of deep learning models. In [63] an overview about different XAI methods in the field of cybersecurity is presented. The conclusion indicates that LRP prevailed in several categories against 6 other frameworks.

LRP works, simply put, by reversing the neural network signal propagation. Instead of using forward propagation of input signals and observing which neurons in the output layer get activated, the LRP technique does the opposite. For a given sample $X$ and its activation $M(X) = \hat{X}$, the last output layer is populated with the values of $\hat{X}$ and preceding neurons are *re-activated* with these signals.

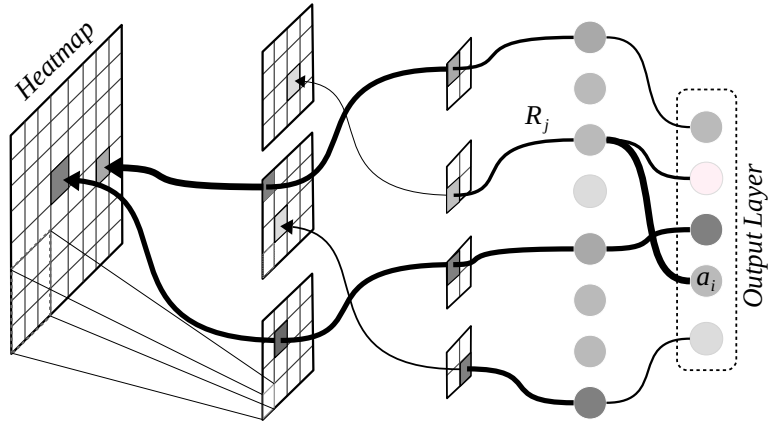

Figure 2.13: Schematic overview of the layer-wise relevance propagation technique

For each hidden layer, the more a neuron gets excited, the more it is relevant for the output. This relevance signal is propagated backwards through the network using defined rules [59]. Figure 2.13[16] shows the schematic overview, where the output values $a_i$ are propagated back through the network of neurons.

The last activation generates a mask or *heatmap* of input factors that had the most impact on generating the corresponding output signal $\hat{X}$. Relevance propagation follows the paradigm that the more a neuron is stimulated, the higher its relevance should be. The relevance ($R$) thus is a non-decreasing function of the neural net's predictions and weights.

The prediction $\hat{X}$ is decomposed into a sum of relevance scores ($r_i$) where $r_i$ denotes the relevance of input $x_i$.

For auxiliary layers like GroupNorm, special conditions have to be regarded in order to derive a meaningful output. A detailed explanation and further information on how the relevance is calculated back through the network can be obtained in [59]. The activation of a deep rectifier neuron *j* 2.4 is composed as:

$$a_k = max(0, \sum_0^j a_j w_{jk}) \text{ with } a_0 = 1 \text{ and define } w_{0k} \text{ to be the neuron bias.}$$

The sum runs over all lower-level connections and sums the weighted ($w_{jk}$) activations. The relevance $R_j$ for at this point in the network can be defined as:

$$R_j = \sum_k^j \frac{a_j w_{jk}}{\sum_0^j a_j w_{jk}} R_k \text{ where } R_0 = 1$$

Intermediate relevance $R_j$ will be conserved in the total relevance. Besides positive evidence for interesting input areas, LRP can also produce a negative relevance value. This *irrelevance* score can indicate what part of the input does not correspond with a given concept.

As stated in [58], the LRP method is a useful tool to evaluate the interpretability of a given model. With LRP, even non-deep-learning methods can be investigated using the method of *neuralizing* [64]. The idea is to express kernel-based methods, like SVM's, as deep neural networks. Once a model is translated as a differentiable ANN, the appliance of LRP is straightforward.

The heatmaps produced by LRP are not comparable to visualization of hidden layers, as they only show prototypical activation patterns [65]. When neurons are visualized a model analysis is conducted. The produced images can help to identify what prototypical evidence for a class can look like.

---

[16]Illustration adapted from [heatmapping.org/](heatmapping.org/)

# 3 Related Work

Anomaly based intrusion detection is a heavily studied field of research not only because of the fundamentally adversarial setting but also because of the different demands an anomaly-oriented IDS. The related work in the field of network-traffic-based anomaly detection is split up between approaches that either inspect the content of a packet on a *deep* level (deep packet inspection) or methods that use accumulated attributes of high-level aspects of network packets for the detection of intrusions (traffic analysis). A comprehensive survey on machine learning applications for networking security is presented in [66].

## 3.1 Traffic Analysis

One way to handle raw network data is to derive statistical features of the underlying collection of packets and to use this aggregated representation as the basis for anomaly detection. The benefit of so-called *traffic-analysis*-based approaches is that they usually work protocol agnostic and that they can process large amounts of traffic into aggregated quantities. These numbers are then used to train a model of normal behaviour. A key disadvantage, on the other hand, is that little to no payload-specific information is regarded in the detection method, which introduces a blind spot attackers may exploit.

Authors in [67] use the concept of *self-taught learning*, where first an unsupervised feature learning is deployed to reduce the given feature space. After this stage a supervised classifier is used to distinguish between classes of the data set. For the first stage a trained encoder network is used to extract latent vectors, which are then used in a softmax-based classification [68, 69]. The work uses the NSL-KDD data set [70] and the original 41 proposed features. The method is evaluated using $PR$, $RC$ and $F_1$-score.

Shallow anomaly detection with an autoencoder-based feature extraction was also explored in [71]. The authors used the KDD99 and NSL-KDD data sets and reduced the number of features to represent a sample with 28 dimensions. Afterwards, a supervised random forest method [72] is trained. Their results show that the classification accuracy does not drop when the unsupervised feature extraction was applied.

Using a hybrid model, researchers in [73] developed a method to extract spatial and temporal patterns of the NSL-KDD data set.

Applying visualization techniques on latent variables of the trained neural network were used to show the internal knowledge of the models. Ultimately their evaluation shows that using only convolutional layers is enough to extract features from aggregated network traces and to classify them with a *softmax* activation function.

Inspired by the few-shot image recognition research, a two-stage approach, presented in [74], explored representation learning on the NSL-KDD data set together with supervised SVM and 1-nearest neighbour based classification.

The authors in [75] benchmarked different ANN configurations on the NSL-KDD data set. The predefined features are ranged into two-dimensional grey scale *images* that represent chunks of packets. The research showed that supervised convolutional and recurrent architectures showed the best performance using several metrics.

In [76] the authors use linear deep neural networks to extract features from provided data. The research not only investigates a network-based intrusion detection, but also looks into host-based intrusion detection using the same general proposed architecture. For the evaluation, different publicly available data sets were used.

Borrowing concepts from natural language processing (NLP), the authors of [77] treat network traffic as *cyber language* and feed the resulting representation into a bidirectional LSTM-based architecture. They produce ordered sequences from raw flow data that follow the format $IP_a IP_b : Transport - Layer : \lfloor log_2(\#bytes) \rfloor$. A logarithmic loss function was used to assign incoming packets an anomaly score.

The *AEGR* framework, proposed by [78], uses an autoencoder architecture to derive features for anomaly detection via a density-based model (LOF). They also focus on the aspect that anomalies could hide in the training set and try to establish a robust loss function. The input for the AEGR network were already derived features from the NSL-KDD data set. The authors also evaluate their model with non-IDS related data sets using *accuracy* and the *receiver operating characteristic*.

The authors of *KITSUNE* [79] propose an online and end-to-end process for detecting anomalies. The basis for their method are incrementally updated statistics about various aspects of the network traffic. From an incoming packet some header fields are extracted and used to update the 23 different handcrafted statistics. For different window sizes the features are also aggregated further, totalling in 155 features. A mapping stage is then used to feed the features into an ensemble of autoencoders. A final autoencoder network is used to derive an anomaly score from the body of AEs to describe the state of the network.

## 3.2 Deep Packet Inspection

Deep packet inspection not only uses available information in headers of layers found inside network packets, but also use the specific bytes of the (decoded) payload to represent the traffic. In order to inspect fields of arbitrary network protocols, a parser has to be implemented which is resource intensive and involves manual labour. The literature proposes several methods, that are able to automatically derive features from network layers without the need of human expertise.

The authors of [80] published a survey paper were they compared successful and proven methods that use *n-gram* techniques on payloads of binary protocols as a detection basis. Thorough benchmarks on real world data they show that investigated methods are not feasible to detect attacks on moderately variable traffic. The authors argue that this is more promising to identify chunks that from a semantic unit and apply a detection on those.

With a focus on *n-gram*-based feature extraction, the authors in [81] use different embedding methods to represent raw payloads in ICS networks. Using the derived embeddings, they detect anomalies based on a similarity measure. Their evaluation compares different distance metrics and was able to identify anomalies in text- and binary-based application protocols.

In [82], the authors proposed an *n-gram* based method to derive feature from raw ICS traffic. They focused on deriving *k-prototypes* of protocol message types from the frequency-based *n-grams*. This not only allowed to remove noise from input vectors but to also derive rules on how to label their collected data set. Using a clustering method on the learned prototypes they could detect anomalies based on a distance metric.

Detection of sequence-based attacks is the focus in [83]. The authors propose a scheme based on probability based finite state machines that is able to identify attacks on ICS networks. With in-depth analysis of network data, they build models of normal communication behaviour between server and clients.

Discussing attack scenarios in industrial applications, the authors in [84], compare *matrix profiles* and *LSTM*-based approaches for intrusion detection on time series data. The time-series was constructed using the signal readings from sensors and activators of field devices and shows that LSTM are outperformed.

An ICS-focused method called *APAD* is proposed in [85]. Using raw signal readings from field devices, the researchers use the SWaT data set [86] to train autoencoders to model normal behaviour. Using a loss function, anomalies are identified if the measured loss is above an optimized threshold value. The method reaches near-perfect precision but only medium recall values.

*Matrix profiles* are also investigated in [87]. The authors apply the technique on analog sensor readings of ICS-related sensors in the SWaT testbed. The evaluation shows that the model is capable of detecting different attacks, but the authors also highlight that certain attacks are maybe easier to detect at higher network layers.

In [88], the SWaT data set is examined using *LSTM* and *OCSVM* methods. Log files generated by a data historian are the basis for an intrusion detection. The result shows, that ANNs produce less false positives than shallow methods while taking more time to train.

An unsupervised and byte-level intrusion detection scheme is presented in [89]. The authors utilize linear autoencoders to approximate the distribution of normal traffic behaviour, and later on use the reconstruction error as a measure to identify abnormal batches of network traffic. They report good results on detection of *inter-packet anomalies* (c.f. section 2.1.3), but remark that attack patterns have to be long enough in order to be detected.

With a focus on ICS networks a two-dimensional convolutional network is trained to extract features using a sliding window method [90]. For each packet, 23 features are parsed and used to represent its content. Notably the authors only focus on the one application layer protocol (*DNP3*) with their approach.

The work presented in [91] uses features chosen by domain experts of protocols in ICS traffic to represent network data. The researchers apply those features in a recurrent-neural-network-based intrusion detection. The supervised model is able to identify anomalies with a near-perfect $F_1$-score.

The authors of *RawPower* [92] explore feature extraction from raw data using deep neural networks. In their publication, the authors propose a spatial-focused extraction from network traffic using one-dimensional convolutional layers. In [93] the same authors also analyse recurrent architectures, which were able to extract temporal aspects of network traffic. Both approaches use supervised neural networks and take the learned representations of network traffic to classify them into binary categories. The conducted experiments showed that a flow-oriented representation of raw network traffic yield better results than a packet-level representation.

A thorough comparison between 6 classical supervised intrusion detection methods and a deep learning based architecture can be viewed in [94]. The experiments show that classical methods like Random Forest when trained on handcrafted features are outperformed by the *RawPower* architecture.

In [95], variational autoencoders are used to detect anomalies in IoT infrastructures. Their model uses data from both IT- and OT-networks, in order to capture side channels. The authors state that their preliminary results show that the *KingFisher* framework is able to identify attacks in both networks. The intrusion detection is archived using unsupervised learning, where anomalies are predicted using the residual error of the model.

Another supervised deep-learning-based architecture is proposed in [96]. The authors focus on how raw traffic flows can be used as input for convolutional neural networks. A single sample is represented using 32 packets and their first 512 bytes of the individual packets' transport and application layer. The neural network can compress the provided data into 256 values, which are used to classify flows into two categories using a cross-entropy loss function.

With a focus on time efficiency, the authors of [97] try to find an architecture that reaches a high accuracy while having a small computational footprint. Their CNN model takes raw network packets from the *USTC-TFC2016* data set [98] and evaluates how different window sizes affect the classification result. The authors compare their results to the *RawPower* network.

Utilizing a stacked autoencoder, the authors of [99] propose an unsupervised feature extraction from a session-based traffic representation. The first 983 bytes of a session, together with some specific header bytes are used to represent a sample. A supervised fine-tuning phase yields a model which is then used to evaluate the approach on the *CTU-13* data set.[100]

An automatic feature extraction framework is presented in [101]. Using a one-dimensional version of stacked autoencoders, the authors extract feature from raw traces. Using a supervised softmax-based classification they could identify encrypted services like *BitTorrent* or *Skype* without any expert knowledge.

With the aim to identify encrypted network communication, the authors in [102] present a one-dimensional CNN architecture which learns to extract features from raw network traffic. The researchers argue that the sequential nature of a byte stream from network data is best processed by one-dimensional architectures and show in experiments that two-dimensional architectures are inferior. The researchers chose the *ISCXVPN2016* data set [103] for the evaluation.

The Authors in [104] use a convolutional ANN that was inspired by LeNet-5 [105]. Their work explores how network traffic is best represented for the representation learning and supervised softmax-based classification. They identify, that flow-based modelling of all network traffic scores the highest accuracy. The evaluation was done using the *USTC-TFC2016* data set which consists of traffic from benign and malicious applications.

Inspired by the success in image classification, the model proposed in [106] uses the *ResNet* [107] architecture to extract features from raw network traffic and use the learned representation to detect anomalous samples in it. The packet-oriented approach is evaluated using a private data set. Through a grid search, used parameters of the ANN were determined.

Using the concept of data augmentation, the authors in [108] propose a semi-supervised model. They train a linear classifier on latent representations obtained from an autoencoder network. With the help of generated anomalous samples obtained by variational autoencoder, the model had two classes to be trained on.

Evaluated on the *DARPA1998* [70] and *ISCX2012* [109] data sets, a spatial-temporal model is able to extract features from raw network traffic [110]. Bytes of traffic flows are encoded using a one-hot method and are fed into CNN layers. The output is then bundled and used for an LSTM layer in order to focus on temporal patterns of network traffic. A final *softmax-layer* is then used for a supervised classification of the provided data. The model is compared to other published methods and yields a better *FPR* while needing less training time.

An NLP-based approach is used in [111] to identify which bytes in a packet's payload are anomalous. The *ATPAD* framework is trained on payloads of web applications via an *attention mechanism* [112]. This gives the model the ability to mark parts of the payload and thus helps investigators explain why a request may be malicious. The model is evaluated on the *CIC-IDS-2017* and *CSIC-2010* data sets [113, 114] using a supervised softmax classification.

Packet2Vec [115], as the name suggests, builds upon the popular *word2vec* [116] concept. For a stream of packets, n-grams are used to break individual packets up into chunks for fixed length. These chunks are then used as input for the word2vec algorithm to build up a vectorized representation. A packet is then represented by the weighted average over all n-grams of the packet and can be used for a classification task. The authors compare a Random Forest and a Naïve Bayes classifier on the trained representation and achieve unusable result in the context of intrusion detection, but acknowledge that it is an initial step in the domain.

Utilizing embedding heuristics of raw payloads frames, the authors in [117] explore different deep learning based architectures in the context of intrusion detection. The evaluation shows that long-term and short-term dependency within extracted features are best discovered by LSTM-based models. All investigated models use a softmax-based approach to detect anomalies.

Authors in [118] also utilize the concept of word embeddings of protocol header fields in their proposed method. The authors focus on a packet-oriented method to process incoming traffic to address shortcomings of flow-oriented approaches. The supervised LSTM-based model takes the embedded tokens and extracts features of them. Lastly, a softmax layer is used to classify in coming samples. The evaluation shows a near perfect detection rate with a minimal false-positive rate.

With the focus on DoS attacks caused by an abundance of insecure IoT devices the others in [119] propose a flow oriented approach for an early detection mechanism. By sampling the first n packets within a flow and extracting the first 80 bytes from a packet the authors train an one-dimensional convolutional ANN to extract features from the given data. Lastly the authors also train an autoencoder network, in order to learn patterns of benign traffic. The unsupervised model uses a combination of cross entropy loss and mean square error to from a decision boundary and report a perfect detection result.

A *wavelet*-based method is presented in [120], which extracts features from a window of packets. The extracted features are analysed using an SVM in order to determine if they are anomalous. The authors compare different possible kernels for the SVM and archive a high-precision value but fail to measure the recall of the method.

## 3.3 Overview

Related works in the field of *content-sensitive* intrusion detection used *n-gram* based methods to extract features in an unsupervised fashion in the past. Their performance decreased as network traffic became more diverse [82]. Advancements in the field of deep learning showed the possibility of another kind of unsupervised representation learning, using hidden layers of neural networks [107]. Many proposed methods for the task of intrusion detection that use deep models, still rely on labels to train a supervised so-called *end-to-end* model. These kinds of models not only extract features, but also distinguish between various classes using a supervised *softmax-based* classification [67, 73, 101, 104, 110, 112, 117].

The seminal work by [121] presented during a BlackHat USA talk was arguably the first to publicly present the connection between unsupervised representation learning used in computer vision to the domain of network security. Using a linear feed-forward architecture, features are extracted from raw network data rather than aggregated statistical features selected by hand.

The authors in [78] were among the first to investigate the use of a hybrid model for anomaly detection. Their approach uses a deep autoencoder to extract features from a given high-level representation of network data. This learned model was then used to extract features, which were applied for a shallow AD. Their work does not take raw byte information as an input, but rather takes the 41 features of the NSL-KDD data set as a basis.

The raw feature learning proposed in [89] has limited power to learn temporal aspects of network traffic as it was trained using a large batch size. The end-to-end model conducts AD using the model's internal loss function. The larger the reconstruction error is, the higher is the anomaly score, that is assigned for a batch of samples. Table 3.1 list an overview about the presented works. For each paper the listed metrics, the used method, and shortcomings are noted. The later proposed method is designed to address the identified shortcomings.

| Author | Data Set | Features | Method | Metrics | Shortcomings |
|---|---|---|---|---|---|
| [108] | CICIDS2018 | statistical | AE | ROC | not protocol agnostic |
| [79] | NSL-KDD | statistical | Assemble of AEs | FNR, FPR, ROC | not protocol agnostic |
| [67] | NSL-KDD | statistical | ANN | RC, PC, F1 | not sequence aware |
| [71] | NSL-KDD | statistical | ANN+RF | AC, PR, RC, F1 | not sequence aware |
| [73] | NSL-KDD | statistical | CNN+RNN | AC, PR, RC, F1 | not sequence aware |
| [74] | NSL-KDD | statistical | CNN+SVM/kNN | AC | not sequence aware |
| [75] | NSL-KDD | statistical | AE, CNN, LSTM | ROC, AC, | not sequence aware |
| [78] | NSL-KDD | statistical | AE+LOF | PR, ROC | not sequence aware |
| [76] | NSL-KDD, UNSW-NB15 | statistical | ANN | AC, PR, F1, TPR, FPR, ROC | not sequence aware |
| [84] | own ICS Data | raw signals | matrix profile & LSTM | - | not protocol agnostic |
| [85] | SWaT | raw signals | AE | AC, RC, FPR | not protocol agnostic |
| [87] | SWaT | raw signals | matrix profile | - | not protocol agnostic |
| [88] | SWaT | raw signals | LSTM & OCSVM | PR, RC, F1 | not protocol agnostic |
| [111] | CIC-IDS-2017, CSIC-2010 | raw payload | attention RNN | PR, FPR | softmax-based training |
| [117] | CSIC 2010 | raw payload | LSTM+CNN | PR, RC, F1, AC | softmax-based training |
| [92, 93] | USTCTFC2016 | raw frames& raw flows | 1D-CNN & 1D-CNN+LSTM | ROC | not sequence aware |
| [110] | ISCX2012 | raw frames & raw flows | CNN+LSTM | AC, PR, FPR | softmax-based training |
| [94] | MAWILab | raw flows | 1D-CNN | ROC | not sequence aware |
| [96] | own Data | raw flows | CNN | PR, RC, AC | not sequence aware |
| [119] | USTC-TFC 2016, Mirai-CCU | raw flows | CNN+AE | PR, RC, F1, FPR | only header information |
| [89] | SwaT, CSET2016 | raw bytes | AE | PR, RC, F1 | intra-packet anomalies |
| [90] | own ICS Data | raw bytes | CNN | PR, RC | not sequence aware |
| [77] | ISCX IDS | raw bytes | Bi-LSTM | ROC | only statistical features |
| [115] | DARPA 2009 | raw bytes | word2vec+RF & NB | PR, RC, F1, ROC | not sequence aware |
| [101] | ISCXVPN2016 | raw bytes | AE+1D-CNN | PR, RC, F1 | not sequence aware |
| [102] | ISCXVPN2016 | raw bytes | 1D-CNN & 2D-CNN | AC, PR | not sequence aware |
| [99] | ISCXIDS2012, CTU-13 | raw bytes | AE | AC, PR, RC, F1 | not sequence aware |
| [97] | USTC-TFC2016 | raw bytes | 1D-CNN & 2D-CNN | AC | not sequence aware |
| [120] | DARPA 99' | raw bytes | wavelet+SVM | RC | not sequence aware |
| [80] | DARPA 99' | raw bytes | n-gram | RC, FPR | not sequence aware |
| [83] | own ICS Data | raw bytes | DFA | - | parser-based |
| [118] | USTC-TFC2016 | raw bytes | word-embedding+ LSTM | PR, RC, F1, FPR | softmax-based training |
| [82] | own ICS Data | n-gram of bytes | Count-Min Sketch | ROC, AUC | not sequence aware |
| [81] | Web07, Aut09 | n-gram of bytes | Clustering | AC, FPR | not sequence aware |
| [91] | own ICS Data | parsed | LSTM+FNN | PR, RC, F1 | not protocol agnostic |
| [95] | own ICS Data | parsed | VAE | FP, FN | not protocol agnostic |

Table 3.1: Overview about the presented related works

# 4 Data Description

This chapter gives insight into the domain-specific data sets this thesis focuses on. The proposed framework in chapter 5 will be evaluated on two data sets. First a public data set which was collected from a test environment in Singapore. And secondly, a not yet disclosed data set which was recorded at a power plant in Germany. Since neither data set has any labelled anomalous data, heuristics have been applied to bridge this shortcoming. The labels are only necessary for the overall model analysis, the training is done in a fully unsupervised fashion. Both data sets are ICS-related. As argued in section 2.1 this fact produces an easier to learn normal baseline. In [122] current private and public cybersecurity related data sets are discussed in detail. They also identify that many IDS related data sets are not labelled well enough for a sound evaluation. The prominent KDD data set [70] is not evaluated since it only offers aggregated features and does not contain traffic from industrial networks.

## 4.1 SWaT

The Secure Water Treatment (SWaT) research testbed was established in 2015 by the Center for Research in Cyber Security of the Singapore University of Technology and Design (*iTrust*). The SWaT testbed is a scaled-down replica of a real water treatment facility. The testbed simulates a modern treatment plant and has the following procedures:

- P1: Raw water intake

- P2: Chemical disinfection

- P3: Ultrafiltration

- P4: Dechlorination using ultraviolet lamps

- P5: Purification by reverse osmosis

- P6: Ultrafiltration membrane backwash and cleaning

The research testbed includes water purification equipment, several levels of communication networks, PLCs, a Supervisory Control and Data Acquisition (SCADA) workstation, HMI workstations, and storage of historical enterprise machines. More information about the SWaT testbed can be found in their initial publication [123]. Figure A.1, which was taken from the published technical details[1], shows a network overview of the testbed. The network consists of two layers: An Ethernet-based PLC network (*Zone-A*) and an SCADA environment (*Zone-B*). For each procedure, two PLCs are installed to ensure a redundant architecture. PLC's communicate over an application protocol called *EtherNet/IP*, which uses TCP as the transport protocol. The *Zone-A* network shares a switch with the *Zone-B* network, which is established for SCADA software and the human machine interface.

During its operation the iTrust research laboratory published numerous data sets, which were gathered from the testbed. Some of these data sets only contain raw readings of sensor values, while others also include network packet captures. Today the SWaT data sets are among the most researched data sets for anomaly detection in the domain of network security.[2]

---

[1] itrust.sutd.edu.sg/wp-content/uploads/sites/3/2020/07/SWaT_technical_details-160720-v4.3.pdf
[2] itrust.sutd.edu.sg/itrust-labs_datasets/

In December 2019, the researchers published the so-called *SWaT A6* data set [123]. These records not only contain periods of benign traffic captures, but also include periods where malware was used to attack the water treatment plant. Unfortunately, the provided data has no labels on the Ethernet level, but does provide a timetable where specific events are annotated on time basis. The proposed framework in chapter 5 will be evaluated on the second attack[3]. While the first attack exfiltrates data using the *ICMP* protocol, the second phase consists of a download of malicious software and the appliance of the malware. The goal of the malware was to disrupt the PLC sensor readings (c.f. $C_1$-communication threat in section 2.1.1). For the training of the model, benign data captures[4] were used. In these two PCAPs the research laboratory states that no attack was conducted.

Figure 4.1 shows a visualization of statistical characteristics of the traffic capture containing the attacks. Around the 50 second mark a spike is visible. This marks the mentioned download event, this was also verified by hand. The disruption of sensor readings is not visible in the graph.

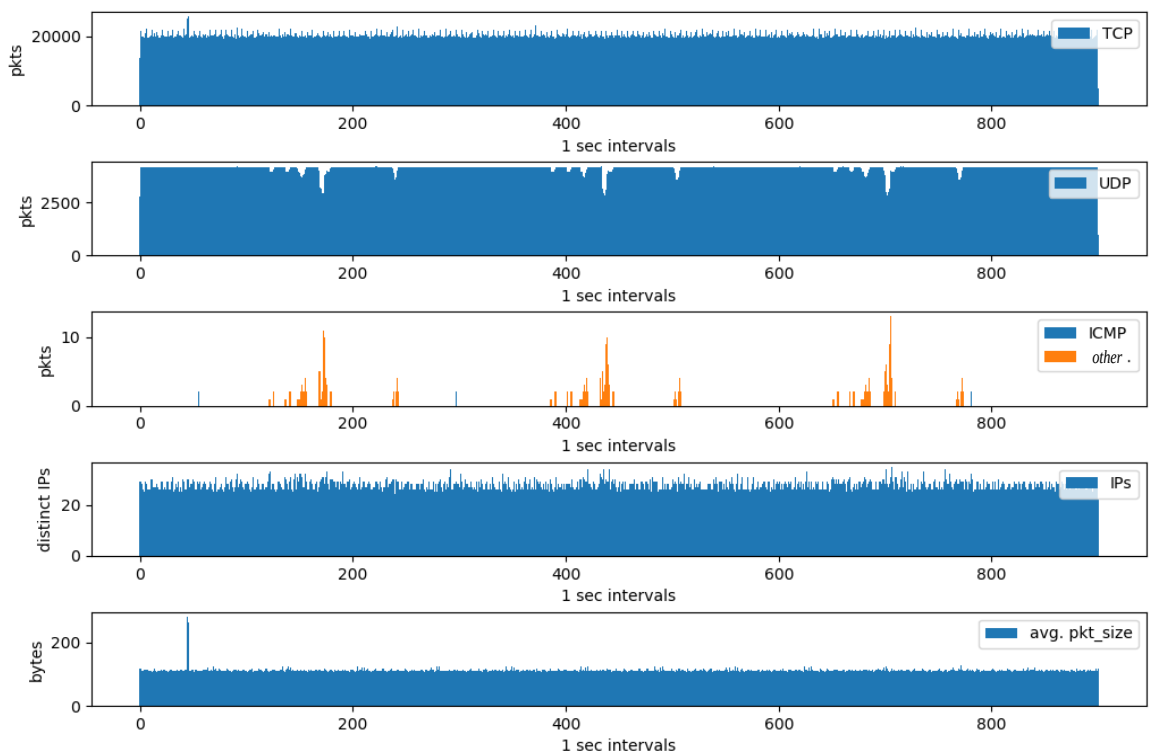

Figure 4.1: High-level traffic analysis of the SWaT data set

## 4.2 Voerde

The *Voerde* data set was collected in 2017, during a research project of the chair of Computer Networks at Brandenburg University of Technology (BTU). The trace of network traffic was captured for a period of 2 hours and 39 minutes. During the packet capture, the network was in part attacked by a delegated *penetration test* from a security company as part of a vulnerability analysis. Figure 4.2 visualizes statistical characteristics of the used data set.

---

[3]The first 300 seconds of PCAP *Dec2019__00010__20191206123000-005*

[4]PCAP *Dec2019__00007__20191206114500-015* and *Dec2019__00008__20191206120000-009*

Section 4.5 describes the data set in more detail, and also highlights how the data has been split for the training and testing of the machine learning models. The data set was chosen in order to evaluate the model on real world ICS network traffic.

Compared to the characteristics visualized in 4.1, the Voerde-based traffic has less UDP-based traffic. The TCP traffic has also fewer fluctuations in segments were no attack was conducted. The Voerde data also shows a continuous ICMP traffic, whereas the SWaT data only has periodic events.

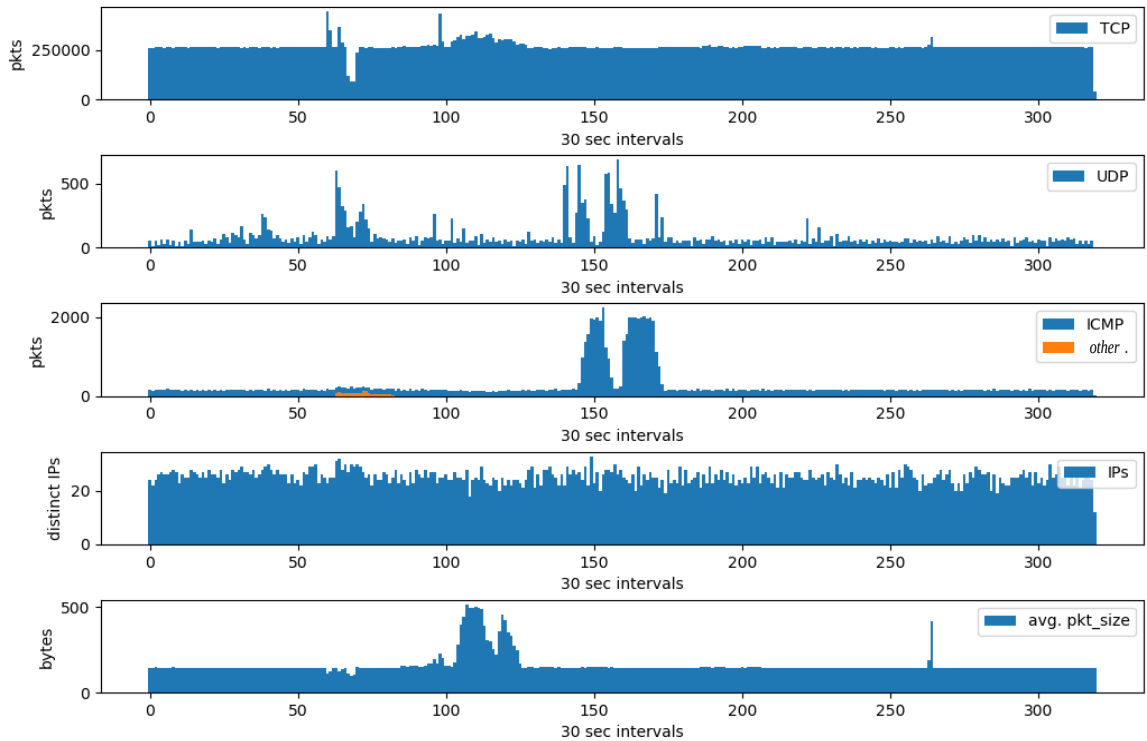

Figure 4.2: High-level traffic analysis of the Voerde data set

## 4.3 Estimated Ground Truth

Since the network traces are not labelled on a packet level, one has to rely on heuristics that estimate labels for the data set. A drawback is that the established metrics from section 2.12 lose some of their expressiveness. An estimation-based labelling was also conducted in [89]. Since unsupervised problems are inherently hard to evaluate, many authors argue that one has to rely on heuristic to generate some form of ground truth data [124], [125], [126], [127]. A semi-automated labelling mechanisms that utilizes a trained model itself to annotated traffic, like proposed in [89], was not considered for the evaluation since it would introduce an unwanted bias into the evaluation. Secondly, the authors only are able to produce labels that flag frames based on the transport layer properties. These properties do not reflect characteristics of attacks on PLCs within the network of the *SWaT A6* trace.

Since many *real* anomalies will not be labelled correctly when an estimation is used, the *type II* error (false negatives) is in-fact higher than when using the true but unknown labelling. False positives likewise have to be investigated, since these events might be true anomalies even though they are not labelled as such.

**Voerde**

As presented in figure 4.2 it is visible that between from approximately the 2250 second (75 *sec* mark) until second 5250 (175 *sec*) in the aggregated plot, the network traffic is irregular. During this period an authorized third party attacked the network. The attacker used an external IP address to access the network. Although the usage of a foreign address is rather simple to detect, it will be the basis for an estimated ground truth labelling and is called $Eval_H$.

**SWaT**

For the malware download in the beginning of the trace (see figure 4.1), all frames in this period are said to be anomalous. Therefore, the estimated ground truth contains 18900 Ethernet frames labelled as positive. The data set will later be referred to as the SWaT $Eval_H$ split. The second attack in the A6 trace was chosen since it is closer related to established CPS threats.

## 4.4 Synthetic Data Generation

Additionally to the roughly labelled evaluation data ($Eval_H$) that mostly focuses on statistical differences, the SWaT and Voerde data sets were also used to systematically insert attacks within each trace using a specially developed tool [128]. The inserted Ethernet frames are referred as *synthetic* anomalies. An advantage of these generated anomalies is that for each of the inserted frame a specific label can be assigned which helps in the evaluation process. Additionally, these generated anomalies mainly express characteristics that cannot be detected with statistical features and rely on temporal or byte-specific detection methods.

Three different types of attacks were added to the data using the framework [128]. The usage of synthetic data in the context of intrusion detection is also discussed in [129].

The first synthetic data set focuses on DoS attacks ($Eval_D$), where a number of frames were inserted that target a specific host in the network. A second synthetic trace was generated, which introduced an MitM-style attack. In this eavesdropping attack ($Eval_E$) the third party listens on the communication between two devices. Lastly, the $Eval_R$ trace was generated whereas an MitM attack a third party is installed to drop frames between communicating devices. The later proposed method is target to detect these inter-packet anomalies (c.f. section 2.1.3).

## 4.5 Characteristics

Table 4.1 and 4.2 give an overview about the selected sets from the used body of network traces. The sets were split into disjoint buckets, in order to be applied to several data driven methods. *Train* denotes the samples that were exclusively used to train the framework. The *Validation* set also consists of only benign samples and is used to test if the model is over-fitted. For the *Voerde* data set, the feature extraction ($Train_{RL}$) and anomaly detection ($Train_{AD}$) training sets are equal. The choice was made because the Voerde set has only limited uninterrupted normal traffic. The Voerde $Eval_H$ set contains traces where a delegated penetration test has been conducted. Those network-probing events are intra-packet anomalies that are contained within a single frame. While the training traces follow a regular pattern, the spikes indicate irregularities in the evaluation data (c.f. figure 4.2).

|                    | *Train*     | *Validation* | *Eval$_H$*       | *Eval$_D$*      | *Eval$_E$*    | *Eval$_R$*     |
|--------------------|-------------|--------------|------------------|-----------------|---------------|----------------|
| Duration [sec]     | 1139.7      | 162          | 2907             | 702             | 702           | 702            |
| Size [GB]          | 1.7         | 0.24         | 0.72             | 1.2             | 1.2           | 1.2            |
| # Frames           | 10,774,388  | 1,539,556    | 1,518,761        | 3,521,649       | 6,157,648     | 5,232,729      |
| # Anomalies (%)    | 0           | 0            | 424,364 (27.9)   | 135,048 (3.8)   | 2,460 (0.04)  | 1,930 (0.037)  |
| # MACs             | 39          | 33           | 110              | 37              | 57            | 54             |
| # IP               | 42          | 34           | 34               | 38              | 40            | 40             |
| # Protocols        | 25          | 19           | 21               | 20              | 22            | 22             |
| # Flows [tcp/udp]  | 993/417     | 144/51       | 6,648/39         | 255/95          | 448/161       | 377/136        |
| ATU (std)          | 146.3 (166) | 145.6 (158)  | 477.8 (1228)     | 144.6 (12)      | 146.3(0.5)    | 146.3(0.5)     |
| $k$packets/sec     | 0.4         | 9.5          | 0.5              | 5.0             | 8.7           | 7.5            |

Table 4.1: Voerde traffic statistics per sub set

|                    | *Train$_{RL}$* | *Validation$_{RL}$* | *Train$_{AD}$* | *Validation$_{AD}$* |
|--------------------|----------------|---------------------|----------------|---------------------|
| Duration [sec]     | 898.6          | 4.1                 | 408.5          | 4.1                 |
| Size [GB]          | 1.4            | 0.14                | 1.4            | 0.14                |
| # Frames           | 10,000,000     | 100,000             | 10,000,000     | 100,000             |
| # Anomalies [%]    | 0              | 0                   | 0              | 0                   |
| # MACs             | 54             | 31                  | 57             | 33                  |
| # IP               | 51             | 32                  | 60             | 35                  |
| # Protocols        | 36             | 21                  | 39             | 21                  |
| # Flows [tcp/udp]  | 87,961 / 172   | 971/19              | 87,368/236     | 1,572/21            |
| ATU (std)          | 113.23 (105)   | 108.23 (104)        | 113.38 (105)   | 112.55 (105)        |
| $k$packets/sec     | 11.1           | 71.4                | 24.0           | 24.0                |

|                    | *Eval$_H$*      | *Eval$_D$*      | *Eval$_E$*     | *Eval$_R$*     |
|--------------------|-----------------|-----------------|----------------|----------------|
| Duration [sec]     | 299.7           | 487.6           | 487.2          | 486.9          |
| Size [GB]          | 1.1             | 1.3             | 1.3            | 1.3            |
| # Frames           | 7,350,000       | 6,284,294       | 6,194,072      | 6,194,576      |
| # Anomalies [%]    | 18,900 (0.25)   | 170,608 (2.7)   | 516 (0.008)    | 148 (0.002)    |
| # MACs             | 56              | 56              | 67             | 67             |
| # IP               | 52              | 48              | 48             | 48             |
| # Protocols        | 37              | 34              | 36             | 33             |
| # Flows [tcp/udp]  | 70,316/148      | 28,649/119      | 28,649/119     | 28,649/122     |
| ATU (std)          | 114.44 (117)    | 113.65 (322)    | 113.12 (101)   | 113.41 (103)   |
| $k$packets/sec     | 24.0            | 12.8            | 12.7           | 12.7           |

Table 4.2: SWaT traffic statistics per sub set

# 5 Methodology

This chapter describes the proposed method for an unsupervised feature learning on raw network traffic. Learned and extracted features are afterwards used for a shallow anomaly detection. The framework aims at utilizing strengths of deep-learning-based methods to learn representations of network data. Already established methods for unsupervised one-class classification (see section 2.5) are evaluated on data which is produced by the framework.

## 5.1 Representation Learning Comparison

There are several approaches to the general problem of representation learning. In order to identify the most promising one for the domain-specific task, a comparison experiment is conducted. Using a small hold-out data set[1] of the SWaT A6 data that encompasses $N = 26,249$ rows. For every frame, the first $D = 1,024$ hexadecimal values were extracted and padded with zeros if a frame contained less information. Column $D_i$ was populated with the $i$-$th$ byte within a frame. All the hexadecimal values were normalized in the interval between 0 and 1 (see section 2.3.6). The dimension reduction was targeted to project $d = 64$ output columns for each row. The initial comparison was done in order to determine different properties about the tested algorithms and to ultimately identify a suitable method for the proposed research questions.

Prominent representatives in the field of dimension reduction, such as different singular value decomposition methods like: principal component analysis (PCA), kernel-based PCA (kPCA) and independent component analysis (ICA), as well as locally linear embedding (LLE), Isomap and spectral embedding (SE) have been compared. A descriptive overview about these algorithms is presented in [130].

Representing deep-learning-based models, an autoencoder (AE) network (see section 2.4.3) with $\theta$ parameters was chosen for this test. Since in section 2.3.5 it was established that comparing DR methods without an explicit target application is a non-trivial task, the evaluation looked at statistical information like: mean, median, minimum and maximum value, standard deviation (std) and the interquartile range (IQR). Secondly, characteristics about the method itself, like the type of transformation, the ability to process new samples (OOS), or to be retrained on new data and complexity demands, were compared. Table 5.1 gives an overview about different dimensions for each feature learning implementation. More insights about the space and time complexity can be obtained in [131–133]. Since feature extraction is ultimately targeted for the task of intrusion detection, the ability to process out-of-sample (OOS) inputs is also compared (see section 2.3.5).

Some listed methods (Isomap, SE) could not be judged fully since the used implementation exceeded the memory limit imposed by the used hardware.[2] The result of the comparative experiment exposes that an AE-based feature extraction is best suited for representation learning. The found representation (compression) by the AE showed for three of the six statistical characteristics the best results compared to the original, uncompressed data whereas no other algorithm could prove to be the best one for more than one of these measures.

Additionally, for autoencoder the space complexity is not bound by the data set size, while the time complexity is linear and can be adjusted using the number of epochs (see section 2.4.1). Hence, in the course of the framework development, an AE-based approach was chosen for the part of representation learning of network traffic based on its computational and statistical advantages.

---

[1] PCAP *Dec2019_00000_20191206100500-013*
[2] A system equipped with 24 GB of RAM.

|  | mean | median | min | max | std | IQR |
|---|---|---|---|---|---|---|
| PCA | $-1.26 \cdot 10^{-7}$ | <u>0.4823</u> | $-4.0832$ | $7.1323$ | $0.3601$ | $0.4690$ |
| k-PCA | $-1.85 \cdot 10^{-8}$ | $-0.0003$ | $-4.0831$ | $7.1324$ | $0.3670$ | <u>0.4781</u> |
| ICA | $-1.51 \cdot 10^{-9}$ | $-2.60 \cdot 10^{-5}$ | <u>$-0.0539$</u> | $0.0661$ | $0.00617$ | $0.008$ |
| Isomap | *n/a* | *n/a* | *n/a* | *n/a* | *n/a* | *n/a* |
| LLE | $1.66 \cdot 10^{-7}$ | $-2.74 \cdot 10^{-5}$ | $-0.3514$ | $0.3098$ | $0.0061$ | $0.0025$ |
| SE | *n/a* | *n/a* | *n/a* | *n/a* | *n/a* | *n/a* |
| AE | <u>0.0162</u> | $0.0016$ | $-0.833$ | <u>0.7694</u> | <u>0.341</u> | $0.4591$ |
| *original* | $0.4844$ | $0.4823$ | $0$ | $1$ | $0.2924$ | $0.5098$ |

|  | non-linearity | OOS processing | ability to retrain | space complexity | time complexity |
|---|---|---|---|---|---|
| PCA |  | ✗ |  | $\mathcal{O}(N*D)$ | $\mathcal{O}(N*D+D^3)$ |
| k-PCA | ✗ | ✗ |  | $\mathcal{O}(N^2)$ | $\mathcal{O}(N^3)$ |
| ICA | ✗ | ✗ |  | $\mathcal{O}(N^2)$ | $\mathcal{O}(D^2*N)$ |
| Isomap | ✗ | ✗ |  | $\mathcal{O}(d*N^2)$ | $\mathcal{O}(N^2)$ |
| LLE | ✗ |  |  | $\mathcal{O}(N^2)$ | $\mathcal{O}(d*N^2)$ |
| SE | ✗ |  |  | $\mathcal{O}(N^2)$ | $\mathcal{O}(N^3)$ |
| AE | ✗ | ✗ | ✗ | $\mathcal{O}(2*\theta)$ | $\mathcal{O}(epoch*D*N)$ |

Table 5.1: Comparison of different dimension reduction methods

## 5.2 Feature Extraction Framework (pcapAE)

This thesis introduces the so-called *pcapAE* framework, which is designed to be trained on raw traffic in order to learn representations and thus extract valuable features. Learned features can afterwards be used for intrusion detection tasks. The framework does not conduct a deep packet inspection using a protocol parser, but rather takes raw application payload and header data. Which was also investigated by other authors [85, 89, 90, 92, 96, 101, 134] with the idea that one has access to more information which results in a better chance to detect sophisticated anomalies. Unlike previous frameworks [110, 111, 117, 118], the *pcapAE* aims to work fully unsupervised and only focuses on feature extraction. Little expert knowledge is used in the framework, the aim was to build a fully protocol-agnostic approach. With the conception of complete and raw Ethernet frame and a central placement of the IDS, one can expect to detect intra- and inter-packet anomalies. In comparison, an IDS that only acts on sensor readings is limited in detecting anomalies, and possible miss MitM attacks for example. On the contrary, considering nearly all OSI layers of the frame covers the monitoring of communication patterns and concrete sensor values at the same time. The proposed method builds on several key observations that are the foundation for the framework:

- With high-level traffic characteristics, zero-day attacks likely are not being reflected by a change in communication and therefore cannot be detected. Thus the framework focuses on using minimal preprocessed raw inputs belonging to any protocol level. The aim is to process protocols on a content-level without the need of a specific parser, in order to derive features that can be used for a content-sensitive AD.

- Since hexadecimal values are used to represent packets in network captures, one can trivially present them as images containing grey-scale patterns, as for a byte value $v \in [0, 2^8 - 1]_{hex} = [0, 255]_{dec}$, one can assign a color $v = R = G = B$ in the RGB spectrum. Therefore unsupervised computer-vision-based machine learning can be used to process this form of data. As the preliminary evaluation showed, deep-learning-based feature extraction also yields attributes which are useful in the context of intrusion detection. A detailed overview about the fragment generation is presented in section 5.2.1.

- Convolutional neurons can be used to detect spatial patterns (see section 2.4.2), while recurrent neurons can be used to learn sequential aspects of data (c.f. section 2.4.2). Hence, the model will use a combination of both neurons in an autoencoder-type network architecture. Temporal patterns are a prominent aspect to network communication and thus learning sequential features of protocols is important for detection inter-packet anomalies. Section 5.2.2 explains in detail how the actual feature extraction is performed.

- Through means of XAI (see section 2.6.2) the framework will be able to use latent information to map relevance values to input samples. These values can be used to aid human operators to investigate the alerted anomalies.

Non-trivial difficulties faced within the framework are the application of *spatial-temporal* feature learning through an autoencoder, and the integration of ways to explain the decisions made by the feature learning framework.

## 5.2.1 Fragment Generation (FG)

The thesis will compare two different ways to represent raw network traffic. The following sections will use the term *fragment* to describe a chunk of hexadecimal bytes values extracted from a stream of Ethernet frames. In this context, fragments are two-dimensional raw representations of network data. Other deep-learning-based methods use one-dimensional input representations [102]. They argue that in context of network data, this would be a more natural depiction. For the pcapAE a CNN-based layer with two-dimensional convolutions was selected as it is prominently used in image processing tasks. A trivial benefit of two-dimensional representations is a natural visualization of the data. Since the input dimensions tend to dictate parts of the ANN model's architecture (input layer), the fragment size was fixed to 32 rows by 32 columns. $1024 = 32^2$ bytes values corresponds to 7 times the average frame size for the Voerde data and 9 times the *ATU* for the SWaT data (c.f section 4.5). Since the fragment size is larger than the average frame, one fragment can contain information belonging to multiple frames. For the chosen fragment size, the used autoencoder thus can have a smooth and gradual bottleneck (c.f. section 2.4.3).

Figure 5.1 illustrates the different kinds of data representations. A detailed explanation how the two types of representation are built follows in the next section. Before fragments are fed into the representation learning model, all values $f_{ij} \in F^{32x32}$ are normalized (see 2.3.6) into the interval of real numbers between 0 and 1 by dividing each value with the largest possible hexadecimal byte value $(2^8 = 255)$ $[f_{ij} \in \mathbb{R} : 0 \leq f_{ij} \leq 1]$. This is done to help the ANN to train the data in more efficient manner. Each fragment contains normalized bytes with 32 rows and columns. Only the last fragment is padded with zeros.

Special protocol fields like IP and MAC addresses are not sanitized, unlike previous works [93] suggests. Because of the fixed network topology found in ICS networks (c.f. section 2.1), were only a limited set of hosts are registered in the domain, new addresses within the communication flow are a highly relevant feature for anomaly detection. Packets are also not reassembled before fragments are built, fragments therefore reflect the mixed communications of the network. A packet-oriented fragment extraction was not evaluated within this thesis, since this type of preprocessing needs to be in tune with the ATU. For this preprocessing, if the fragment size $D$ is larger than the *ATU*, fragments are padded since ANNs expect the input to be in a certain shape. Otherwise, if the $ATU >> D$ then a lot of information is discarded by this extraction design.

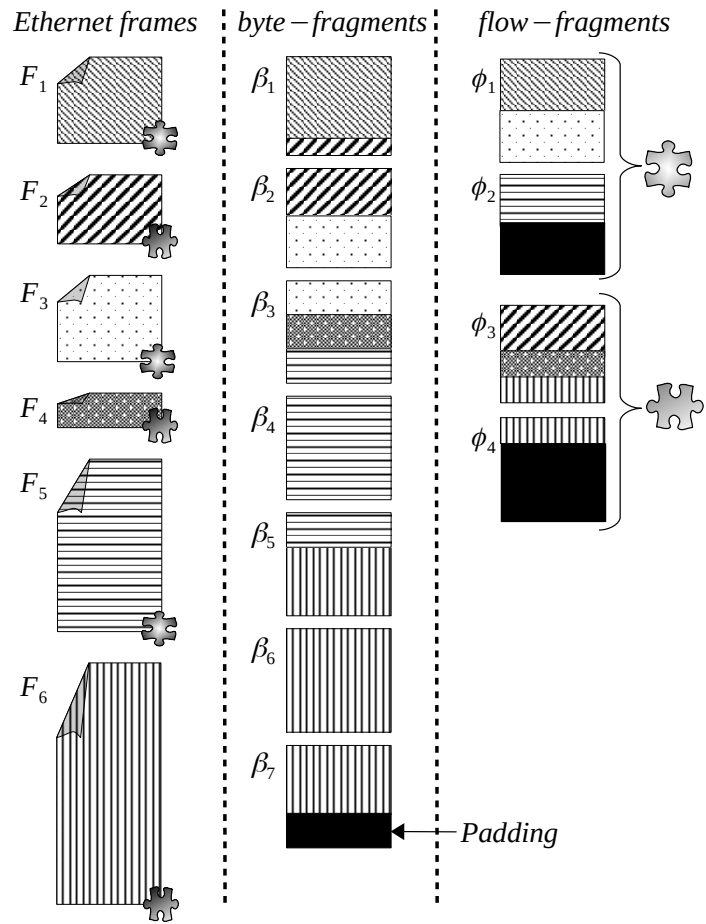

Figure 5.1: Visualisation of the transformation for byte and flow-fragments

**Byte-Fragment**

With a focus on raw byte streams, this preprocessing heuristic transforms the incoming stream of Ethernet frames into fixed-length fragments of size 32 by 32. The first 1024 bytes are used for every frame. If frames contain fewer bytes, fragments host information from multiple frames. An advantage the byte-oriented processing offers is, that no $MTU$ has to be considered. Also, the complete network data is processed using this representation as no protocol layer is skipped. This implicates that also information from the link- and network-layer is considered, which contains more metadata about communications (c.f. section 2.1.2). Figure 5.2 shows how patterns in byte-fragments are expressed when converted into images. Although the four examples look random, on closer inspection one can see darker clusters of bytes, which hints that they do contain some structure. Byte-fragments might be better suited to capture intra-packet anomalies as they do not discard any information.

**Flow-Fragment**

The flow-oriented representation of network data uses a five tuple (network flows) consisting of the transport protocol, source address and port as well as the destination address and port to sort incoming frames into distinct bins. Fragments are built up from the first 64 bytes of the network-layer data (c.f. section 2.3) of each packet from a given flow aggregation.

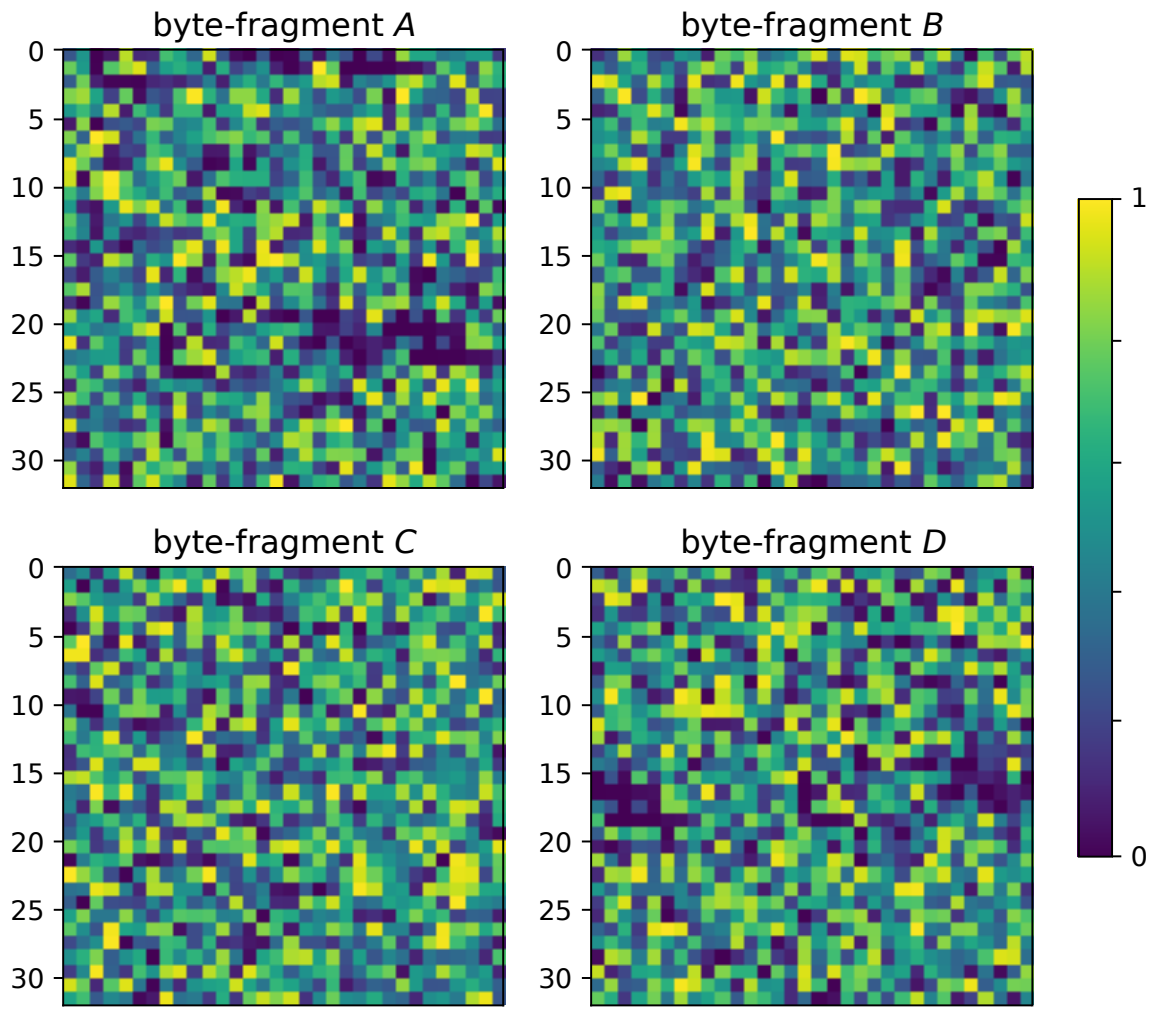

Figure 5.2: Four consecutive examples of byte-fragments shaded using a *viridis* color map

The lower layers are excluded, since the flow-focused preprocessing uses this information indirectly when building its fragments. The remaining data contains information about the TCP or UDP session as well as the application information that for example can contain control commands for a PLC. In figure 5.3, the internal structure of the flow-oriented extraction is highlighted. It is clearly visible that, compared to the byte-oriented representation, a clear structure is present. The transport protocol header and payload display distinct patters within the fragments. Intra-packet anomalies may also be better represented with flow-fragments since the semantic order of frames is valued higher within this type of fragment. Intra-packet anomalies may better be represented with flow-fragments since this representation, in contrast to the byte-representation, maps semantically identical information of the IP payload to always the same fragment areas.

Although one can expect that this more stringent representation yields a more precise input for the machine learning algorithms, it is also possible that an autoencoder is even capable to equally derive condensed characteristics from the more fluently mapped byte-fragments. If the ladder is the case, the effort for flow-based preprocessing would be expendable.

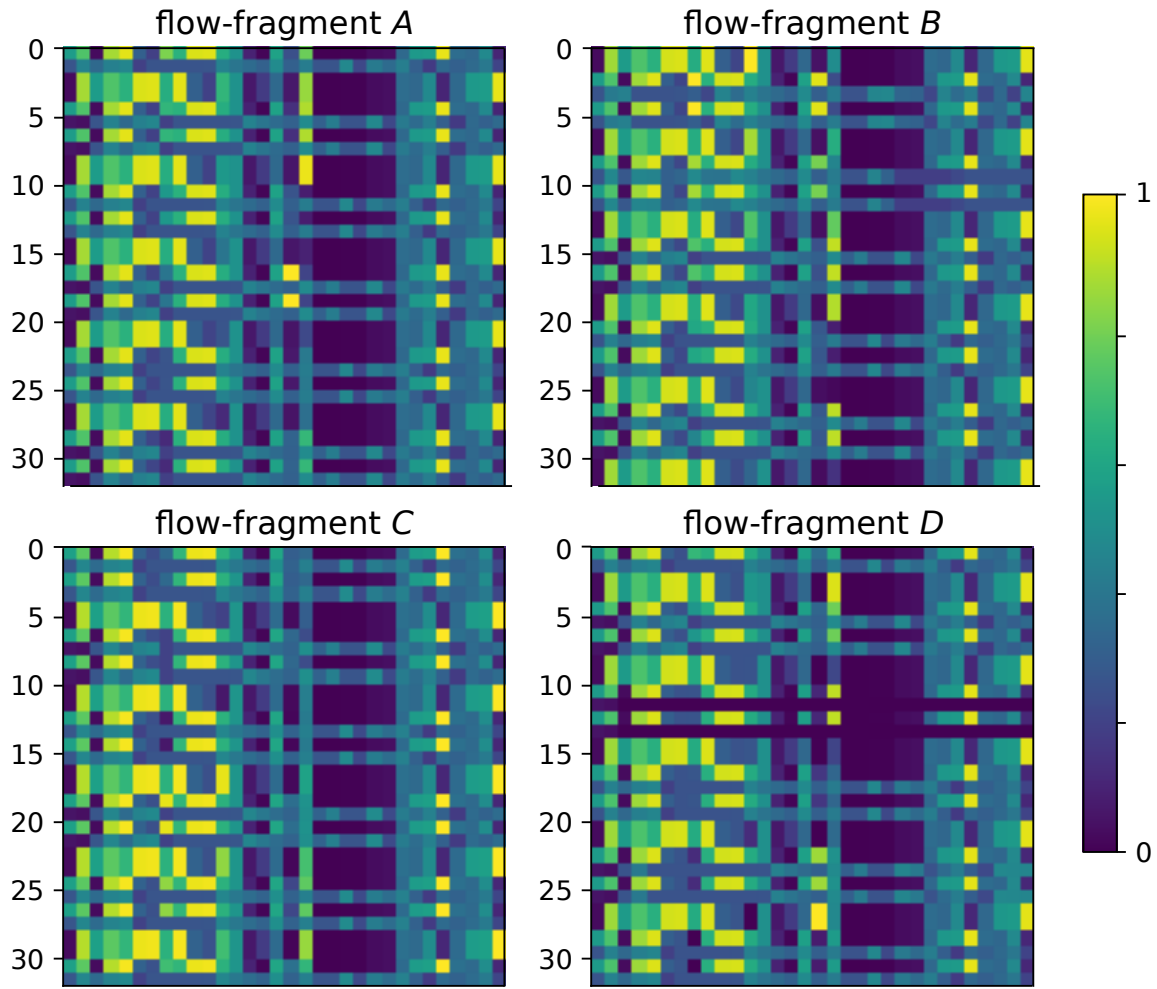

Figure 5.3: Four consecutive examples of flow-fragments shaded using a *viridis* color map

## 5.2.2 Unsupervised Feature Extraction

The model for the pcapAE framework is an autoencoder. As established in section 2.4.3, deep architectures are designed to learn a compressed representation of the provided data set. The encoder network is designed to down-sample the two-dimensional input fragment of size 32 by 32 ($1,024$ bytes) to a hidden representation of size $8^2 = 64$ features. The network uses 4 hidden layers to achieve this compression of input features, while the decoder network on the opposite uses also 4 hidden layers to up-sample the code to its original shape. The decoder network, as stated is only used to train the encoder, as the reconstructed samples is compared to the original and the resulting loss is used to adjust the weights of the ANN.

After the model has trained on a subset of the data, the *trained* encoder network can be used to compress new samples. The latent variables produced by the encoder encapsulate patterns of the input fragments. These *codes* (c.f. section 2.4.3) can then be used to train a downstream anomaly detection algorithm.

Since the network extracts 64 features from $1,024$ input variables, the extracted representations is thus 16 times smaller than the original input dimension. After each layer a dropout layer is deployed in order to prevent the ANN from over-fitting. Group normalization layers are used to aid the training process. Exact dimensions and the number of weights for each layer ($\sim 20,000$ adjustable parameters) is listed in the appendix under section C.1. The pcapAE's autoencoder is designed for a spatial-temporal extraction of features in order to capture aspects of both intra- and inter-packet anomalies. The following section describes how this spatial-temporal learning is conducted.

**Spatial-Temporal Feature Learning**

The proposed framework is based on concepts presented in [135]. Here, recurrent Long short-term memorys (LSTMs) cells [136] are combined with neurons that perform convolutions transformations (see section 2.4.2). Their so-called *convLSTM* cell combines spatial feature extraction for local learning of patterns with sequential learning using an RNN. The RNN works as expected, but the multiplications ($\otimes$) at each gate are replaced with convolutional operations. The convolutions are placed at input-to-gate transitions. The proposed method was first used for precipitation forecasting. Later experiments showed that a convRNN-based model is able to learn a variety of tasks [137]. Other research augmented the convLSTM architecture and also replaced the LSTM base cell with a GRU cell (*convGRU*). A convGRU cell can be written as:

$$r_t = \sigma_r(W_r \star [h_{t-1}; x_t] + b_r)$$
$$z_t = \sigma_z(W_z \star [h_{t-1}; x_t] + b_z)$$
$$\hat{h}_t = \phi(W_h \star [x_t; r_t \otimes h_{t-1}] + b_h)$$
$$h_t = (1 - z_t) \otimes h_{t-1} \oplus z_t \otimes \hat{h}_t$$

where brackets denote a vector concatenation and $\star$ the convolution operation. These convolutions reduce the number of dimensions using small kernel sizes. Resulting feature maps therefore have to be padded to preserve matrix multiplication rules.

The proposed framework will utilize the introduced type of spatial-temporal neuron in its autoencoder design. This has the advantage that not only patterns from within snippets of data are extracted, but also that sequential aspects of the underlying data are observed for the task of representation learning. The recurrent architecture is able to consume a series of fragments with a variable length $n$. $S_n$ therefore denotes a model with trained on sequences of length $n$. For longer sequences of samples the encoder uses to produce a final latent variable (code), it is harder to find adequate weights that encode the data set. From the sequence of samples, the decoder network is trained to always reconstruct the last input fragment of a series of fragments.

Since individual fragments can be made up from multiple frames, and multiple fragments can be used to form a hidden representation, the code thus holds information about a sequence of frames. This is also the case for a sequence length of $n = 1$, as the fragment size is larger than the frame size. For each fragment $F_i$ that is part of a sequence F-sequence$_x = [F_i | i \in [x, x+n]]$ with $x \in [0, |F| - n]$ where the total number of fragments (samples) is equal to $|F|$, the network uses its internal latent variables ($H_i$) at a given step in a recurrent manner. For fragments $F_i$, the latent variable from previous steps ($H_{i-1}$) are fed back into the GRU cell (c.f. section 2.4.2). The initial latent variable ($H_0$), for a the first fragment in a sequence, is set to be a null matrix. This has the effect that sequences of fragments are independent of each other. Depending on the latent variable, the reset and update gates get more or less excited.

This recurrent design is exploited to capture temporal aspects about the data, while the convolutional aggregations are used to pick up spatial novelties. Using the last latent variable of the encoder network, decoder is trained to reconstruct the last input fragment in a sequence. This is achieved by calculating the error between the generated fragment and the last input fragment. This error term (residual loss) is then used backpropagation step.

Figure 5.4 depicts this mechanism for sequence length of $n = 3$. The last hidden state ($H_3$) is the compressed representation (*code*) which is used for the feature extraction process. The figure also shows that information from more than one frame is present within one code. Lastly, it can be noted that only the encoder network is used for the feature extraction process — the decoder network is only need for the training loop.

Since the recurrent architecture uses its internal hidden state for each new fragment, different-ordered sequences using the same fragments will produce different codes. The histogram in figure 5.5 highlights this fact by comparing the loss of a trained model that is presented novel sequences and sequences that contain randomly ordered fragments. From the comparison one can see that a trained model expects certain ordered sequences, as the loss values are often lower than in random sequences. Thus, the plot highlights that the temporal aspects are important in the extraction process.

Two models are compared in the later evaluation. One model will be trained on a minimal sequence length of $n = 1$, while a second model will use a length of $n = 3$. The idea behind these two approaches is to test if models that are trained on longer sequences are better in detecting inter-packet anomalies, like MitM attacks.

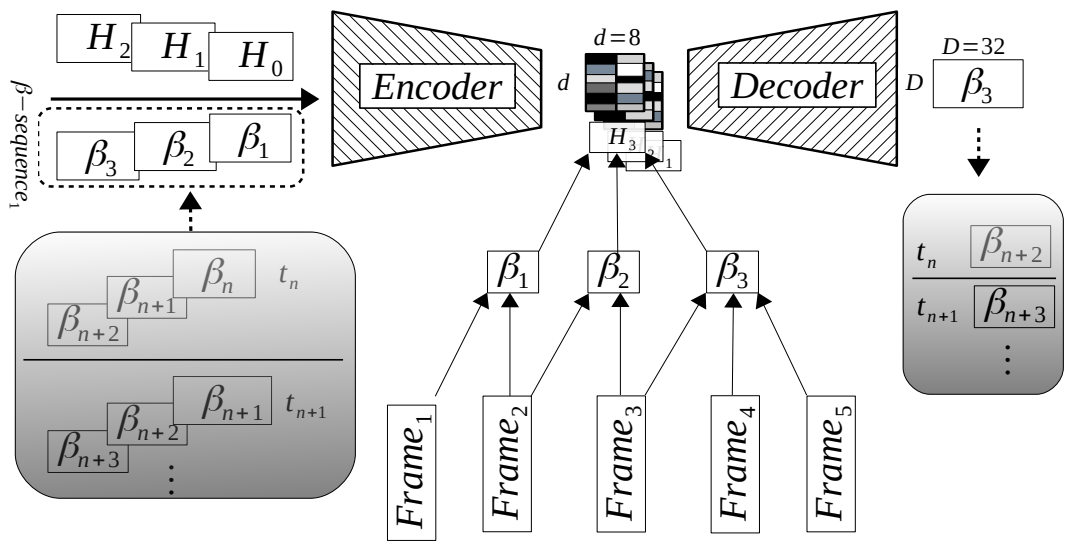

Figure 5.4: Extracting features from a sequence of fragments with length 3 using the networks hidden representations

### 5.2.3 Feature Relevance

As argued in chapter (see demand two of list 2.2.2 and section 2.6) a decision analysis on the model's behaviour is a crucial factor for its usability.

The framework will therefore be able to produce heatmaps about the processed input samples, which can used to understand the model's decision. These heatmaps will shed light which input values contributed the most in the feature extraction mechanism. To generate said heatmaps relevance propagation (c.f. section 2.6.2) is utilized. By taking the encoder network and a given sequence of fragments, LRP can calculate what regions within an input fragment contributed the most for extracted features.

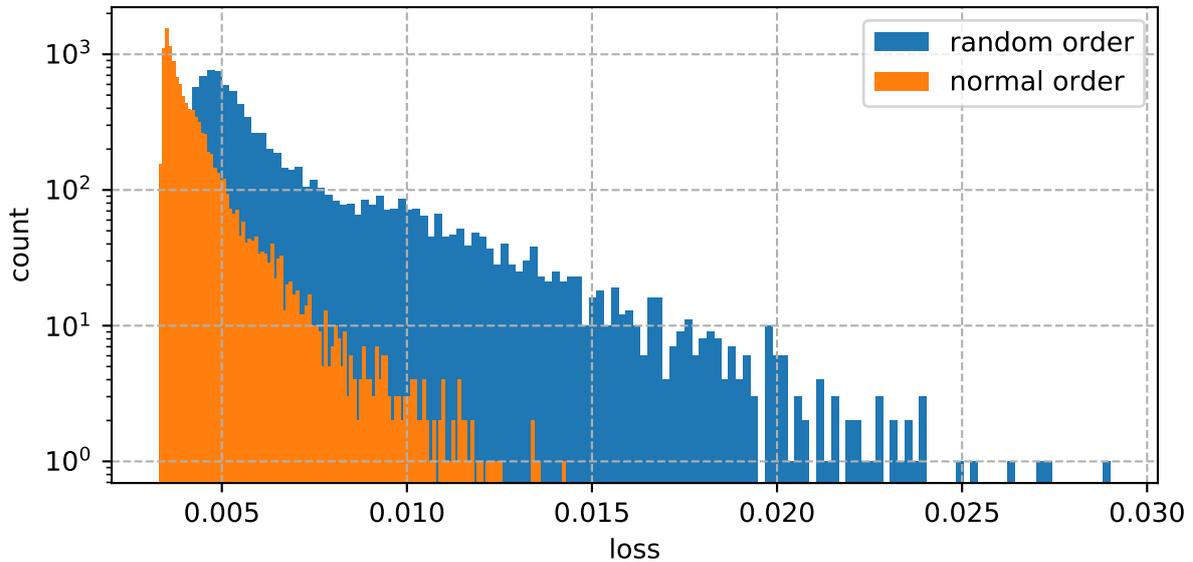

Figure 5.5: Loss distribution for expected and randomly-ordered fragment sequences

If a model is trained to reconstruct the last fragment of a sequence, LRP is also only able to generate a heatmap for the last fragment. The produced heatmap can later also be used to enrich the generated alerts about potential anomalies and thus help to close the semantic gap.

### 5.2.4 Method Orchestration

The proposed pcapAE framework can be used in three operations. Figure 5.6 depicts each stage in detail. Each stage starts by generating *byte-* or *flow-oriented* fragments from a set of PCAP files. The subsequent anomaly detection is independent of the chosen data representation as the pcapAE extracts the same number of features for either type. A chosen sequence length $n$ also does not have to be regarded in the data-preprocessing step, since this hyperparameter is implemented in a programmatic fashion inside the pcapAE autoencoder training procedure.

Generated alerts by the arbitrary downstream anomaly detection consists of four parts: anomaly score, model loss, relevance heatmap, and the suspected input packet identifiers.

1. *AE training:*

   This first stage is used to train an autoencoder network on preprocessed PCAP data. This results in an trained encoder network which can be used in later stages.

2. *AD training:*

   The second stage uses the fully trained autoencoder to compress new PCAP data. The compressed representations, which encapsulate features of the initial network data using only a fraction of dimensions, can be used to build a *new* data set. Using this compressed numerical data, an unspecific anomaly detection algorithm can then be trained to learn a model of benign patterns.

3. *Evaluation:*

   Lastly, in stage three, the complete pipeline can be used to evaluate new data in order to generate alerts for potential anomalies. These alerts can be enriched using heatmaps produced by the trained encoder network.

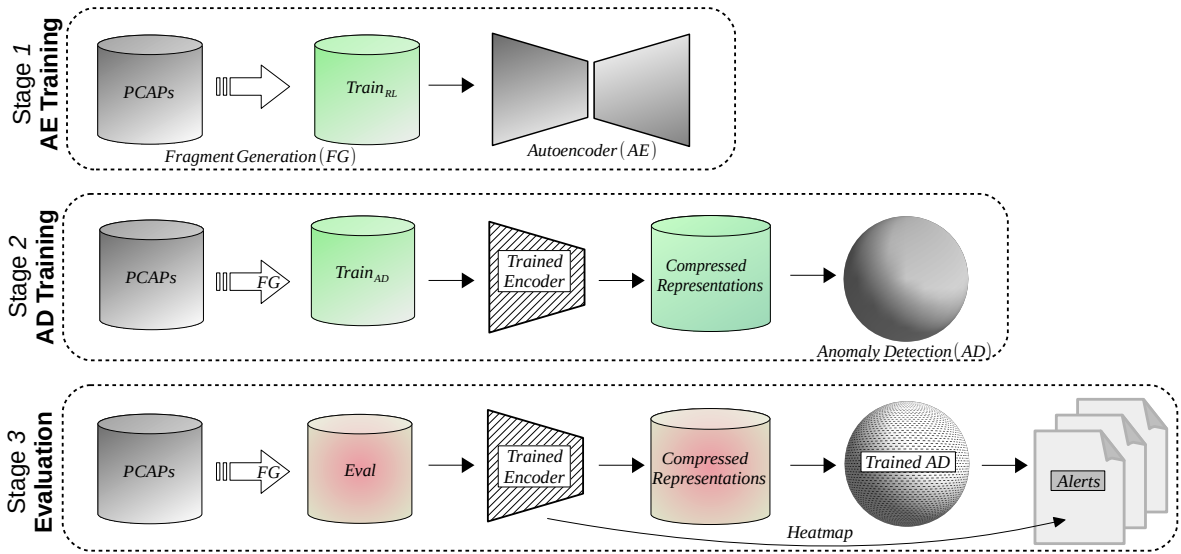

Figure 5.6: Overview about the order of events for each stage of the framework

## 5.2.5 Implementation Details

The framework is implemented using *python3* since it has a rich machine learning ecosystem and supports bindings for hardware accelerate computing. To handle the reading of network data the *dpkt* library [138] showed performance advantages over other available libraries. The PCAP preprocessor can process a given list of PCAP files in a parallel fashion. On average, the preprocessor consumes ∼2,000 frames/sec per computing core.

The convGRU-based autoencoder was implemented using the *PyTorch* library [139]. The integration of the LRP method was done using a public implementation.[3] The downstream anomaly detection was achieved using the implementation published within the *scikit-learn* [140] machine learning library.

The autoencoder training procedure was implemented to use graphical processing units. The experiments and evaluation were conducted using an *NVIDIA RTX 3090*. To track and visualize different experiment trails, the *Tensorboard* library was used. An overview of one experiment trail can be viewed visiting the uploaded Tensorboard.[4]

The *pcapAE* framework can be obtained by visiting the public code repository.[5] The implementation was made public since reproducibility and access is a common problem within academia [141]. Along with implementation details, supplementary *jupyter notebooks* were also published to show the different stages.

From a data point of view, a set of *PCAP* files is first transformed into a *HDF5* [142] binary data store of decimal values. The preprocessed data is then given to the autoencoder interface, which generates a trained *PyTorch* model. Using the trained encoder, new data can then be transformed into another *HDF5* data store containing compressed representations of real values. This data can then be applied to train an anomaly detection algorithm. Finally, a set of alerts for a given network trace is generated using the hybrid model.

---

[3] github.com/dmitrysarov/LRP__decomposition
[4] tensorboard.dev/experiment/OTD0qLBeQYK1DajlvkwoyQ
[5] github.com/dreizehnutters/pcapAE

# 6 Evaluation

This chapter shows the experiential results for the introduced data sets from chapter 4. In particular three anomaly detection methods (see section 2.5) were evaluated on learned representations generated using the introduced framework from chapter 5. Two minimal data preprocessing heuristics were compared within this evaluation (see section 5.2.1), as well as two different models are compared using an input fragment length of $n = 1$ and $n = 3$ (c.f. section 5.2.2). For all anomaly detection methods, model-specific hyperparameters were determined after an initial grid search. This grid search used compressed data from a fully trained pcapAE encoder. The used hyperparameters are noted in the appendix under section B.1. To compare different runs, the $F_1$-score was used as a metric.

Authors of [89] used the *SWaT A3* data trace to train their end-to-end model. To evaluate their model, they used the SUTD Security Showdown 2017 data set (*S317*). In order to compare different models, a common data set is necessary, therefore the proposed framework was set out to be trained and evaluated in the same setting. Figure 6.1 shows, the residual loss (MSE) over time, plotted for mixed-data and a SWaT A6 only evaluation. As presented in the figure, the loss for fragments belonging to the *S317* data set were always above the validation set for supposedly normal and abnormal samples. The pcapAE model either seemed to be *over-fitted* on the SWaT A3 train data, or the *S317* traces are truly different from a network protocol point of view. When any autoencoder was trained and tested on SWaT A6 data, fragments that do contain patterns of anomalous traffic, produce a higher residual loss, while benign traffic was reconstructed just like trained examples.
Since the mixed evaluation using two data sets was not applicable, the thesis used just one SWaT data set (A6) for its evaluation. The A6 data set was preferred to the A3 data, as it also contained roughly labelled anomalies.

## 6.1 Experimental Setup

For each of the 4 pcapAE models (2 fragment types and 2 temporal learning approaches) that were examined, 3 shallow anomaly detection methods were evaluated. For each of the 2 data sets, an AD model was trained first. Using a validation set, a trained model was tested. Afterwards, the trained models were evaluated on different test data. Additionally, two baseline models (see section 6.1.2 and 6.1.3) were also compared to rate the performance. For one data set this concludes in $(3 * (2 * (2 + 1)) + (1 * (2 * 2))) * 5 = 110^1$ measurements. For every measurement, established metrics for the evaluation of OCC problems (see section 2.5.4) were used. Secondly, the time a model took to fit (T2F) or evaluate (T2T) a given data split was also measured. Characteristics such as number of frames per data set can be obtained in chapter 4.
Although more data was available, this initial research only used parts of the data to answer the proposed research questions (compare section 1.1). As deep learning models tend to increase their performance with more data, shallow methods like the ones introduced in section 2.5 are not as likely to increase their performance.

---

[1] $(\#AD\ methods * (\#fragment\ types * (\#pcapAE\ models + raw\ baseline)) + naive\ baseline) * \#data\ split$

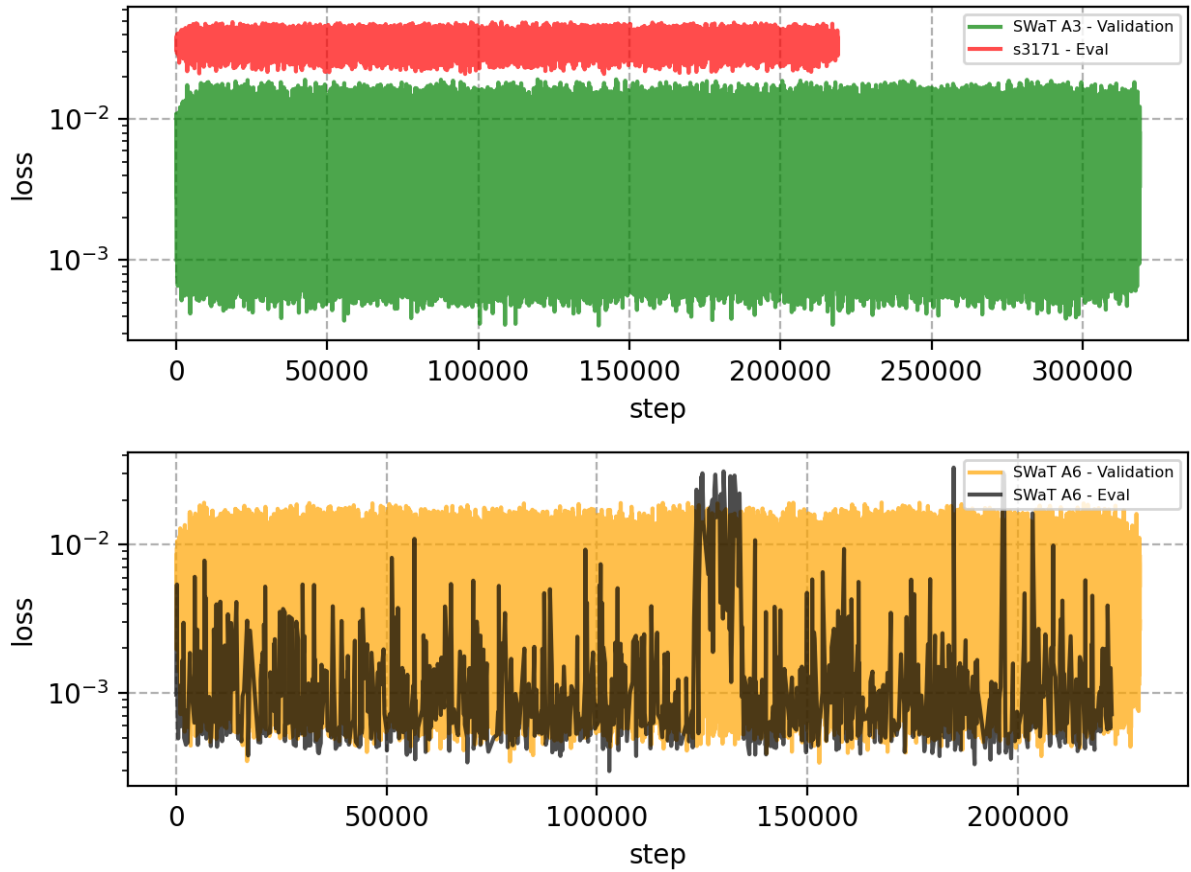

Figure 6.1: Residual loss for different iTrust data sets

### 6.1.1 Autoencoder Setup

A descriptive overview about the specific autoencoder network arrangement can be found in the appendix under section C.2.[2] *RelU* was chosen as a non-linear activation function for the ANN. The network used the mean square error ($MSE$) to calculate a distance between ground truth and output sample and was trained in a mini-batch approach with a batch size of 2. This small batch size was chosen since other models like [89] note that larger batch sizes can lead to disadvantages when *small* attack traces are analysed. Lastly, the learning rate was dynamically adjusted after each epoch using the method described in [37]. After observing many training experiments, the initial learn rate was set to 0.00002 as it produced stable loss curves. As stated in section 2.4.1, larger learning rates may affect how well a model is able to generalize. Pre-training of individual layers was not utilized, as the network only used four layers per subnetwork [143].

---

[2]An interactive model overview can be seen under: tensorboard.dev/experiment/OTD0qLBeQYK1DajlvkwoyQ/#graphs

Using a hold-out data set with $100,000$ samples, different training and model-related hyperparameters like: the type of RNN cell (LSTM or GRU), the used loss function (*MSE* and *BCE*[3]), the different gradient descent strategies (SGD[4] and adamW [34]), and dynamic learn-rate schedulers (*cycle* [37], *plateau* [145], and *step* [146]) were compared in a grid search, and summarized in table 6.1. Based on this parameter search and prior knowledge about the training of RNN models, a GRU RNN cell was chosen in the pcapAE framework. Secondly, the model was trained using the MSE loss function, since over all runs this criterion produced models with lower loss values. Using the MSE, loss values range from a maximum error of 1 and a minimal value of 0. Weights were updated using a cyclic learn-rate scheduler using a weighted gradient descent strategy [35] (c.f. section 2.4.1), because the loss curves depicted a gradual decrease compared to other strategies.

| RNN-Cell | Loss Function | Optimization | LR-Scheduler | Loss after Epoch 5 |
|----------|---------------|--------------|--------------|--------------------|
| GRU | MSE | SGD | cycle | $2.82 \cdot 10^{-4}$ |
| GRU | MSE | SGD | plateau | $1.72 \cdot 10^{-4}$ |
| GRU | MSE | SGD | step | $1.72 \cdot 10^{-4}$ |
| GRU | MSE | adamW | cycle | $\underline{2.34 \cdot 10^{-5}}$ |
| GRU | MSE | adamW | step | $6.11 \cdot 10^{-5}$ |
| GRU | MSE | adamW | plateau | $1.26 \cdot 10^{-4}$ |
| GRU | BCE | SGD | cycle | $1.14 \cdot 10^{-3}$ |
| GRU | BCE | SGD | step | $2.93 \cdot 10^{-4}$ |
| GRU | BCE | SGD | plateau | $6.34 \cdot 10^{-4}$ |
| GRU | BCE | adamW | cycle | $3.15 \cdot 10^{-5}$ |
| GRU | BCE | adamW | step | $6.92 \cdot 10^{-5}$ |
| GRU | BCE | adamW | plateau | $1.71 \cdot 10^{-4}$ |
| LSTM | MSE | SGD | cycle | $9.49 \cdot 10^{-5}$ |
| LSTM | MSE | SGD | step | $5.81 \cdot 10^{-5}$ |
| LSTM | MSE | SGD | plateau | $5.81 \cdot 10^{-5}$ |
| LSTM | MSE | adamW | cycle | $\underline{2.21 \cdot 10^{-5}}$ |
| LSTM | MSE | adamW | step | $2.84 \cdot 10^{-4}$ |
| LSTM | MSE | adamW | plateau | $7.16 \cdot 10^{-4}$ |
| LSTM | BCE | SGD | cycle | $9.75 \cdot 10^{-5}$ |
| LSTM | BCE | SGD | step | $6.52 \cdot 10^{-5}$ |
| LSTM | BCE | SGD | plateau | $6.33 \cdot 10^{-5}$ |
| LSTM | BCE | adamW | cycle | $2.39 \cdot 10^{-5}$ |
| LSTM | BCE | adamW | step | $1.15 \cdot 10^{-3}$ |
| LSTM | BCE | adamW | plateau | $9.35 \cdot 10^{-4}$ |

Table 6.1: Hyperparameter grid search results

---

[3]Binary Cross-Entropy $(X, \hat{X}) = -\frac{1}{n} \sum_{i=1} (x_i * log(\hat{x_i}) + (1 - x_i) * log(1 - \hat{x_i}))$.
[4]Vanilla Stochastic gradient descent [144].

### 6.1.2 Raw Baseline ($B_r$)

To gauge the proposed framework's added value regarding the detection of anomalies using AE-compressed fragment information, a so-called *raw* baseline is established by evaluating an anomaly detection method on uncompressed fragments. Thus, the raw baseline ($B_r$) is not trained on extracted features, it only uses the established fragments in this setting. Input samples are therefore a flat representation of fragments ($1,024$ values per sample) and are normalized into the interval of $[0, 1]$. If the detection accuracy using one of the considered AD algorithms in conjunction with the pcapAE model is lower than the $B_r$ this implies, that the feature extraction has no clear benefit.

### 6.1.3 Naive Baseline ($B_n$)

In order to compare the shallow anomaly detection schemes that work on top of the proposed pcapAE method, a naive baseline ($B_n$) was additionally evaluated. This naive approach for AD only uses the proposed representation learning to detect anomalies. Similarly to other methods [85, 89, 95, 119], the model's internal loss values are exploited to assign an anomaly score for each sample. Through the training process, this residual loss value is minimized over time. Since anomalies are outside the learned norm, one expects the model to have a higher loss value on those samples.

A higher loss for a sample would also be reflected within the *code* the decoder network reconstructed the input sample from. Thus, any downstream anomaly detection is in theory also capable of detecting these deviations.

To turn the residual error term into a decision boundary for anomaly detection, for a trained model the average model loss (AML) and its standard deviation ($\sigma$) is calculated using a validation set. Using these two values an upper-bound threshold (thr) can be defined as $thr := AML + \nu\sigma$ where $\nu \in \mathbb{R}$ dictates how much slack is allowed in the detection of anomalies. $\nu$ is similar to the trade-off parameter used in OCSVM that determines the upper-bound of the fraction of training errors the model is allowed to make.

If for some sample the calculated loss $\epsilon > thr$, then the sample is classified as abnormal. Given the assumption that loss follows a normal distribution and using the value 2.5 for $\nu$, one can expect that 98.9% of examples are classified as normal for training data. This allows for a more general decision boundary, for the naive autoencoder-based anomaly detection model.

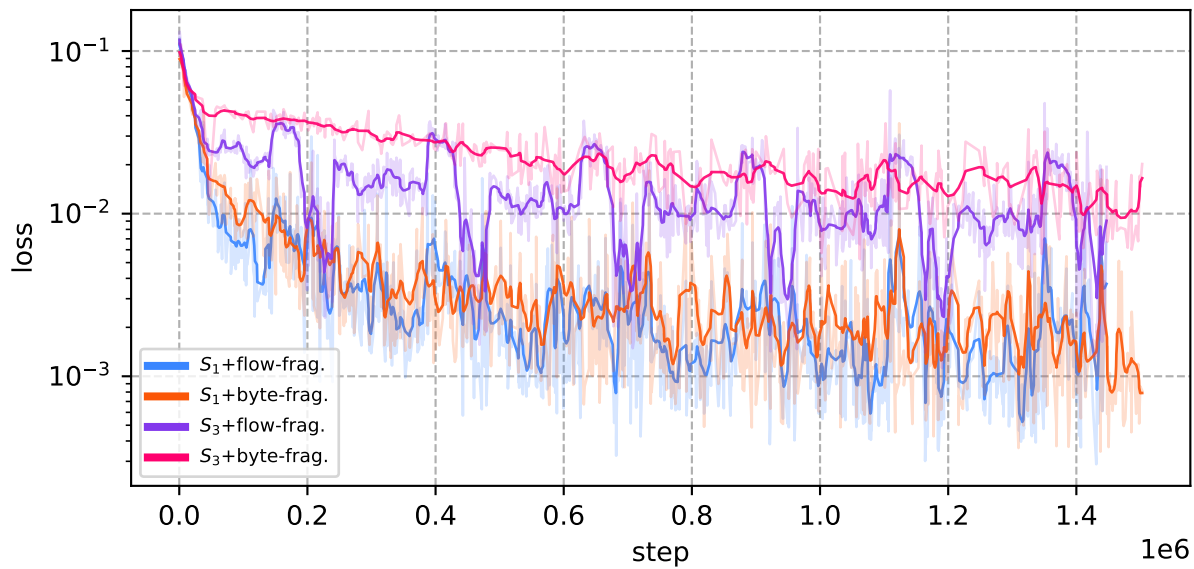

Figure 6.2: Training loss-curves for the 4 different trained models for the SWaT A6 data set

## 6.2 Model Analysis

A model analysis was conducted to get a quantitative measure of each model's performance. For the actual representation learning only the model loss was examined, while the binary confusion matrix (c.f. section 2.12) and its metrics were used to evaluate the downstream anomaly detection.

### 6.2.1 Representation Learning

Figure 6.2 and figure 6.4 show the residual loss (MSE) over time for each data set. For all pcapAE model trained for 6 epochs. The training was stopped in order to prevent the models to over-fit on the training data set. All models shared the same training hyperparameters such as initial learning rate, weight optimization strategy and others as described in section 6.1.1. A conceptual overview of the used pcapAE PyTorch model is depicted in appendix under section C.2.

The learning progress is visible in the decreasing loss value. Interestingly, for both data sets the flow-based preprocessing with an input sequence length of $n = 3$ showed visible periodic patterns within their loss values. This indicates that these models are not fitted as well to the data as their $S_1$-based counterparts. The plots also show that for the $S_1$-based models the loss is significantly smaller, compared to the $S_3$-models. This indicates that longer sequences are harder to learn.

By comparing the distributions of loss values per model and data set (compare figure 6.3 and figure 6.5), one can also see the similarities as in both cases the $S_3$-based models are shifted father to the right. As these models on average produce greater loss values, one can expect that these models are worse in reconstruction the input data, and thus intern are inferior for the extraction of features. To explain the phenomena across the different data sets, one has to keep in mind, that learning longer input sequences of fragments is harder to accomplish for a model. Although the model has more information, which can be used to reconstruct the last fragment, the recurrent signal form the hidden variables, make the task more difficult.

Figure 6.6 visualizes flow-oriented fragments that are compressed by a model trained on sequences with length $n = 1$. One can see that input fragments depict an internal structure which are dictated the structure of the associated IP payload's, which containing the first 64 bytes of each frame (compare section 5.2.1).

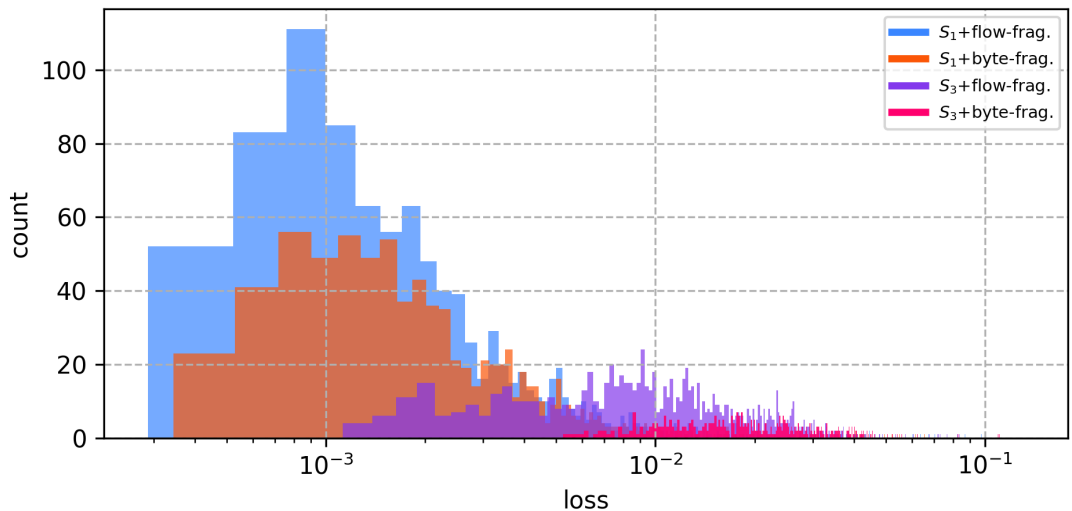

Figure 6.3: Training loss distribution for the SWaT A6 data set

The same can be said for the latent variable (code) retrieved from the pcapAE network. Since extracted features are not descriptive, one can not tell what the patterns represent in detail. For changing input samples, the code also changes in certain areas.

The evaluation shows that the model is capable of learning structural aspects of the fragments as the black *gird* lines are reproduced. On close inspection one can see that some areas are not as *bright* as in the original fragment. This reflects the lossy characteristic of autoencoders. Since the model inspection was conducted at the end of the model training, the reconstructed fragment is near identical copy of the input sample.

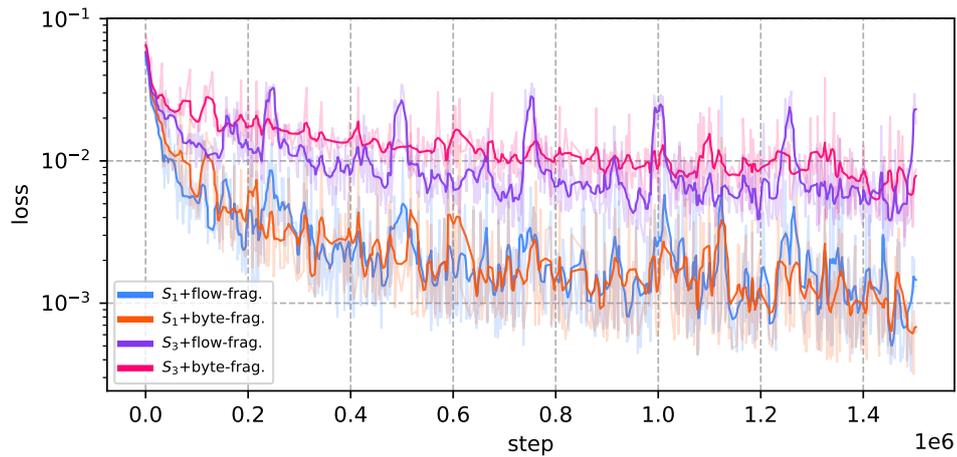

Figure 6.4: Loss curves for the different trained models for the Voerde data set

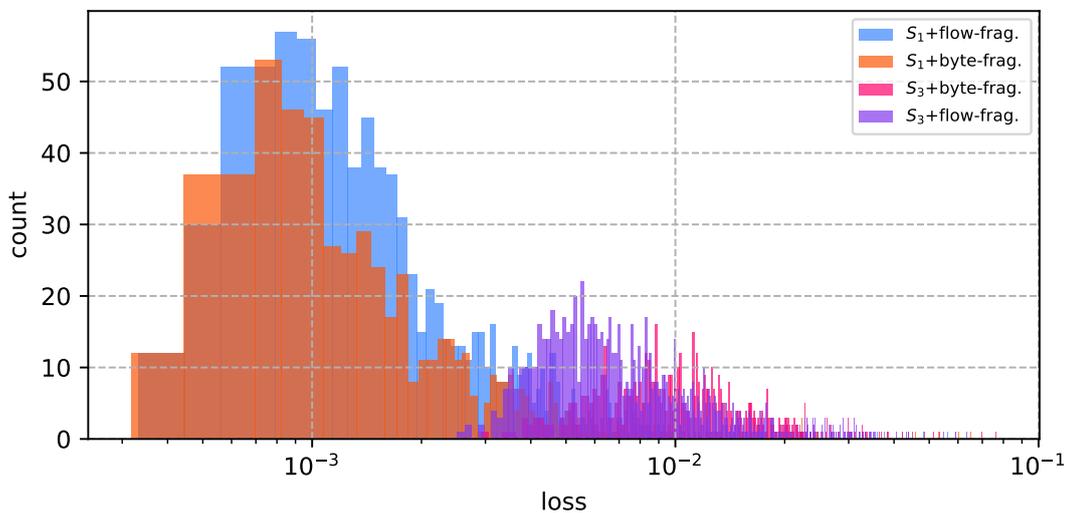

Figure 6.5: Training loss distribution for the Voerde data set

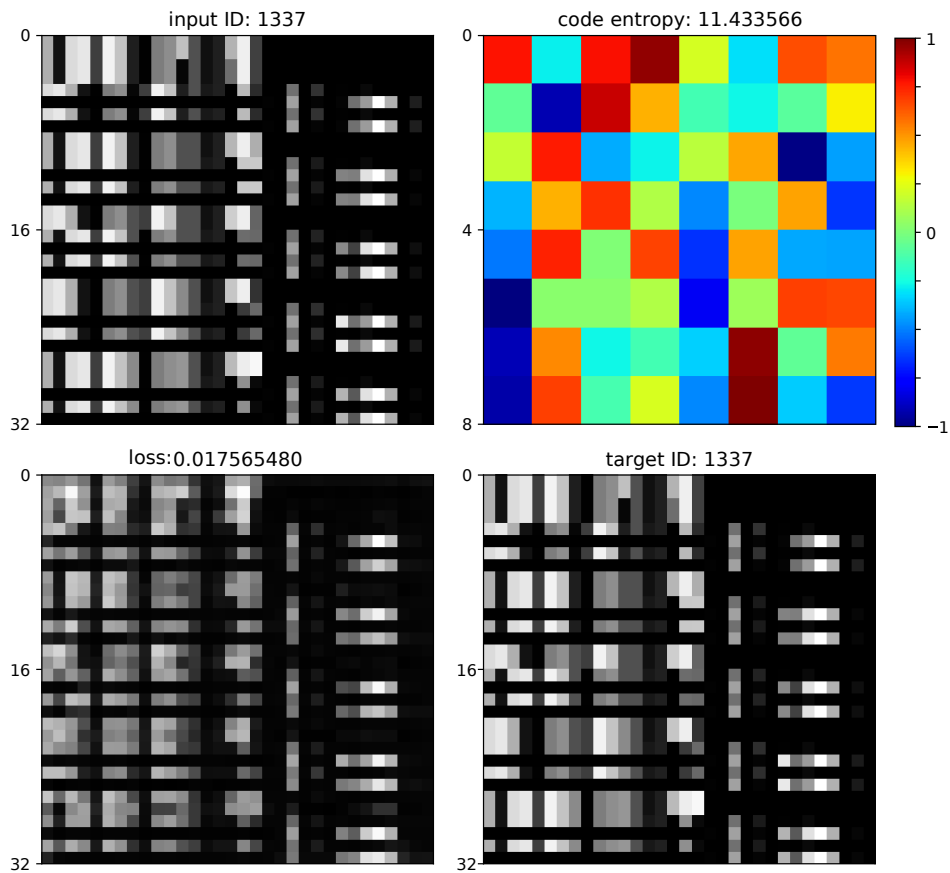

Figure 6.6: Input fragment (top left), extracted features (top right), reconstructed fragment (lower left), target fragment (lower right)

## 6.2.2 Anomaly Detection

In order to conduct a quantitative analysis of the performance of an anomaly detection method, one needs ground-truth information. This fact was already established in section 2.5.4. Since the representation learning is ultimately used to identify anomalous frame sequences, a network trace with frame-specific labels is needed for the ground truth. Figure 6.7 shows how sequences of byte-fragments (compare figure 5.1) are compressed by the encoder network using a sequence length of $n = 3$. The evaluation was conducted in a context where codes were counted as anomalous, if one of the associated input frames was labelled as an anomaly. This setting thus marked $code_1$-$code_3$ as anomalous since fragment $\beta_3$ was part of the input of each code and $\beta_3$ contains bytes from a known abnormal frame. The codes contain as stated, compressed representations of raw network data, these features were therefore used to train and test a downstream anomaly detection algorithm. Figure 6.7 also shows how the evaluation metrics are calculated using the ground-truth information. If the AD model alerted an anomaly, an auxiliary data structure was queried to resolve codes back to frame identifiers. These identifiers, along with an AD specific anomaly score and relevance heatmaps (see section 5.2.3) are the contents of raised alerts.

For the different data splits that were introduced in section 4.5, table B.2 shows how many codes were extracted on each setting and how long the extraction took. Since flow-fragments always use byte information of 16 Ethernet frames, the subsequent methods had fewer examples compared to the byte-fragment setting. Extracted codes from sequences of fragments are marked as abnormal even if only one of the contained frames is labelled as anomalous. This discrete approach has the effect that the $S_3$-related data sets have a higher percentage of anomalous samples.

The Voerde $Eval_H$ split does contain more samples which are labelled as abnormal since every frame issued from the MAC address of the penetration tester are recognised as malicious. The frequency in which this malicious network interface issued frames into the network was so high that nearly every fifth frame was sent by it. This resulted in the large number of anomalous samples.

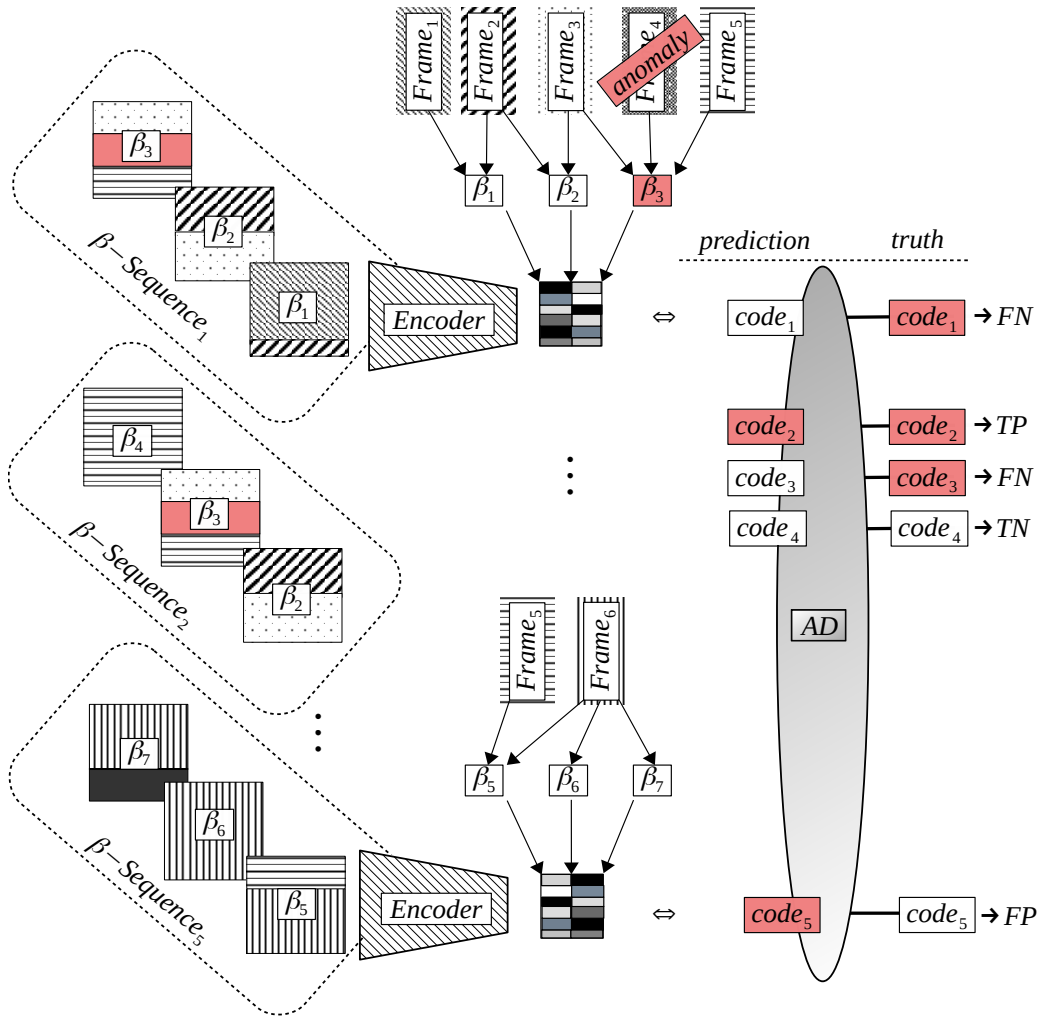

Figure 6.7: From 6 Ethernet frames, 7 byte-fragments are constructed and used to extract features with an input sequence length of n=3 resulting in 5 codes.

### 6.2.3 Discussion

The experiment result tables (B.3 to B.12) in the second appendix B contain metrics about each experiment trail. For each data set and for every combination of the investigated algorithms and type of data representation, precision (PR), recall (RC), false positive rate (FPR), and the $F_1$-score were measured. The time an AD model took to fit to the *Train* data is noted as *time to fit* (T2F). Likewise, the inference time for an AD model is noted as *time to test* (T2T). For each of the two data sets, after the initial training the models were tested using the *Validation* data split (compare table B.3 and table B.8). These traces of course do not contain any anomalies, since they are designed to test if the model over-fitted to its training data. Precision and recall are not meaningful here, since both metrics depend on the measure of true positives. TPs are not present for normal training data, which is assumed to be negative-only data. The false positive rate in this case is the only valuable metric. For the SWaT data set both *IF* and *LOF* showed a perfect FPR, while the *OCSVM* models had more difficulties on the *Validation* set. The results for the Voerde-based experiments express similar characteristics, although the nominal values are even lower.

Notably, the raw baseline also indicated a poor fitting to the data, while also taking much longer to train. The raw baseline measurements show that using compressed features, any AD algorithm converges significantly faster. As both measurements are not acceptable, one can not judge if the feature extraction framework impacted the AD performance negatively.

When the contaminated data splits ($Eval_H$, $Eval_D$, $Eval_E$, $Eval_R$) are evaluated, one can see that a given AD models is not capable to produce a high $F_1$-score combined with a low $FPR$-rate. Only the naive baseline model was able to produce an acceptable detection results in the case of the $Eval_H$ data set. For the naive model, one can see that in settings were byte-fragments are used, better results are achieved. For the synthetic data sets, virtually none of the models detected any anomalies.

This leads to the conclusion that the extraction and respectively the learning of feature from raw network data was in part successful. This is also supported by the analysis in section 6.2.1 and is reflected in figure 5.5, as the order of fragments influences the reconstruction error. The first research question (see section 1.1), can therefore be approved. Since mostly only intra-packet anomalies (c.f. section 2.1.3) were successfully identified and due to the fact that $S_3$-based models have higher loss curves, one can also see that the temporal extraction of features was not as successful as the spatial extraction.

For the Voerde-based evaluation, the $Eval_H$ split containing mostly assumed intra-packet anomalies, was tested first. Table B.9 suggests that the isolation forest-based model is best suited to detect these anomalies since it achieved an $F_1$-score of *0.9334*. But since the model flagged every packet as anomalous, the false positive rate is therefore unacceptable (see section 2.2.2). The naive baseline in this setting produced an $F_1$-score of only *0.8452*. Although this indicates that the model is worse, the FPR in this case is much lower (*0.0419*). This clearly indicates an advantage of the naive baseline model compared to the shallow AD methods. The raw baseline, over all experiment trails, shows that the feature extraction helps the AD methods to converge faster but does not raise its detection rate. Having this inside, one can conclude that the second research question is denied through the evaluation.

One reason for the rather bad detection result can be explained by examining the AE based feature extraction. For parts of the traffic, the model is able to produce a residual loss value which is comparable to other autoencoder-based methods like [85, 89, 95, 119]. In case of the SWaT-based evaluation, the MSE loss value periodicity raises above 0.1 for strictly normal samples. Packets that produce higher values, were identified as large number of TCP reset events, which occur when an unexpected frame is sent to a host. As the model is under-fitted for these events, the resulting feature extraction build unsuitable features which affect the following AD. An extended training of the pcapAE model on TCP reset examples could be used to teach the model of the specific data characteristic.

Overall one can see that models that were trained on byte-oriented fragments accomplish better result when for example compared on the $Eval_H$ sub set for both trails, hinting that this heuristic might be better suited as an input representation and answers the third research question. In both cases the $Eval_H$ set contained mostly intra-packet anomalies. The result is not unsurprisingly as flow-oriented fragments only use a fraction of the application layer and thus have limited access to information that might be useful in the feature extraction process. Secondly, models had limited access to flow-fragments as more data was discarded during the creation process resulting on fewer samples when compared to byte-fragments. For inter-packet anomalies like DoS attacks no AD method reported any usable results. This leads to the conclusion that the evaluation can only make qualified statements about intra-packet anomalies.

Since the proposed pcapAE framework failed to achieve similar results for AD compared to other publications, one has to reflect on its design decisions that influenced the outcome. Foremost one has to acknowledge that the majority of the presented related work used supervised classification instead of a fully unsupervised approach. Secondly, the highlighted works mostly focused on end-to-end methods that did not rely on shallow anomaly detection, resulting in a slimmer architecture.

For the proposed pcapAE framework one can question the decision that no packet reassembling was conducted on the input traffic, which resulted more noise the framework was exposed to. This noisy input data might influence the feature extraction and ultimately contribute to the bad detection results. A second design decision was the usage of two-dimensional convolutional networks to be used to extract features on one-dimensional input sequences. Since fragments were arranged into two-dimensional representation, an artificial correlation between rows of bytes was imposed. Another explanation for the evaluated results is that some of the pcapAE models ($S_3$) also indicated that their training was stopped prematurely, as their loss curves did not fully plateau. This resulted in an unfavourable feature extraction. Lastly one can also hold the hyperparameter grid search accountable, since it only searched the parameter space for byte-related fragments trained with an input sequence length of $n = 1$.

### Naive Baseline

As stated earlier, a naive baseline was constructed to test the pcapAE's capabilities to detect anomalies. This was done following the paradigm that one expects a higher residual error stemming from a vastly different *code* for a fragment containing malicious bytes. Likewise, benign samples should produce a low residual loss, since the model was trained to reconstruct them.

Figure 6.8 depicts the model loss for the different subsets of data. In this experiment trail of the SWaT-based evaluation, the model was trained on byte-oriented fragments with an input sequence length of $n = 1$. After the blue $Training_{AD}$ subset was processed, the naive baseline computed an upper threshold value of $0.01 = AML + 2.5\sigma$ (c.f. section 6.1.3). It is clear that the training loss is not static, as periodic spikes raise the loss to around 0.12. The model seems to be *under-fitted* for those kinds of events. A second subset ($Validation_{AD}$) was used to test the developed threshold value. Lastly, using actual testing data ($Eval_H$) containing roughly labelled anomalies (c.f. section 4.3) marked with red, the naive baseline measurement could be evaluated. Further, plots for the SWaT-based evaluation as well as for the Voerde $Eval_D$ data sets are listed in the appendix under section B.3. The four figures indicate that models trained with an input sequence length of $n = 1$ are seemingly better suited for byte-fragments, while the opposite is true for flow-fragments. The autoencoders that trained with longer sequences also show a higher model loss, indicating that these models might have stopped the training procedure too early.

This particular trace contained a malicious file download event, which was successfully detected by the model, as the fragments around step 1.2 million have unusual loss values and proving a successful learning of features. As the metrics show, only half of the selected anomalies were true positives. This again is a result of the seemingly *under-fitted* model. A second explanation is that the estimated labels are not close enough to the real distribution of anomalies.

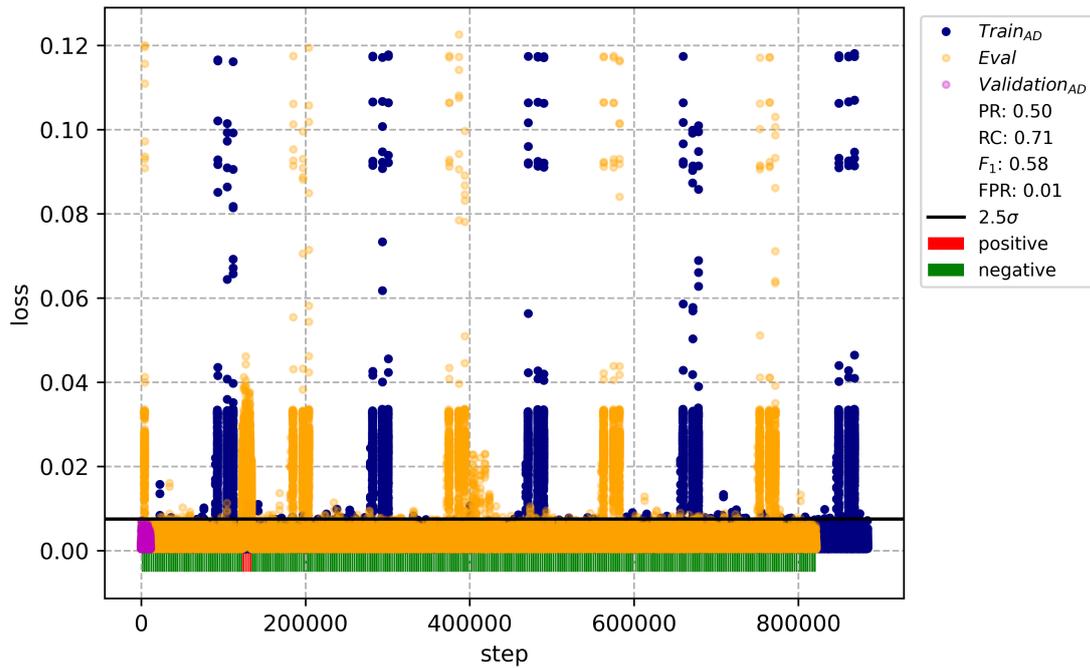

Figure 6.8: SWaT A6 *Eval_H* naive baseline evaluation loss visualisation for byte-fragments trained on an input sequence length of n=1

A second cluster of irregularities can also be observed in figure 6.8. Around step 4.2 million, a cloud of high-loss values appears. These events were not marked as anomalous. A manual investigation of packets that were used to produce these high residual loss values was conducted. This analysis revealed some *strange* user datagram protocol (UDP) packets that are likely to be a side effect of the conducted attack on the SWaT data (c.f. section 4.1). An example of such an observed UDP packet is highlighted in figure 6.9.

| | Time | Source | Destination | Protocol | Leng Info |
|---|---|---|---|---|---|
| 844... | 3.420098 | 192.168.1.183 | 118.201.255.203 | UDP | 802 54028 → 1195 Len=760 |
| 84498 | 3.420099 | 192.168.1.183 | 118.201.255.203 | UDP | 806 54028 → 1195 Len=760 |
| 84499 | 3.420186 | 192.168.1.183 | 239.255.255.250 | UDP | 698 54029 → 3702 Len=656 |

Frame 84497: 802 bytes on wire (6416 bits), 802 bytes captured (6416 bits) on interface 0
Ethernet II, Src: Giga-Byt_6c:68:78 (e0:d5:5e:6c:68:78), Dst: AdiEngin_0b:0d:aa (00:08:a2:0b:0d:aa)
Internet Protocol Version 4, Src: 192.168.1.183, Dst: 118.201.255.203
User Datagram Protocol, Src Port: 54028, Dst Port: 1195
Data (760 bytes)

```
000   00 08 a2 0b 0d aa e0 d5   5e 6c 68 78 08 00 45 00   ........^lhx..E.
010   03 14 b6 22 00 00 80 11   48 c2 c0 a8 01 b7 76 c9   ..."....H.....v.
020   ff cb d3 0c 04 ab 03 00   f9 45 4b 00 00 00 85 6d   .........EK....m
030   ee ea cd 51 7f e7 f1 2a   86 69 e9 d4 29 3e 27 01   ...Q...* .i..)>'.
040   5c 72 9d 29 d4 03 8f 80   75 8e b4 d6 e6 e7 ef bf   \r.).... u.......
050   7a be 39 9d 9c ed be 3a   f3 6c f7 f3 8d 6f 6c af   z.9....: .l...ol.
060   76 35 80 5b 08 e6 50 81   3c cd e8 5e de a6 c6 58   v5.[..P. <..^...X
070   c1 ca f6 dc cf 21 07 95   7a f0 9b e3 ff 9a 06 37   .....!.. z......7
080   23 13 3d 9e 66 7b ea b0   d5 11 60 ab a3 55 27 f4   #.=.f{.. ..`..U'.
090   7b a8 b8 88 ce 28 25 9e   db 04 cc ed b8 d8 54 23   {....(%. ......T#
0a0   b4 50 06 76 05 03 71 b4   24 25 ee 1f 92 bf e4 21   .P.v..q. $%.....!
```

Figure 6.9: *Wireshark* excerpt from a UDP packet with an unusual payload

## 6.3 Decision Analysis

Lastly, this thesis has conducted a decision analysis (see chapter 2.6) for the proposed framework. Deep-learning- based methods are often not easy to be interpreted, since they represent their learned knowledge with hierarchically structured weights and biases. This also effects the extracted features by the pcapAE framework. The 64 values that represent a certain sequence of fragments do follow some internal structure, as one can see from figure 6.6. A model however does not express what certain areas within the code truly represent.

In understanding what parts of an input fragment contributed the most to the final code, one can get some sense on what feature the model was particular interested on extracting. Therefore, the method of layer-wise relevance propagation (c.f. section 2.6.2) was used to generate heatmaps of input signal relevance.

Figure 6.10 shows a generated heatmap for an abnormal fragment belonging to the SWaT *Eval_H* set. The relevance for the sample is clustered in the upper area, indicating a focused attention of the pcapAE feature extraction method. Revealing the correspondent bytes within the fragment does not indicate why the fragment is anomalous as the data is a binary blob. Further protocol parsing would be necessary to help to interpret the highlighted areas.

Figure B.9 shows the decision analysis for the examined *false* FP frame highlighted in figure 6.9. As the binary UDP protocol is not easily readable, the highlighted area also needs further protocol-specific aggregation in order to be truly helpful. For unencrypted and public implementations, protocol dissectors could be used to aid the investigation for example.

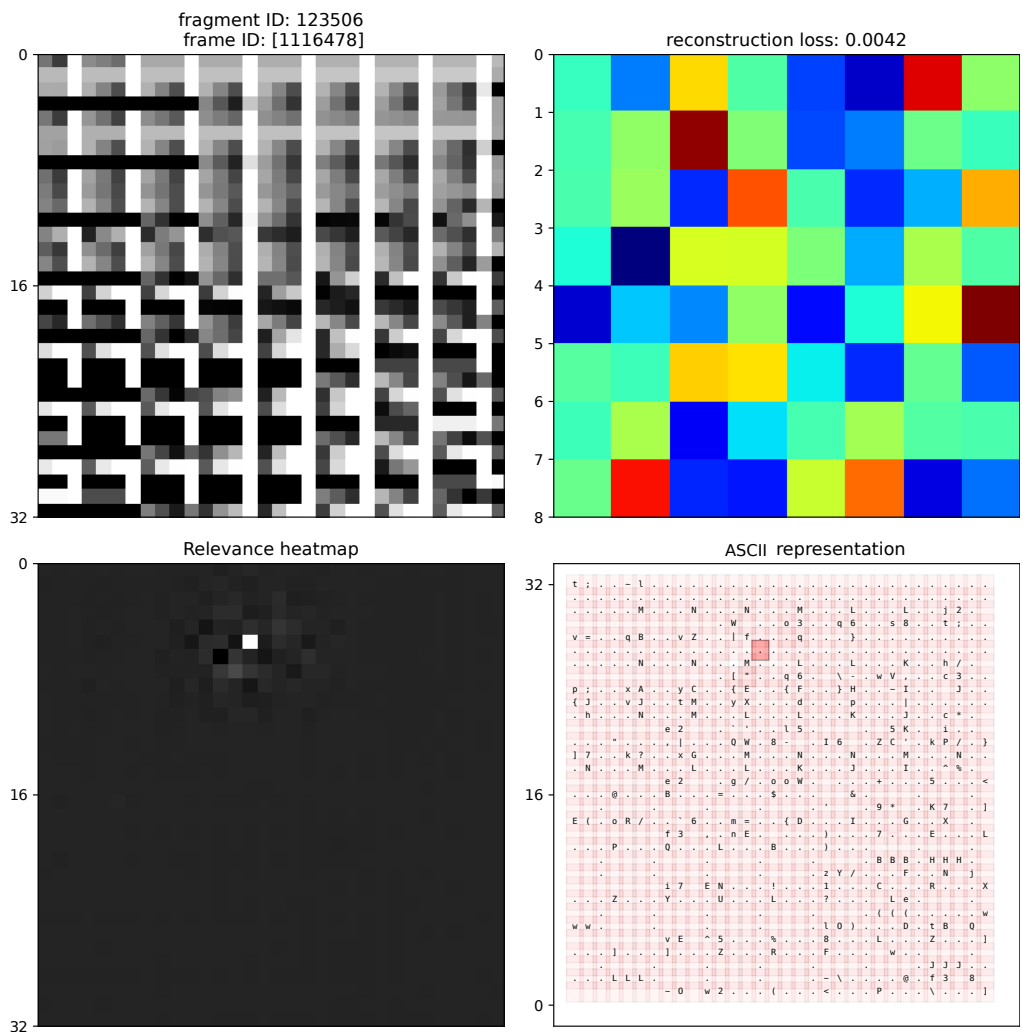

Figure 6.10: Byte-fragments relevance heatmaps of the SWaT file download event

When heatmaps for flow-fragments are analysed, one can see that the autoencoder is always interested on the right side of a fragment (see figure 6.11 and further examples in the appendix under section B.4). Since for flow-fragments the transport layer payload is written from left to right for each frame, mostly application layer data is contained in every odd row of a fragment. Surprisingly, for byte-fragments the LRP method also showed that the right side of a fragment is often more relevant. Since byte-fragments have no clear internal structure, this result was not expected.

Through the usage of LRP, the decision analysis discovered what part of the input payload data was more relevant for the used pcapAE model.

For the above-mentioned samples, it is clear that generated heatmap could be used to enrich generated alerts by any anomaly detection algorithm. Operators have a chance to intuitively know what part of the fragment were more relevant in generating a code for the model. Since these extracted features are the basis for any AD technique, analysing the relevance heatmaps can help to bridge the semantic gap for a number of AD methods as long as the underlying network protocols can be parsed accordingly. The decision analysis therefore also helps to answer the last research question.

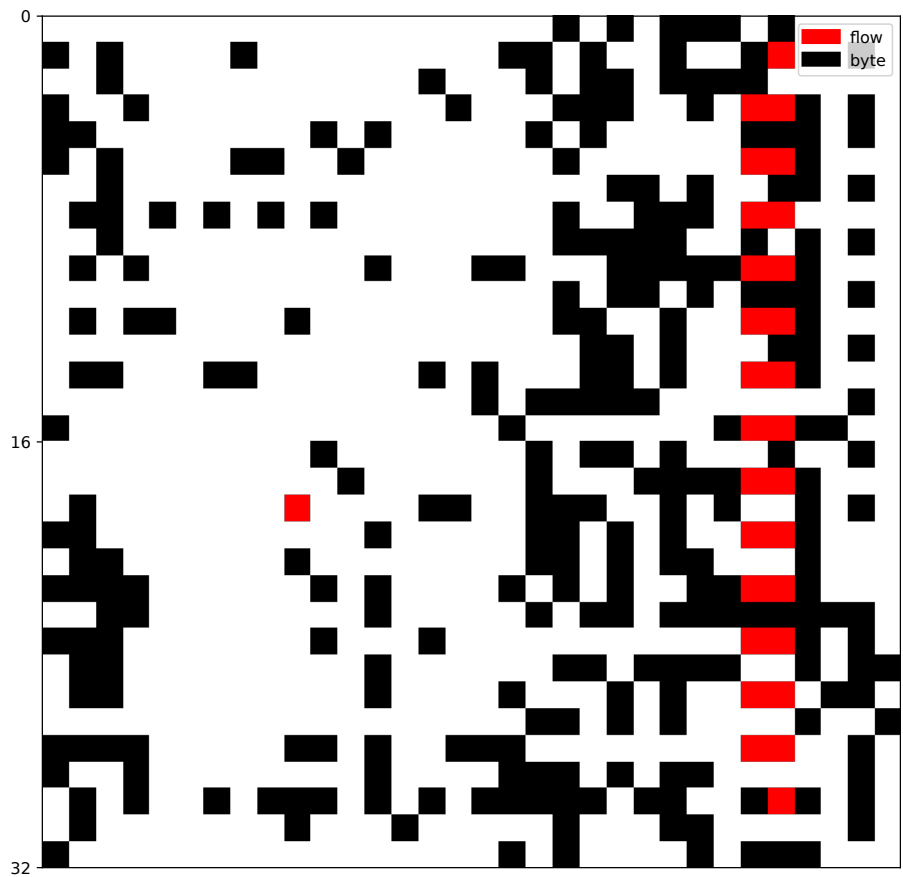

Figure 6.11: Aggregated relevance heatmaps for 1% of randomly selected evaluation fragments

# 7 Conclusion & Future Work

This chapter concludes the presented thesis. Achievements and problems are addressed in the following section. Proposals for future work, indicated by the conducted research, is lastly suggested.

In summary, this thesis dealt with the problem of extracting features for a content-sensitive AD, and to also use explainability approaches to analyse extracted features. Since network protocols can follow diverse specifications, an unsupervised and automatic method was developed to extract features from raw network traces. Two minimal pre-processing steps were evaluated for this matter. Using the proposed method, learned representations were afterwards applied for existing methods of anomaly detection and baseline models. With help of a used explainability heuristic, the framework was able to visualize data-driven decisions.

## 7.1 Experiment Conclusion

The success of an IDS depends on the data the system has access to, where it is placed within the network's topology and how good the detection heuristic is. If for example an attacker has already managed to infiltrate lower levels of an CPS-infrastructure and does not try to move laterally, it is hard to detect from a high-level network perspective. However, since the attacker may try to change certain parameters, the underling fingerprint of the behaviour of the cyber physical process can be used to detect these kinds of attacks (see section 2.1.1). On the other hand, certain attacks that do not alter the behaviour itself and instead network resources are used to move, extract or upload data. Even though this does not directly impact the process, it constitutes unwanted activity. These kinds of attacks could be detected by considering communication graphs. For the conducted experiments, anomalies that affected the cyber physical process as well as usual communication patterns were both investigated.

Through the analysis of the representation learning model, research question I could be positively answered. The different models showed a capability of extraction relevant information from a sequence of fragments. Since extracted feature that contain know anomalies, were poorly reconstructed, one can conclude the model has learned how benign fragments are structured. The distribution of the reconstruction loss also indicated, that anomalous samples not always produced significant higher loss values, which can be explained with an under-fitting in the feature extraction model. A longer training of the model with a larger validation set should be used for future evaluations.

Reflecting posed research question II, it is clear that the pcapAE framework failed to produce acceptable results when used in conjunction with classic anomaly detection methods. The AD models failed to detect anomalies based on the extracted features. One reason for the low $F_1$-scores could be explained with an over-fitting to the grid search data set which resulted in unfavourable hyperparameters. Secondly, the AE may also compress the input samples in such a way, that attack-related traffic features are encoded adversely for the subsequent anomaly detection methods. Lastly, as the pcapAE feature extraction model seems to be partly under-fitted, the extracted features are influenced by the data-driven model and results in bad detection results.

Comparably, one can see that the naive baseline heuristic for the task of anomaly detection is partially capable to alert intra-packet anomalies in byte-oriented settings for the same amount of training examples. Which indicates that the pcapAE framework in this case was able to learn content-sensitive (spatial) features of the network traffic.

The difference in the detection result might be explained with the fact that the naive heuristic is an end-to-end model and is thus less prone to specific limitation and hyperparameters imposed by an individual AD algorithm. The naive AD model can directly infer on the actual extracted features, without the need to train a second type of model.

Research question III investigated what kind of minimal preprocessed representations would yield better results. Through different experiments, a slight advantage for models that trained on byte-oriented fragments of Ethernet frames can be attributed. This result is biased because less flow fragments were extracted in the training process, resulting in an unbalanced distribution. Secondly, the type and amount of anomalies that were present in the data sets could also be more favorable for byte fragments. Future evaluations should try to balance the amount of input examples for each type and also use cross validation to produce results that are less dependent by the chosen data split. Using different randomly selected validation sets to test a model against and also randomize the training data split itself has the effect of a higher computational effort, but also produces results that are more comparable.

Lastly, research question IV investigated how extracted features could be explained in a transparent manner. Therefore, heatmaps of input feature relevance are produced for each generated alert by any anomaly detection method. This could be achieved using techniques that reverse the signal propagation within an ANN. The produced heatmaps highlight areas the feature extraction framework focused on when fragments were compressed. And for example revealed what part of flow fragments were more influential in the feature extraction process. Since the highlighted areas could trivially be converted back into the initial raw hexadecimal byte values, these maps can therefore help to bridge the semantic gap for the problem of anomaly detection, as long as the underling network packet can be dissected for human operators.

The evaluation analysed the representation learning framework as well as the subsequent anomaly detection. The experiments showed that the pcapAE model was able to learn aspects of the underlying raw network traffic. Unfortunately, the intrusion detection was in general not able to identify anomalies on an acceptable scale. As the representation learning was presumably stop prematurely, the resulting features are suboptimal. Secondly, the AD methods used hyperparameters that were chosen unfavourable. Future evaluation should try to address the problems by using larger validation data sets, deploy a cross validation strategy and also expend the hyperparameter grid search. As the evaluation data sets only contained roughly labelled anomalies or used synthetically-generated anomalies, future efforts should also try to focus on providing data sets were intra- and inter-packet anomalies can be found with concert labels.

The representation learning framework was also analysed from a second point of view. The decision analysis showed that the LRP method is a valuable tool for the domain of anomaly detection. With generated heatmaps, one has the opportunity to comprehend made decisions by the deep neural network. Since the pcapAE framework, the anomaly detection implementations, and the SWaT data set are public accessible, current results are reproducible and could be improved further.

## 7.2 Recommendations for Future Work

The pcapAE framework showed a potential approach for subsequent works. Mainly the initial research efforts for a protocol-agnostic and content-sensitive feature extraction using spatial-temporal attributes is valuable concept for future AD methods. Deviations of the *pcapAE* framework can implement an essential services necessary for the domain of network-based intrusion detection, as raw network data is processed in such a manner that is can be used for intrusion detection.

Additionally, the thesis presented a novel strategy for the field of industrial networks in intrusion detection by investigating existing methods that try to explain data-driven decisions of an ANN. By generating relevance heatmaps, one can inspect the decision-making process of the framework which also helps to close the semantic gap (see demand 2 from section 2.2.2) between alert and actionable result. Future implementations should try to also highlight areas with in the input signal, that give evidence why a sample is normal.

As pointed out in the earlier evaluation discussion in section 6.2.3, several aspects can be addressed. For future investigations in the realm of convLSTM-based autoencoders, one should be sure that the training process is not halted prematurely. By examining the loss distribution of a larger validation set, one can better test the model. Another factor one should correct for, is a balanced distribution of different types of input fragments. As fewer flow-fragments were present, the evaluation was not fair in this regard.

To aid the AD models, one could restrict the number of application layer protocols presented to the feature learning framework. This would impose a limitation on the detection results, but would help the framework to train, as the input signal is less diverse, which could result in extracted features that are *easier* to learn by the AD methods.

Likewise, one could train an ensemble of models for each networking device in a fixed network topology. With the idea, that each feature extraction model is tuned to a specific MAC address resulting in a less noisy environment. Since for the two approach the input signal is reduced, one has to verify that no blind spot is introduced in the detection model.

Since the naive baseline showed some promising results for the task of intrusion detection, the simple detection heuristic could also be further investigated with more suitable curve-fitting methods as described in [147].

The experiments also showed limited success in learning temporal extraction of features. Future works can explore the use of different code dimensions, longer input sequences and also evaluate models that are trained to predict the next fragment in a sequence. By training the network to predict future fragments, the temporal nature of the representation learning is emphasized more.

Lastly, this thesis encourages the use of synthetically-generated data sets in the evaluation process. A trivial extension that can be drawn from this fact, is the idea to also incorporate said generated synthetic anomalies in the training process. The unsupervised problem intrusion detection faces could be enriched with the use of small portions of labelled data that contain specific unwanted behaviour. This process can lift the problem into the category of semi-supervised problems which can be regraded with models as described in [148, 149]. This has the benefit that a suitable decision boundaries may be found in less time and to a greater degree of correctness for specific attacks while also be able to identify unknown anomalies.

# Appendix A

## iTrust Testbed Overview

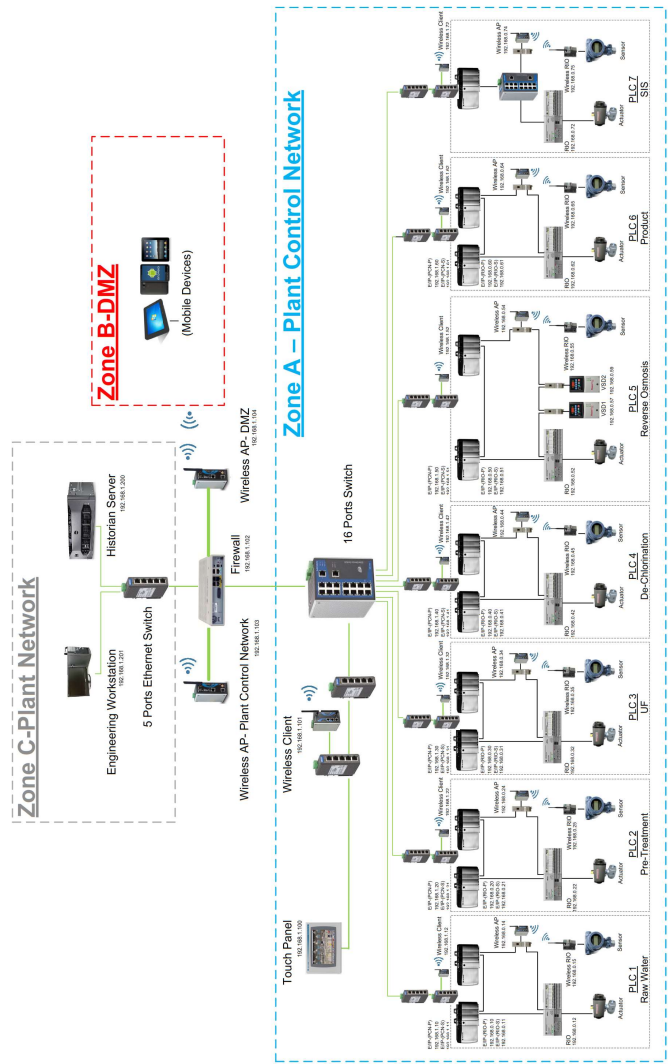

Figure A.1: SWaT testbed network overview

# Appendix B

## Evaluation

| AD Model | Parameter | Value | Parameter Grid |
|---|---|---|---|
| **IsolationForest** | | | |
| | n_estimators | 150 | [100, 150, 200] |
| | max_samples | auto | [auto, 64] |
| | contamination | 0.00001 | [0.00001, 0.0001, 0.01] |
| | max_features | 16 | [8, 16, 32] |
| | bootstrap | True | [True] |
| **LocalOutlierFactor** | | | |
| | n_neighbors | 25 | [25, 50] |
| | algorithm | auto | [auto] |
| | leaf_size | 20 | [20, 30, 40] |
| | metric | minkowski | [minkowski] |
| | p | 2 | [1, 2] |
| | contamination | 0.00001 | [0.00001, 0.0001, 0.01] |
| | novelty | True | [True] |
| **OneClassSVM** | | | |
| | kernel | poly | [poly] |
| | degree | 40 | [30, 40, 50] |
| | gamma | scale | [auto, scale] |
| | tol | 0.01 | [0.01] |
| | nu | 0.0001 | [0.01, 0.0001] |
| | shrinking | True | [True] |
| | max_iter | -1 | [-1] |

Table B.1: Overview about the used hyperparameters for each AD model

| Fragment | $S_n$ | Split | Time | #Samples | #Anomalies | Time | #Samples | #Anomalies |
|---|---|---|---|---|---|---|---|---|
| Byte | | | | SWaT | | | Voerde | |
| | $S_1$ | | | | | | | |
| | | $Train$ | 01:17:12 | 1,107,198 | 0 | 02:09:22 | 1,540,638 | 0 |
| | | $Validation$ | 00:00:44 | 10,990 | 0 | 00:17:08 | 220,089 | 0 |
| | | $Eval_H$ | 01:06:35 | 821,442 | 11,208 | 00:55:20 | 708,697 | 620,150 |
| | | $Eval_E$ | 00:42:25 | 684,501 | 288 | 01:17:17 | 880,363 | 1,125 |
| | | $Eval_D$ | 00:40:44 | 697,304 | 25,883 | 00:38:34 | 497,315 | 22,871 |
| | | $Eval_R$ | 00:49:32 | 684,426 | 101 | 00:57:52 | 747,924 | 1,302 |
| | $S_3$ | | | | | | | |
| | | $Train$ | 03:06:50 | 1,107,196 | 0 | 04:03:09 | 1,540,636 | 0 |
| | | $Validation$ | 00:01:47 | 10,988 | 0 | 00:33:26 | 220,087 | 0 |
| | | $Eval_H$ | 02:12:25 | 821,440 | 11,210 | 02:01:29 | 708,695 | 675,240 |
| | | $Eval_E$ | 01:35:33 | 684,499 | 578 | 02:46:01 | 880,361 | 2,930 |
| | | $Eval_D$ | 01:32:32 | 697,302 | 35,642 | 01:32:51 | 497,313 | 24,441 |
| | | $Eval_R$ | 00:49:32 | 684,424 | 232 | 02:21:21 | 747,922 | 2,979 |
| Flow | | | | SWaT | | | Voerde | |
| | $S_1$ | | | | | | | |
| | | $Train$ | 00:27:18 | 482,280 | 0 | 00:42:55 | 505,539 | 0 |
| | | $Validation$ | 00:00:16 | 4624 | 0 | 00:06:11 | 73,164 | 0 |
| | | $Eval_H$ | 00:22:09 | 346,786 | 1,123 | 00:05:03 | 70,908 | 19,456 |
| | | $Eval_E$ | 00:20:37 | 291,373 | 29 | 00:05:03 | 292,702 | 39 |
| | | $Eval_D$ | 00:18:03 | 295,128 | 3,776 | 00:13:01 | 167,410 | 9,293 |
| | | $Eval_R$ | 00:18:45 | 291,443 | 14 | 00:19:14 | 248,764 | 1,158 |
| | $S_3$ | | | | | | | |
| | | $Train$ | 00:28:08 | 482,278 | 0 | 01:23:16 | 505,537 | 0 |
| | | $Validation$ | 00:00:16 | 4,622 | 0 | 00:11:46 | 73,162 | 0 |
| | | $Eval_H$ | 00:21:33 | 346,784 | 1,349 | 00:11:43 | 70,906 | 19,493 |
| | | $Eval_E$ | 00:45:10 | 291,371 | 55 | 00:05:03 | 292,700 | 53 |
| | | $Eval_D$ | 00:38:51 | 295,126 | 3,808 | 00:12:41 | 167,408 | 10,089 |
| | | $Eval_R$ | 00:17:32 | 291,441 | 42 | 00:07:54 | 248,762 | 2,968 |

Table B.2: Anomaly detection samples per data split

# B.1 SWaT

| AD | Fragments | Model | T2F↓ | FPR↓ | PR↑ | RC↑ | $F_1$↑ | T2T↓ |
|---|---|---|---|---|---|---|---|---|
| IF | | | | | | | | |
| | Byte | | | | | | | |
| | | $S_1$ | 00:07:31 | 0 | 1 | 1 | 1 | 00:03:56 |
| | | $S_3$ | 00:05:58 | 0 | 1 | 1 | 1 | 00:02:44 |
| | | $B_r$ | 00:45:14 | 0.0001 | 0 | 1 | 0 | 00:50:27 |
| | Flow | | | | | | | |
| | | $S_1$ | 00:05:15 | 0 | 1 | 1 | 1 | 00:01:23 |
| | | $S_3$ | 00:05:11 | 0 | 1 | 1 | 1 | 00:01:43 |
| | | $B_r$ | 00:36:12 | 0.0089 | 0 | 1 | 0 | 00:21:39 |
| OCSVM | | | | | | | | |
| | Byte | | | | | | | |
| | | $S_1$ | 00:05:52 | 0 | 1 | 1 | 1 | 00:05:52 |
| | | $S_3$ | 00:04:32 | 0.7869 | 0 | 1 | 0 | 00:04:32 |
| | | $B_r$ | 19:51:57 | 0.5220 | 0 | 1 | 0 | 19:51:57 |
| | Flow | | | | | | | |
| | | $S_1$ | 00:01:02 | 0 | 1 | 1 | 1 | 00:01:02 |
| | | $S_3$ | 00:01:07 | 0 | 1 | 1 | 1 | 00:01:07 |
| | | $B_r$ | 10:13:46 | 0.5220 | 0 | 1 | 0 | 10:13:46 |
| LOF | | | | | | | | |
| | Byte | | | | | | | |
| | | $S_1$ | 00:01:19 | 0 | 1 | 1 | 1 | 00:01:19 |
| | | $S_3$ | 00:05:50 | 0 | 1 | 1 | 1 | 00:05:50 |
| | | $B_r$ | 40:13:40 | 0 | 1 | 1 | 1 | 40:13:40 |
| | Flow | | | | | | | |
| | | $S_1$ | 00:01:12 | 0 | 1 | 1 | 1 | 00:01:12 |
| | | $S_3$ | 01:24:37 | 0 | 1 | 1 | 1 | 01:24:37 |
| | | $B_r$ | 06:00:33 | 0 | 1 | 1 | 1 | 06:00:33 |
| $B_n$ | | | | | | | | |
| | Byte | | | | | | | |
| | | $S_1$ | 05:31:37 | 0.0035 | 0 | 1 | 0 | 05:31:37 |
| | | $S_3$ | 11:07:54 | 0.0234 | 0 | 1 | 0 | 11:07:54 |
| | Flow | | | | | | | |
| | | $S_1$ | 02:19:16 | 0.0702 | 0 | 1 | 0 | 02:19:16 |
| | | $S_3$ | 04:41:41 | 0.0014 | 0 | 1 | 0 | 04:41:41 |

Table B.3: SWaT train validation results

| AD | Fragments | Model | FPR↓ | PR↑ | RC↑ | $F_1$↑ | T2T↓ |
|---|---|---|---|---|---|---|---|
| IF | | | | | | | |
| | Byte | | | | | | |
| | | $S_1$ | 0 | 1 | 0 | 0 | 00:13:29 |
| | | $S_3$ | 0 | 0 | 0 | 0 | 00:11:38 |
| | | $B_r$ | 0 | 0.9783 | 0.0120 | 0.0238 | 01:15:11 |
| | Flow | | | | | | |
| | | $S_1$ | 0 | 0 | 0 | 0 | 00:09:32 |
| | | $S_3$ | 0 | 1 | 0 | 0 | 00:03:33 |
| | | $B_r$ | 0 | 0 | 0 | 0 | 00:31:27 |
| OCSVM | | | | | | | |
| | Byte | | | | | | |
| | | $S_1$ | 0 | 1 | 0 | 0 | 00:03:36 |
| | | $S_3$ | 0.7915 | 0.0132 | 0.7656 | 0.0260 | 00:03:14 |
| | | $B_r$ | 0.4316 | 0.0022 | 0.2876 | 0.0043 | 62:53:53 |
| | Flow | | | | | | |
| | | $S_1$ | 0 | 1 | 0 | 0 | 00:02:45 |
| | | $S_3$ | 0 | 1 | 0 | 0 | 00:01:14 |
| | | $B_r$ | 0.4316 | 0.0022 | 0.2876 | 0.0043 | 62:53:53 |
| LOF | | | | | | | |
| | Byte | | | | | | |
| | | $S_1$ | 0 | 1 | 0 | 0 | 01:32:54 |
| | | $S_3$ | 0 | 0 | 0 | 0 | 06:03:28 |
| | | $B_r$ | 0 | 0.5294 | 0.0008 | 0.0016 | 69:04:37 |
| | Flow | | | | | | |
| | | $S_1$ | 0 | 0 | 0 | 0 | 01:28:06 |
| | | $S_3$ | 0 | 0 | 0 | 0 | 00:21:47 |
| | | $B_r$ | 0 | 0.0349 | 0.0001 | 0.0001 | 10:14:01 |
| $B_n$ | | | | | | | |
| | Byte | | | | | | |
| | | $S_1$ | 0.01 | 0.4959 | 0.7079 | <u>0.5832</u> | 05:31:39 |
| | | $S_3$ | 0.0302 | 0.3059 | 0.9620 | 0.4642 | 11:07:56 |
| | Flow | | | | | | |
| | | $S_1$ | 0.0732 | 0.0185 | 0.4239 | 0.0354 | 02:19:17 |
| | | $S_3$ | 0.0040 | 0.2463 | 0.3306 | <u>0.2823</u> | 04:41:42 |

Table B.4: SWaT anomaly detection results of $Eval_H$

| AD | Fragments | Model | FPR↓ | PR↑ | RC↑ | $F_1$↑ | T2T↓ |
|---|---|---|---|---|---|---|---|
| IF | | | | | | | |
| | Byte | | | | | | |
| | | $S_1$ | 0.9999 | 0.0371 | 0.9999 | 0.0716 | 00:01:08 |
| | | $S_3$ | 0.9987 | 0.0511 | 0.9992 | <u>0.0973</u> | 00:01:09 |
| | | $B_r$ | 1 | 0.0371 | 1 | 0.0716 | 00:18:27 |
| | Flow | | | | | | |
| | | $S_1$ | 0.0705 | 0.0261 | 0.1457 | 0.0442 | 00:00:28 |
| | | $S_3$ | 0 | 1 | 0 | 0 | 00:00:17 |
| | | $B_r$ | 0.9346 | 0.0137 | 1 | 0.0270 | 00:08:07 |
| OCSVM | | | | | | | |
| | Byte | | | | | | |
| | | $S_1$ | 1 | 0.0371 | 1 | 0.0716 | 00:00:36 |
| | | $S_3$ | 0.0005 | 0.0384 | 0.0004 | 0.0008 | 00:00:38 |
| | | $B_r$ | 1 | 0.0369 | 0.9950 | 0.0712 | 07:14:04 |
| | Flow | | | | | | |
| | | $S_1$ | 1 | 0.0128 | 1 | 0.0253 | 00:00:16 |
| | | $S_3$ | 1 | 0.0129 | 1 | 0.0255 | 00:00:18 |
| | | $B_r$ | 1 | 0.0127 | 0.9902 | 0.0250 | 03:52:22 |
| LOF | | | | | | | |
| | Byte | | | | | | |
| | | $S_1$ | 0.9999 | 0.0371 | 1 | 0.0716 | 06:33:01 |
| | | $S_3$ | 1 | 0.0511 | 1 | 0.0973 | 06:14:50 |
| | | $B_r$ | 1 | 0.0371 | 1 | 0.0716 | 11:13:28 |
| | Flow | | | | | | |
| | | $S_1$ | 0.9387 | 0.0136 | 1 | 0.0269 | 01:05:51 |
| | | $S_3$ | 1 | 0.0129 | 1 | 0.0255 | 00:48:12 |
| | | $B_r$ | 1 | 0.0127 | 0.9952 | 0.0251 | 05:32:56 |
| $B_n$ | | | | | | | |
| | Byte | | | | | | |
| | | $S_1$ | 0.0091 | 0.0152 | 0.0036 | 0.0059 | 05:32:02 |
| | | $S_3$ | 0.0281 | 0.0543 | 0.03 | 0.0386 | 11:17:33 |
| | Flow | | | | | | |
| | | $S_1$ | 0.0085 | 0.2227 | 0.1870 | <u>0.2033</u> | 02:25:40 |
| | | $S_3$ | 0.0030 | 0.0618 | 0.0150 | 0.0241 | 04:52:41 |

Table B.5: SWaT anomaly detection results of $Eval_D$

| AD | Fragments | Model | FPR↓ | PR↑ | RC↑ | $F_1$↑ | T2T↓ |
|---|---|---|---|---|---|---|---|
| IF | | | | | | | |
| | Byte | | | | | | |
| | | $S_1$ | 0.9999 | 0.0004 | 1 | 0.0008 | 00:00:45 |
| | | $S_3$ | 0.9987 | 0.0008 | 1 | 0.0017 | 00:00:46 |
| | | $B_r$ | 1 | 0.0004 | 1 | 0.0008 | 00:18:57 |
| | Flow | | | | | | |
| | | $S_1$ | 0.0706 | 0.0001 | 0.1034 | 0.0003 | 00:00:18 |
| | | $S_3$ | 0 | 1 | 0 | 0 | 00:00:19 |
| | | $B_r$ | 0.9346 | 0.0001 | 1 | 0.0002 | 00:07:56 |
| OCSVM | | | | | | | |
| | Byte | | | | | | |
| | | $S_1$ | 1 | 0 | 1 | 0.0001 | 00:00:12 |
| | | $S_3$ | 1 | 0.0001 | 1 | 0.0003 | 00:00:12 |
| | | $B_r$ | 0.0169 | 0.0004 | 0.0169 | 0.0008 | 06:54:35 |
| | Flow | | | | | | |
| | | $S_1$ | 1 | 0.0001 | 1 | 0.0002 | 00:00:18 |
| | | $S_3$ | 1 | 0.0002 | 1 | 0.0004 | 00:00:12 |
| | | $B_r$ | 0.0010 | 0.0001 | 0.0010 | 0.0002 | 02:48:17 |
| LOF | | | | | | | |
| | Byte | | | | | | |
| | | $S_1$ | 0.9999 | 0.0004 | 1 | 0.0008 | 06:31:27 |
| | | $S_3$ | 1 | 0.0008 | 1 | 0.0017 | 06:08:44 |
| | | $B_r$ | 0.1050 | 0.0003 | 0.1050 | 0.0007 | 09:46:53 |
| | Flow | | | | | | |
| | | $S_1$ | 0.9387 | 0.0001 | 1 | 0.0002 | 01:05:39 |
| | | $S_3$ | 1 | 0.0002 | 1 | 0.0004 | 00:42:48 |
| | | $B_r$ | 0.0968 | 0.0001 | 0.0968 | 0.0002 | 04:17:47 |
| $B_n$ | | | | | | | |
| | Byte | | | | | | |
| | | $S_1$ | 0.0090 | 0 | 0 | 0 | 05:28:49 |
| | | $S_3$ | 0.0278 | 0.0015 | 0.0484 | <u>0.0029</u> | 11:05:08 |
| | Flow | | | | | | |
| | | $S_1$ | 0.0085 | 0.0020 | 0.1724 | <u>0.0040</u> | 02:19:47 |
| | | $S_3$ | 0.0030 | 0.0012 | 0.0182 | 0.0022 | 04:50:35 |

Table B.6: SWaT anomaly detection results of $Eval_E$

| AD | Fragments | Model | FPR↓ | PR↑ | RC↑ | $F_1$↑ | T2T↓ |
|---|---|---|---|---|---|---|---|
| IF | | | | | | | |
| | Byte | | | | | | |
| | | $S_1$ | 0.9999 | 0.0001 | 1 | 0.0003 | 00:00:59 |
| | | $S_3$ | 0.9987 | 0.0003 | 0.9957 | 0.0007 | 00:00:10 |
| | | $B_r$ | 1 | 0.0001 | 1 | 0.0003 | 00:19:11 |
| | Flow | | | | | | |
| | | $S_1$ | 0.0706 | 0 | 0 | 0 | 00:00:27 |
| | | $S_3$ | 0 | 1 | 0 | 0 | 00:00:26 |
| | | $B_r$ | 0.9346 | 0.0001 | 1 | 0.0001 | 00:07:59 |
| OCSVM | | | | | | | |
| | Byte | | | | | | |
| | | $S_1$ | 1 | 0.0001 | 1 | 0.0003 | 00:00:28 |
| | | $S_3$ | 0.0006 | 0 | 0 | 0 | 00:00:28 |
| | | $B_r$ | 0.0004 | 0.0001 | 0.0004 | 0.0002 | 06:08:07 |
| | Flow | | | | | | |
| | | $S_1$ | 1 | 0.0128 | 1 | 0.0253 | 00:00:16 |
| | | $S_3$ | 1 | 0.0129 | 1 | <u>0.0255</u> | 00:00:18 |
| | | $B_r$ | 0 | 0 | 0 | 0 | 02:46:23 |
| LOF | | | | | | | |
| | Byte | | | | | | |
| | | $S_1$ | 0.9999 | 0.0001 | 1 | 0.0003 | 05:10:38 |
| | | $S_3$ | 1 | 0.0003 | 1 | 0.0007 | 04:33:09 |
| | | $B_r$ | 0 | 0.0001 | 0 | 0 | 08:44:23 |
| | Flow | | | | | | |
| | | $S_1$ | 0.9387 | 0.0001 | 1 | 0.0001 | 00:44:55 |
| | | $S_3$ | 1 | 0.0001 | 1 | 0.0003 | 00:42:13 |
| | | $B_r$ | 0 | 0 | 0 | 0 | 04:01:58 |
| $B_n$ | | | | | | | |
| | Byte | | | | | | |
| | | $S_1$ | 0.0089 | 0.0005 | 0.0297 | <u>0.0010</u> | 05:24:23 |
| | | $S_3$ | 0.0276 | 0.0005 | 0.0431 | 0.0010 | 11:09:59 |
| | Flow | | | | | | |
| | | $S_1$ | 0.0084 | 0.0004 | 0.0714 | 0.0008 | 02:20:37 |
| | | $S_3$ | 0.0028 | 0 | 0 | 0 | 04:46:05 |

Table B.7: SWaT anomaly detection results of $Eval_R$

## B.2 Voerde

| AD | Fragments | Model | T2F↓ | FPR↓ | PR↑ | RC↑ | $F_1$↑ | T2T↓ |
|---|---|---|---|---|---|---|---|---|
| IF | | | | | | | | |
| | Byte | | | | | | | |
| | | $S_1$ | 00:09:11 | 0 | 1 | 1 | 1 | 00:08:49 |
| | | $S_3$ | 00:09:23 | 0 | 1 | 1 | 1 | 00:08:48 |
| | | $B_r$ | 02:53:19 | 1 | 0 | 1 | 0 | 02:14:33 |
| | Flow | | | | | | | |
| | | $S_1$ | 00:02:47 | 0 | 1 | 1 | 1 | 00:02:36 |
| | | $S_3$ | 00:02:14 | 0.0001 | 0 | 1 | 0 | 00:01:50 |
| | | $B_r$ | 00:46:08 | 0.9999 | 0 | 1 | 0 | 00:25:03 |
| OCSVM | | | | | | | | |
| | Byte | | | | | | | |
| | | $S_1$ | 00:05:14 | 0 | 1 | 1 | 1 | 00:04:48 |
| | | $S_3$ | 00:04: | 0.8073 | 0 | 1 | 0 | 00:04:46 |
| | | $B_r$ | 05:51:04 | 1 | 0 | 1 | 0 | 01:51:14 |
| | Flow | | | | | | | |
| | | $S_1$ | 00:04:53 | 0 | 1 | 1 | 1 | 00:01:03 |
| | | $S_3$ | 00:04:29 | 0.9581 | 0 | 1 | 0 | 00:01:04 |
| | | $B_r$ | 00:43:17 | 1 | 0 | 1 | 0 | 00:21:42 |
| LOF | | | | | | | | |
| | Byte | | | | | | | |
| | | $S_1$ | 08:13:37 | 0.0001 | 0 | 1 | 0 | 20:43:11 |
| | | $S_3$ | 07:42:55 | 0.0001 | 0 | 1 | 0 | 18:36:49 |
| | | $B_r$ | 01:24:54 | 0.2151 | 0 | 1 | 0 | 03:11:54 |
| | Flow | | | | | | | |
| | | $S_1$ | 01:09:22 | 0.0034 | 0 | 1 | 0 | 01:37:37 |
| | | $S_3$ | 00:58:43 | 0.0031 | 0 | 1 | 0 | 01:16:24 |
| | | $B_r$ | 01:54:13 | 0.9215 | 0 | 1 | 0 | 01:38:53 |
| $B_n$ | | | | | | | | |
| | Byte | | | | | | | |
| | | $S_1$ | 06:41:19 | 0.0401 | 0 | 1 | 0 | 06:45:46 |
| | | $S_3$ | 13:32:55 | 0.0946 | 0 | 1 | 0 | 13:37:38 |
| | Flow | | | | | | | |
| | | $S_1$ | 01:29:13 | 0.0001 | 0 | 1 | 0 | 01:30:29 |
| | | $S_3$ | 03:18:48 | 0.0286 | 0 | 1 | 0 | 03:20:08 |

Table B.8: Voerde train validation results

| AD | Fragments | Model | FPR↓ | PR↑ | RC↑ | $F_1$↑ | T2T↓ |
|----|-----------|-------|------|-----|-----|--------|------|
| IF | | | | | | | |
| | Byte | | | | | | |
| | | $S_1$ | 1 | 0.8751 | 1 | <u>0.9334</u> | 00:01:29 |
| | | $S_3$ | 0 | 1 | 0 | 0 | 00:02:54 |
| | | $B_r$ | 1 | 0.8751 | 1 | 0.9334 | 02:35:33 |
| | Flow | | | | | | |
| | | $S_1$ | 1 | 0.2744 | 1 | 0.4306 | 00:00:11 |
| | | $S_3$ | 1 | 0.2749 | 1 | 0.4313 | 00:00:12 |
| | | $B_r$ | 1 | 0.2744 | 1 | 0.4306 | 00:27:14 |
| OCSVM | | | | | | | |
| | Byte | | | | | | |
| | | $S_1$ | 1 | 0.8751 | 1 | 0.9334 | 00:02:27 |
| | | $S_3$ | 0.1432 | 0.8705 | 0.0477 | 0.0905 | 00:01:08 |
| | | $B_r$ | 0 | 1 | 0 | 0 | 02:35:33 |
| | Flow | | | | | | |
| | | $S_1$ | 1 | 0.2744 | 1 | <u>0.4306</u> | 00:00:09 |
| | | $S_3$ | 0.0738 | 0.0800 | 0.0169 | 0.0279 | 00:00:10 |
| | | $B_r$ | 1 | 0.0352 | 0.0600 | 0.0444 | 00:27:14 |
| LOF | | | | | | | |
| | Byte | | | | | | |
| | | $S_1$ | 0.9996 | 0.8748 | 0.9969 | 0.9318 | 08:01:23 |
| | | $S_3$ | 1 | 0.9528 | 0.9998 | 0.9757 | 07:11:40 |
| | | $B_r$ | 0.1249 | 1 | 0.1249 | 0 | 02:35:33 |
| | Flow | | | | | | |
| | | $S_1$ | 1 | 0.2737 | 0.9967 | 0.4295 | 00:11:08 |
| | | $S_3$ | 1 | 0.2727 | 0.9888 | 0.4275 | 00:10:46 |
| | | $B_r$ | 0.9993 | 0.0014 | 0.0009 | 0.0011 | 02:33:07 |
| $B_n$ | | | | | | | |
| | Byte | | | | | | |
| | | $S_1$ | 0.0419 | 0.9919 | 0.7363 | 0.8452 | 06:46:28 |
| | | $S_3$ | 0.0716 | 0.6195 | 0.0058 | 0.0115 | 13:38:21 |
| | Flow | | | | | | |
| | | $S_1$ | 0 | 1 | 0 | 0 | 01:30:42 |
| | | $S_3$ | 0.0498 | 0.5666 | 0.1715 | 0.2634 | 03:20:22 |

Table B.9: Voerde anomaly detection results of $Eval_H$

| AD | Fragments | Model | FPR↓ | PR↑ | RC↑ | $F_1$↑ | T2T↓ |
|---|---|---|---|---|---|---|---|
| IF | | | | | | | |
| | Byte | | | | | | |
| | | $S_1$ | 1 | 0.0460 | 1 | 0.0879 | 00:01:34 |
| | | $S_3$ | 1 | 0.0491 | 1 | 0.0937 | 00:01:35 |
| | | $B_r$ | 1 | 0.0460 | 1 | 0.0879 | 00:13:48 |
| | Flow | | | | | | |
| | | $S_1$ | 1 | 0.0555 | 1 | 0.1052 | 00:00:34 |
| | | $S_3$ | 0.9999 | 0.0603 | 1 | 0.1137 | 00:00:35 |
| | | $B_r$ | 1 | 0.0555 | 1 | 0.1052 | 00:04:40 |
| OCSVM | | | | | | | |
| | Byte | | | | | | |
| | | $S_1$ | 1 | 0.0460 | 1 | 0.0879 | 00:01:19 |
| | | $S_3$ | 0.1944 | 0.0405 | 0.1588 | 0.0645 | 00:01:19 |
| | | $B_r$ | 0.2772 | 0.0401 | 0.2772 | 0.07 | 04:02:34 |
| | Flow | | | | | | |
| | | $S_1$ | 1 | 0.0555 | 1 | 0.1052 | 00:00:29 |
| | | $S_3$ | 0.0934 | 0.0527 | 0.0811 | 0.0639 | 00:00:29 |
| | | $B_r$ | 0.3185 | 0.0485 | 0.3185 | 0.0841 | 03:23:47 |
| LOF | | | | | | | |
| | Byte | | | | | | |
| | | $S_1$ | 1 | 0.0460 | 1 | 0.0879 | 04:41:52 |
| | | $S_3$ | 0.9999 | 0.0491 | 1 | 0.0937 | 04:33:38 |
| | | $B_r$ | 0.8875 | 0.0447 | 0.8875 | 0.0851 | 03:21:55 |
| | Flow | | | | | | |
| | | $S_1$ | 0.9912 | 0.0555 | 0.9919 | 0.1052 | 00:25:01 |
| | | $S_3$ | 0.9953 | 0.0605 | 0.9999 | <u>0.1141</u> | 00:24:28 |
| | | $B_r$ | 0.9929 | 0.0539 | 0.9929 | 0.1023 | 02:48:37 |
| $B_n$ | | | | | | | |
| | Byte | | | | | | |
| | | $S_1$ | 0.0408 | 0.0552 | 0.0494 | 0.0521 | 05:59:10 |
| | | $S_3$ | 0.1033 | 0.0921 | 0.2028 | <u>0.1267</u> | 12:11:19 |
| | Flow | | | | | | |
| | | $S_1$ | 0.0003 | 0 | 0 | 0 | 01:56:02 |
| | | $S_3$ | 0.0349 | 0.0590 | 0.0341 | 0.0432 | 03:57:47 |

Table B.10: Voerde anomaly detection results of $Eval_D$

| AD | Fragments | Model | FPR↓ | PR↑ | RC↑ | $F_1$↑ | T2T↓ |
|---|---|---|---|---|---|---|---|
| IF | | | | | | | |
| | Byte | | | | | | |
| | | $S_1$ | 1 | 0.0013 | 1 | 0.0026 | 00:03:03 |
| | | $S_3$ | 1 | 0.0033 | 1 | 0.0066 | 00:02:55 |
| | | $B_r$ | 1 | 0.0013 | 1 | 0.0026 | 00:24:31 |
| | Flow | | | | | | |
| | | $S_1$ | 1 | 0.0001 | 1 | 0.0003 | 00:00:59 |
| | | $S_3$ | 1 | 0.0002 | 1 | 0.0004 | 00:00:57 |
| | | $B_r$ | 1 | 0.0001 | 1 | 0.0003 | 00:08:01 |
| OCSVM | | | | | | | |
| | Byte | | | | | | |
| | | $S_1$ | 1 | 0.0013 | 1 | 0.0026 | 00:02:19 |
| | | $S_3$ | 0.1950 | 0.0019 | 0.1119 | 0.0038 | 00:02:18 |
| | | $B_r$ | 0.0551 | 0.0011 | 0.0551 | 0.0022 | 04:54:11 |
| | Flow | | | | | | |
| | | $S_1$ | 1 | 0.0001 | 1 | 0.0003 | 00:00:48 |
| | | $S_3$ | 0.0757 | 0 | 0.0189 | 0.0001 | 00:00:48 |
| | | $B_r$ | 0.1667 | 0.0001 | 0.1667 | 0.0003 | 03:58:24 |
| LOF | | | | | | | |
| | Byte | | | | | | |
| | | $S_1$ | 1 | 0.0013 | 1 | 0.0026 | 08:18:12 |
| | | $S_3$ | 0.9999 | 0.0033 | 1 | 0.0066 | 08:04:56 |
| | | $B_r$ | 0.1692 | 0.0012 | 0.1692 | 0.0025 | 04:18:47 |
| | Flow | | | | | | |
| | | $S_1$ | 0.9987 | 0.0001 | 1 | 0.0003 | 00:44:09 |
| | | $S_3$ | 1 | 0.0002 | 1 | 0.0004 | 00:42:35 |
| | | $B_r$ | 1 | 0.0001 | 1 | 0.0003 | 03:25:15 |
| $B_n$ | | | | | | | |
| | Byte | | | | | | |
| | | $S_1$ | 0.0367 | 0.0078 | 0.2249 | 0.0151 | 07:15:44 |
| | | $S_3$ | 0.0909 | 0.0173 | 0.4785 | <u>0.0333</u> | 14:45:10 |
| | Flow | | | | | | |
| | | $S_1$ | 0.0002 | 0 | 0 | 0 | 02:22:10 |
| | | $S_3$ | 0.0277 | 0.0012 | 0.1887 | <u>0.0024</u> | 04:55:49 |

Table B.11: Voerde anomaly detection results of $Eval_E$

| AD | Fragments | Model | FPR↓ | PR↑ | RC↑ | $F_1$↑ | T2T↓ |
|---|---|---|---|---|---|---|---|
| IF | | | | | | | |
| | Byte | | | | | | |
| | | $S_1$ | 1 | 0.0017 | 1 | 0.0035 | 00:02:23 |
| | | $S_3$ | 1 | 0.0040 | 1 | 0.0079 | 00:02:24 |
| | | $B_r$ | 1 | 0.0017 | 1 | 0.0035 | 00:21:16 |
| | Flow | | | | | | |
| | | $S_1$ | 1 | 0.0047 | 1 | 0.0093 | 00:00:50 |
| | | $S_3$ | 1 | 0.0119 | 1 | <u>0.0236</u> | 00:00:50 |
| | | $B_r$ | 1 | 0.0047 | 1 | 0.0093 | 00:07:12 |
| OCSVM | | | | | | | |
| | Byte | | | | | | |
| | | $S_1$ | 1 | 0.0017 | 1 | 0.0035 | 00:01:56 |
| | | $S_3$ | 0.1943 | 0.0043 | 0.2085 | <u>0.0084</u> | 00:01:58 |
| | | $B_r$ | 0.2039 | 0.0015 | 0.2039 | 0.0030 | 04:01:21 |
| | Flow | | | | | | |
| | | $S_1$ | 1 | 0.0047 | 1 | 0.0093 | 00:00:41 |
| | | $S_3$ | 0.0676 | 0.0124 | 0.0701 | 0.0210 | 00:00:41 |
| | | $B_r$ | 0.3427 | 0.0037 | 0.3427 | 0.0073 | 03:30:52 |
| LOF | | | | | | | |
| | Byte | | | | | | |
| | | $S_1$ | 1 | 0.0017 | 1 | 0.0035 | 07:01:24 |
| | | $S_3$ | 1 | 0.0040 | 1 | 0.0079 | 06:51:09 |
| | | $B_r$ | 0.3003 | 0.0016 | 0.3003 | 0.0032 | 03:31:52 |
| | Flow | | | | | | |
| | | $S_1$ | 0.9987 | 0.0047 | 1 | 0.0093 | 00:37:27 |
| | | $S_3$ | 1 | 0.0119 | 1 | 0.0236 | 00:36:02 |
| | | $B_r$ | 1 | 0.0047 | 1 | 0.0093 | 02:54:29 |
| $B_n$ | | | | | | | |
| | Byte | | | | | | |
| | | $S_1$ | 0.0370 | 0.0008 | 0.0169 | 0.0015 | 06:38:15 |
| | | $S_3$ | 0.0929 | 0.0020 | 0.0463 | 0.0038 | 13:51:14 |
| | Flow | | | | | | |
| | | $S_1$ | 0.0002 | 0 | 0 | 0 | 02:12:43 |
| | | $S_3$ | 0.0280 | 0.0109 | 0.0256 | 0.0153 | 04:36:28 |

Table B.12: Voerde anomaly detection results of $Eval_R$

## B.3 Naive Baseline

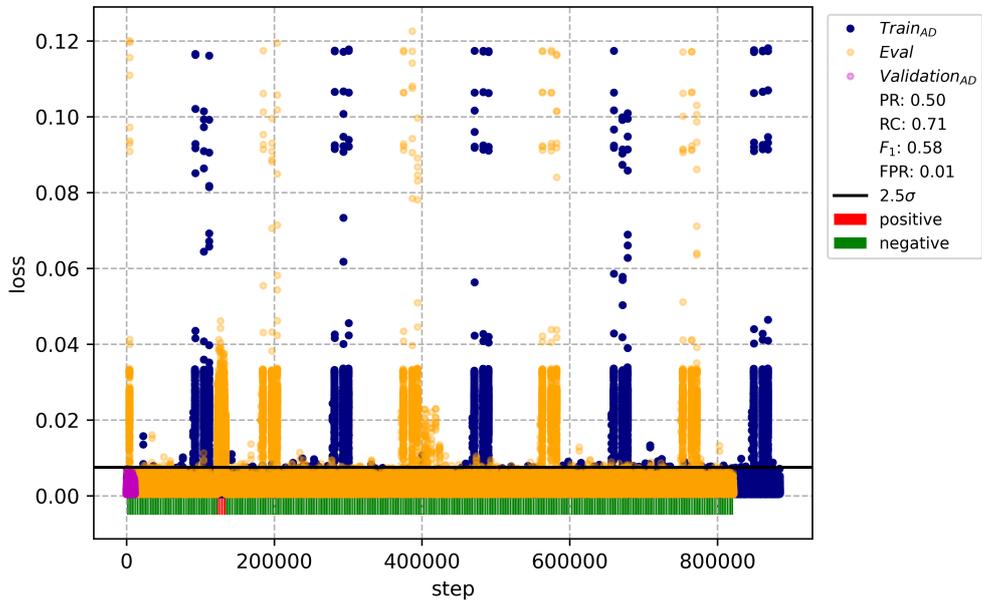

Figure B.1: SWaT A6 $Eval_H$ naive baseline loss visualisation for byte-fragments trained on an input sequence length of $n = 1$

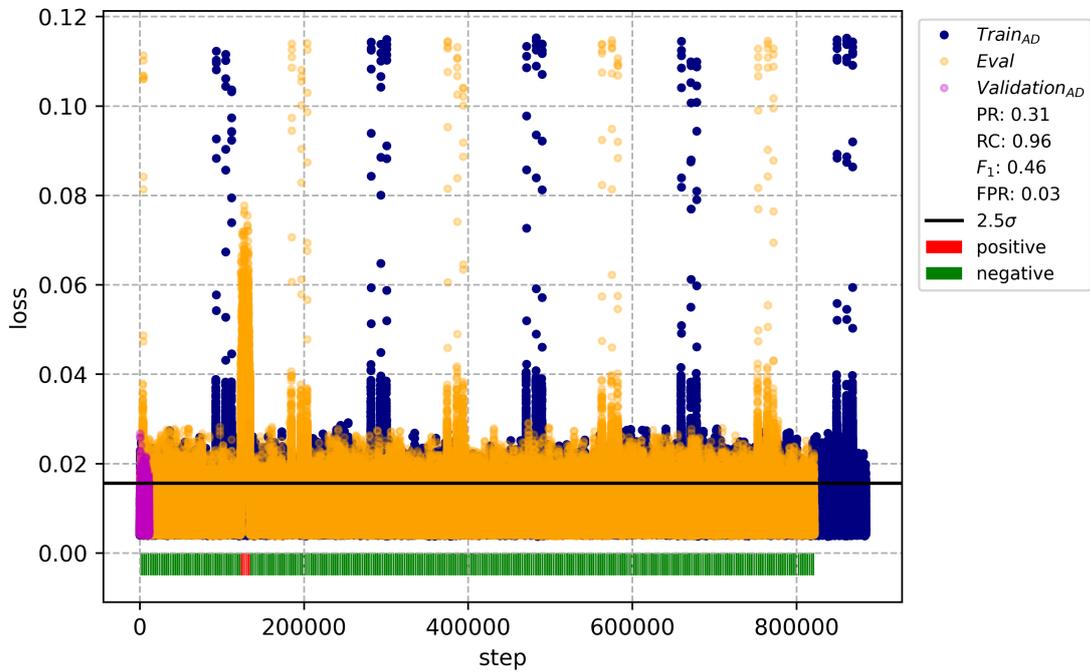

Figure B.2: SWaT A6 $Eval_H$ naive baseline loss visualisation for byte-fragments trained on an input sequence length of $n = 3$

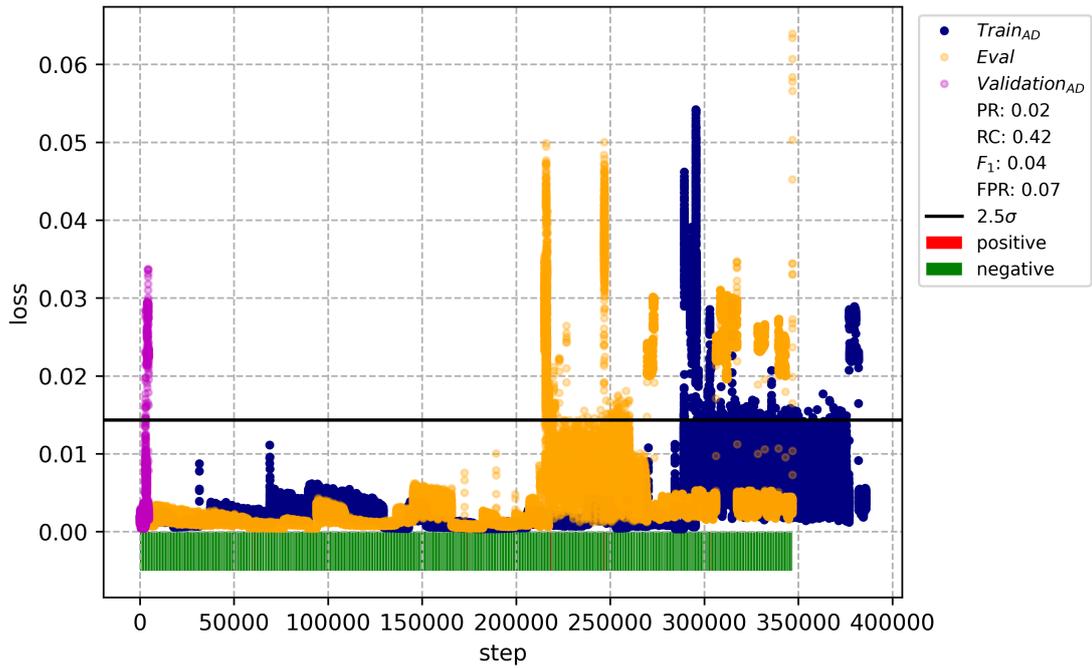

Figure B.3: SWaT A6 $Eval_H$ naive baseline loss visualisation for flow-fragments trained on an input sequence length of $n = 1$

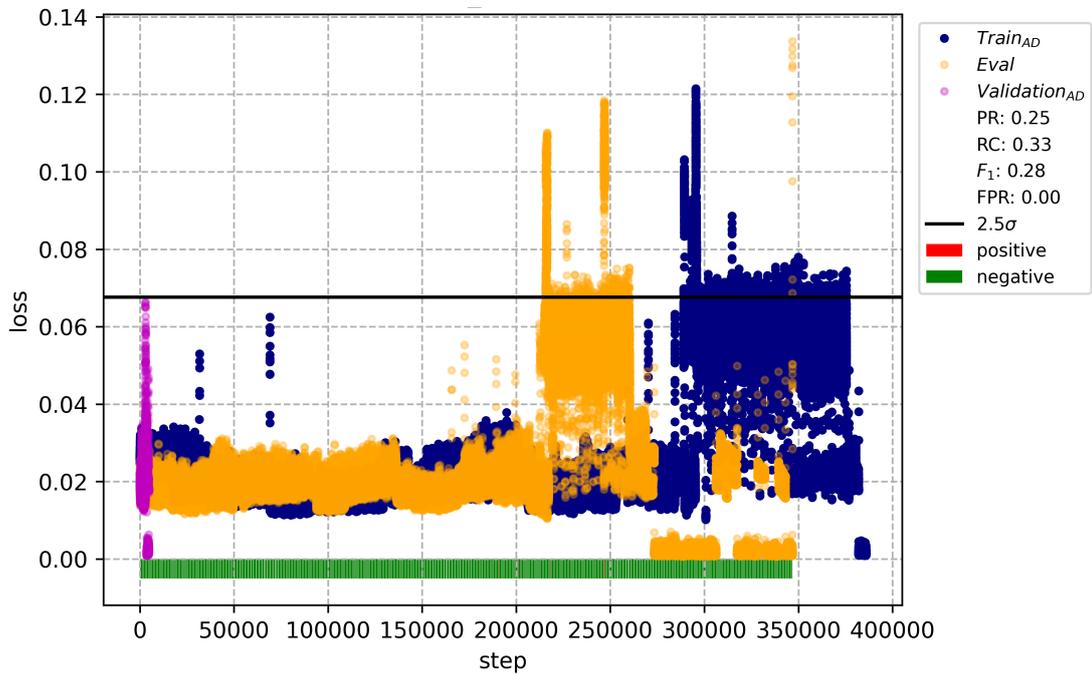

Figure B.4: SWaT A6 $Eval_H$ naive baseline loss visualisation for flow-fragments trained on an input sequence length of $n = 3$

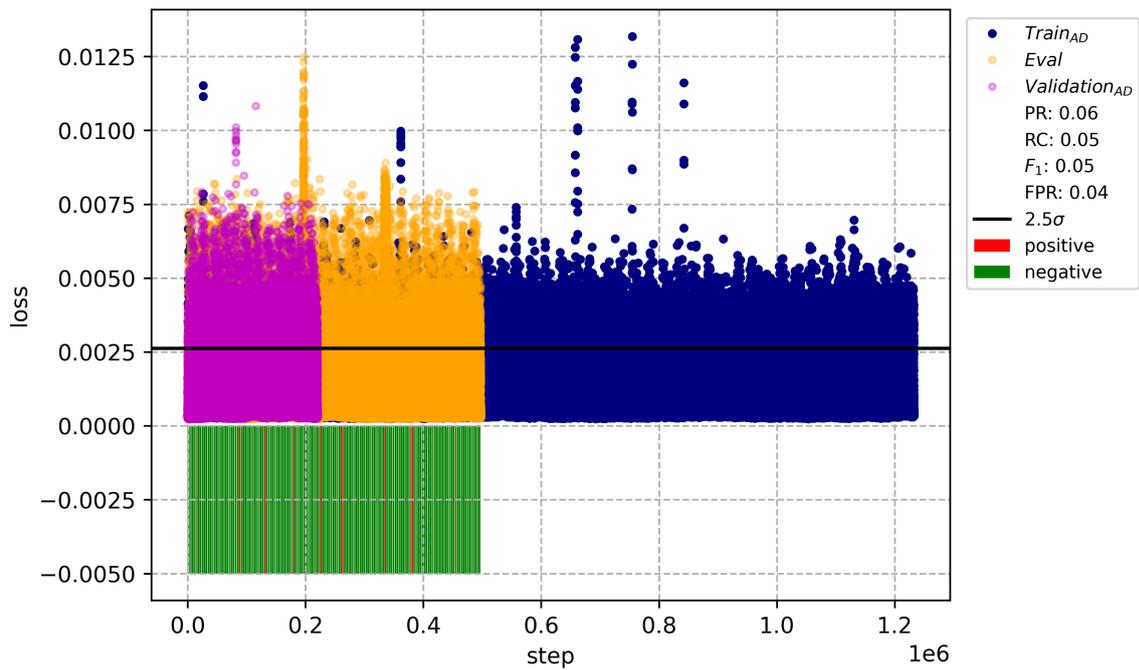

Figure B.5: Voerde $Eval_D$ naive baseline loss visualisation for byte-fragments trained on an input sequence length of $n = 1$

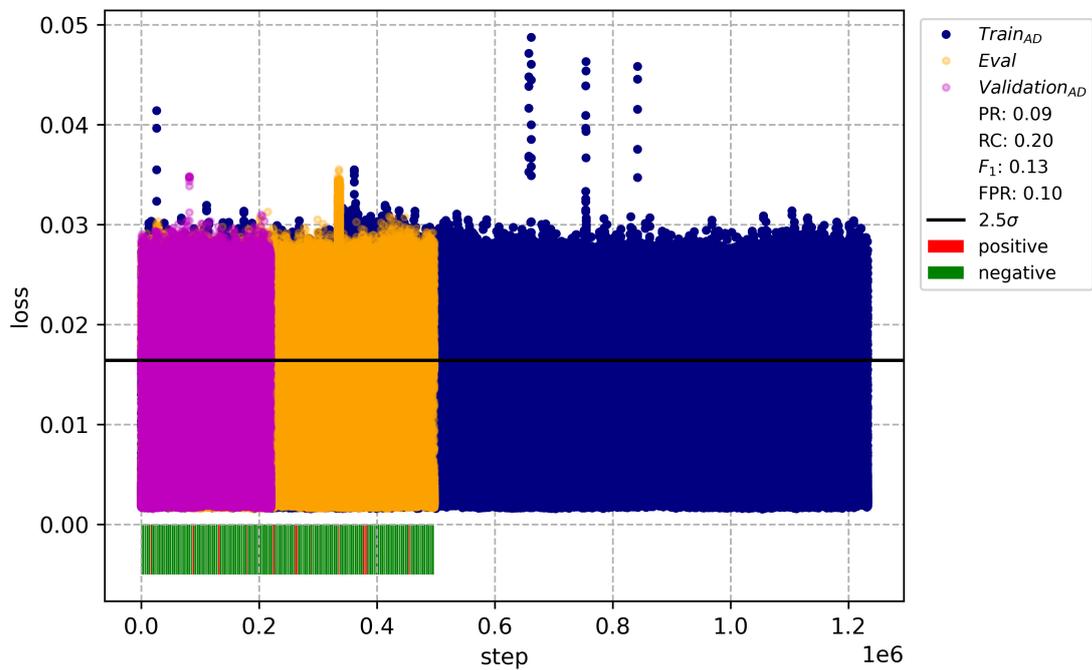

Figure B.6: Voerde $Eval_D$ naive baseline loss visualisation for byte-fragments trained on an input sequence length of $n = 3$

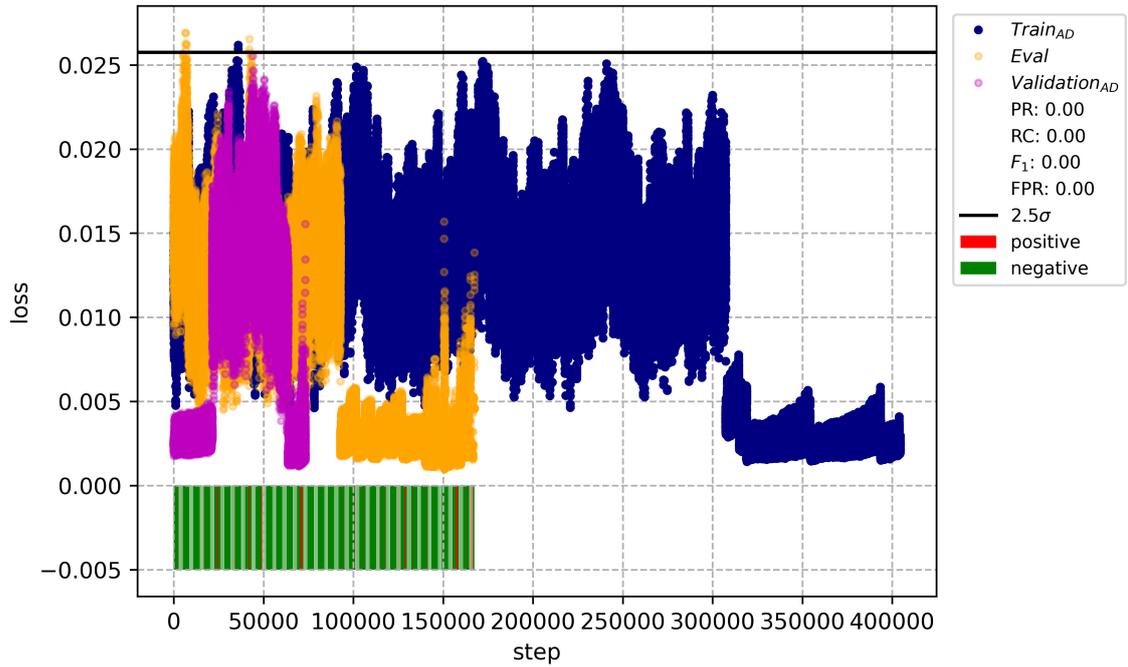

Figure B.7: Voerde $Eval_D$ naive baseline loss visualisation for flow-fragments trained on an input sequence length of $n = 1$

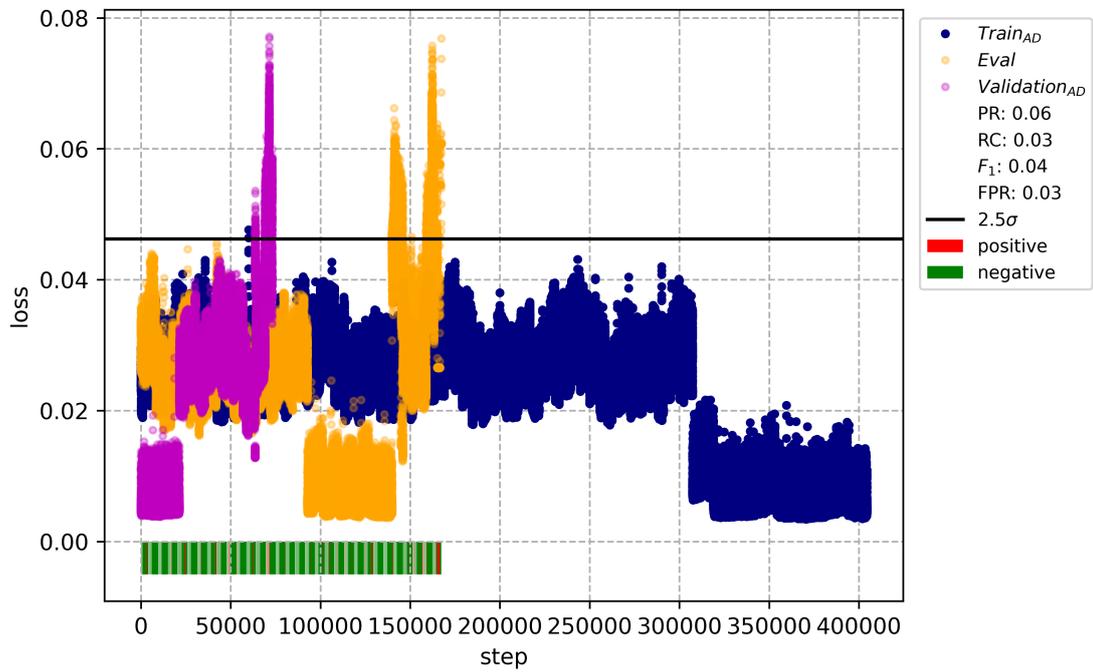

Figure B.8: Voerde $Eval_D$ naive baseline loss visualisation for flow-fragments trained on an input sequence length of $n = 3$

## B.4  Decision Analysis

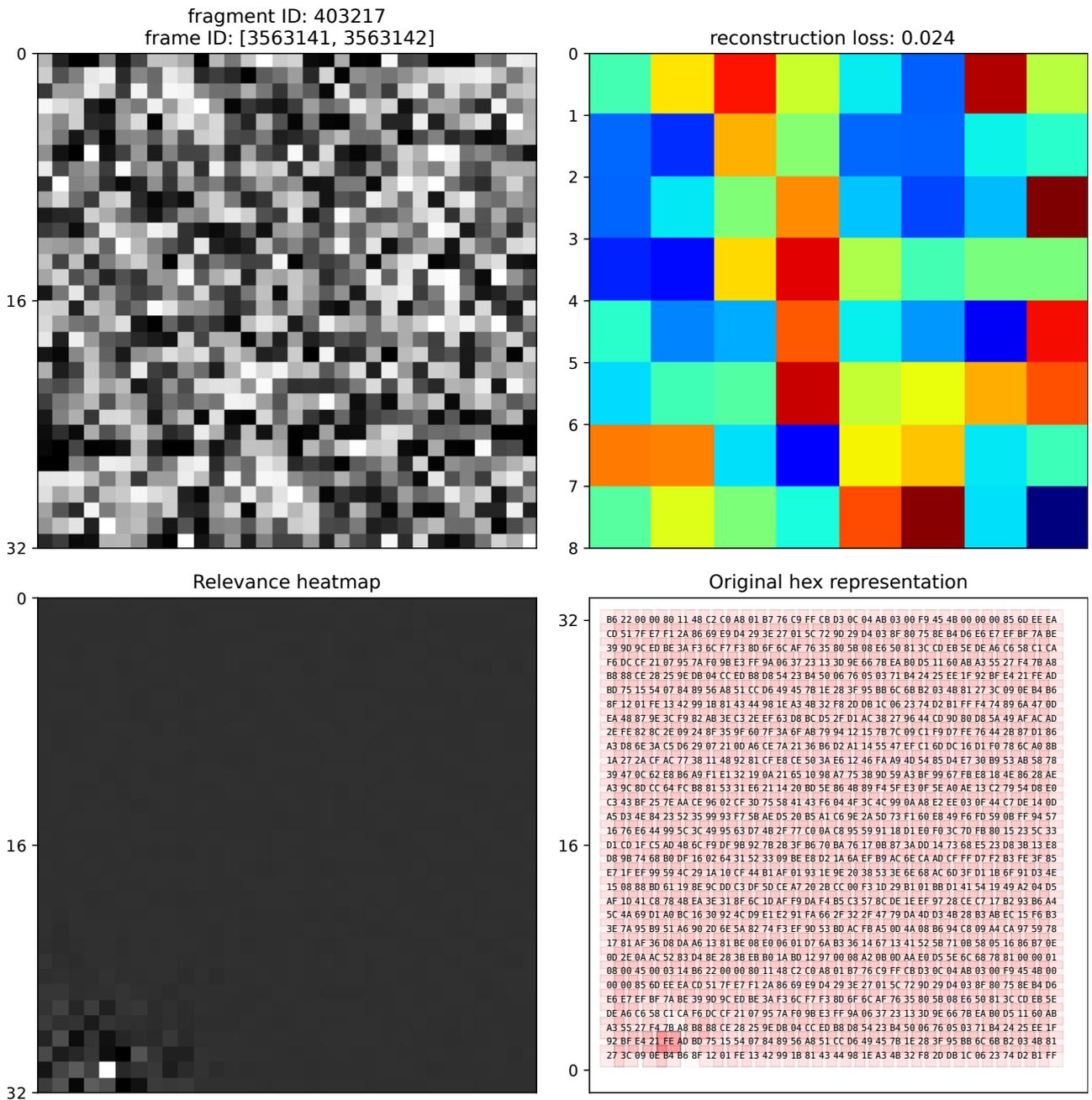

Figure B.9: Byte-fragments relevance heatmaps of the unusual UDP packet

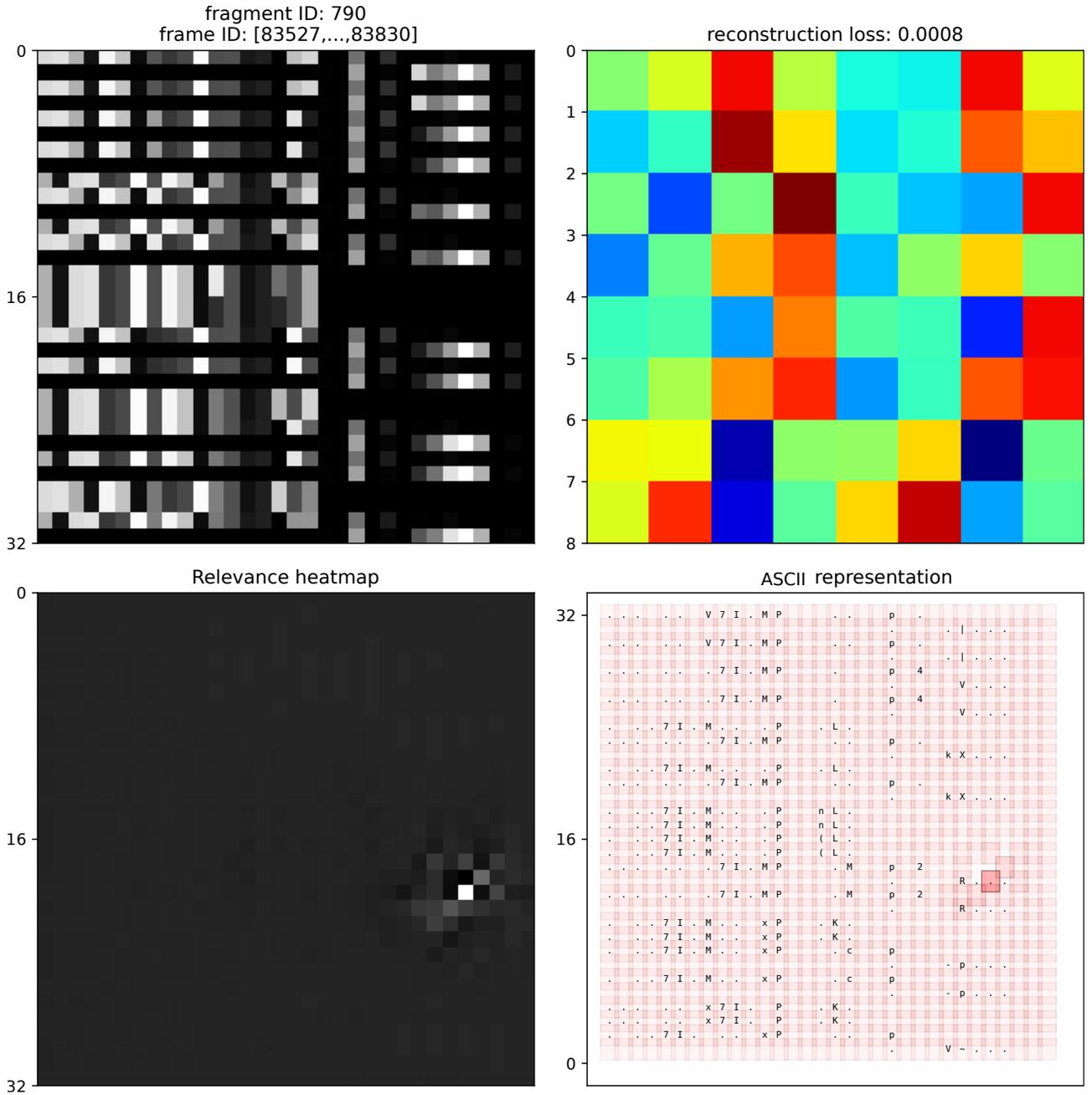

Figure B.10: Relevance heatmap for a benign flow-fragment

Figure B.11: Flow-fragments relevance heatmaps of the file download event

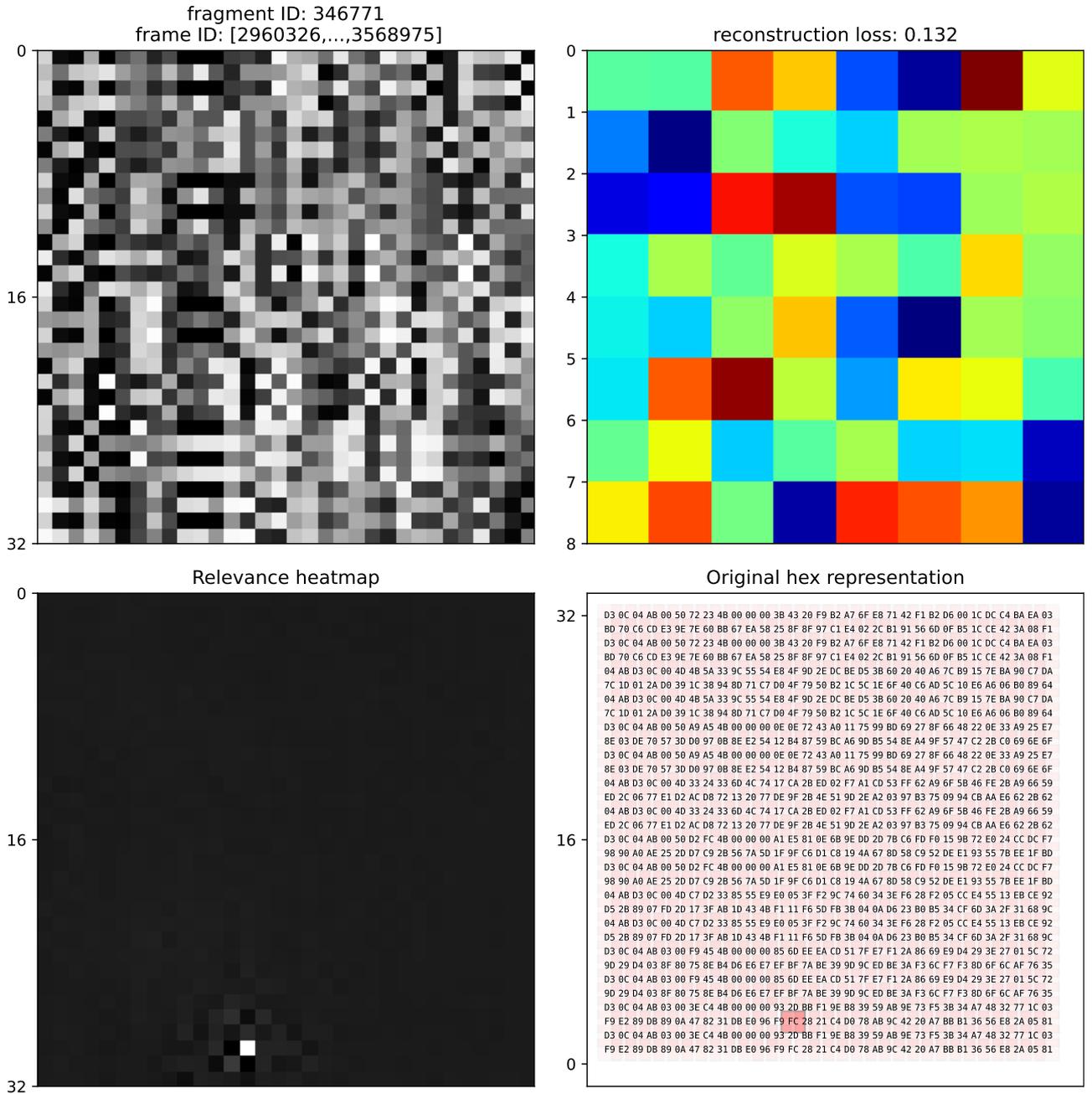

Figure B.12: Flow-fragments relevance heatmaps of the unusual UDP packet

# Appendix C

## Autoencoder Architecture

### C.1 Autoencoder parameter overview

| Layer (type) | Output Shape Encoder | # Parameter | Layer (type) | Output Shape Decoder | # Parameter |
|---|---|---|---|---|---|
| Conv2d-1 | (1, 2, 32, 32) | 20 | CGRU_cell-33 | (1, 4, 4, 4) | 0 |
| LeakyReLU-2 | (1, 2, 32, 32) | 0 | Dropout2d-34 | (1, 4, 4, 4) | 0 |
| Conv2d-3 | (1, 8, 32, 32) | 1,208 | GroupNorm-35 | (1, 4, 4, 4) | 8 |
| GroupNorm-4 | (1, 8, 32, 32) | 16 | Conv2d-36 | (1, 4, 4, 4) | 804 |
| Conv2d-5 | (1, 4, 32, 32) | 604 | GroupNorm-37 | (1, 8, 4, 4) | 16 |
| GroupNorm-6 | (1, 4, 32, 32) | 8 | ConvTrans2d-38 | (1, 8, 4, 4) | 1,608 |
| Dropout2d-7 | (1, 4, 32, 32) | 0 | LeakyReLU-39 | (1, 4, 4, 4) | 0 |
| CGRU_cell-8 | (1, 4, 32, 32) | 0 | Conv2d-40 | (1, 4, 4, 4) | 148 |
| Conv2d-9 | (1, 4, 16, 16) | 148 | CGRU_cell-41 | (1, 4, 8, 8) | 0 |
| LeakyReLU-10 | (1, 4, 16, 16) | 0 | Dropout2d-42 | (1, 4, 8, 8) | 0 |
| Conv2d-11 | (1, 8, 16, 16) | 1,608 | GroupNorm-43 | (1, 4, 8, 8) | 8 |
| GroupNorm-12 | (1, 8, 16, 16) | 16 | Conv2d-44 | (1, 4, 8, 8) | 804 |
| Conv2d-13 | (1, 4, 16, 16) | 804 | GroupNorm-45 | (1, 8, 8, 8) | 16 |
| GroupNorm-14 | (1, 4, 16, 16) | 8 | ConvTrans2d-46 | (1, 8, 8, 8) | 1,608 |
| Dropout2d-15 | (1, 4, 16, 16) | 0 | LeakyReLU-47 | (1, 4, 8, 8) | 0 |
| CGRU_cell-16 | (1, 4, 16, 16) | 0 | Conv2d-48 | (1, 4, 8, 8) | 148 |
| Conv2d-17 | (1, 4, 8, 8) | 148 | CGRU_cell-49 | (1, 4, 16, 16) | 0 |
| LeakyReLU-18 | (1, 4, 8, 8) | 0 | Dropout2d-50 | (1, 4, 16, 16) | 0 |
| Conv2d-19 | (1, 8, 8, 8) | 1,608 | GroupNorm-51 | (1, 4, 16, 16) | 8 |
| GroupNorm-20 | (1, 8, 8, 8) | 16 | Conv2d-52 | (1, 4, 16, 16) | 804 |
| Conv2d-21 | (1, 4, 8, 8) | 804 | GroupNorm-53 | (1, 8, 16, 16) | 16 |
| GroupNorm-22 | (1, 4, 8, 8) | 8 | ConvTrans2d-54 | (1, 8, 16, 16) | 1,608 |
| Dropout2d-23 | (1, 4, 8, 8) | 0 | LeakyReLU-55 | (1, 4, 16, 16) | 0 |
| CGRU_cell-24 | (1, 4, 8, 8) | 0 | Conv2d-56 | (1, 4, 16, 16) | 148 |
| Conv2d-25 | (1, 4, 4, 4) | 148 | CGRU_cell-57 | (1, 4, 32, 32) | 0 |
| LeakyReLU-26 | (1, 4, 4, 4) | 0 | Dropout2d-58 | (1, 4, 32, 32) | 0 |
| Conv2d-27 | (1, 8, 4, 4) | 1,608 | GroupNorm-59 | (1, 4, 32, 32) | 8 |
| GroupNorm-28 | (1, 8, 4, 4) | 16 | Conv2d-60 | (1, 4, 32, 32) | 604 |
| Conv2d-29 | (1, 4, 4, 4) | 804 | GroupNorm-61 | (1, 8, 32, 32) | 16 |
| GroupNorm-30 | (1, 4, 4, 4) | 8 | ConvTrans2d-62 | (1, 8, 32, 32) | 1,208 |
| Dropout2d-31 | (1, 4, 4, 4) | 0 | LeakyReLU-63 | (1, 2, 32, 32) | 0 |
| CGRU_cell-32 | (1, 4, 4, 4) | 0 | Conv2d-64 | (1, 1, 32, 32) | 20 |

## C.2 PyTorch Model

```
AutoEncoder(
  (encoder): Encoder(
    (stage1): Sequential(
      (conv1_leaky_1): Conv2d(in=1, out=2, kernel_size=(3, 3),
                              stride=(1, 1),
                              padding=(1, 1))
      (leaky_conv1_leaky_1): LeakyReLU(negative_slope=0.2, inplace=True)
    )
    (rnn1): CGRU_cell(
      (dropout): Dropout2d(p=0.1, inplace=False)
      (conv_gates): Sequential(
        (0): Conv2d(in=6, out=8, kernel_size=(5, 5),
                    stride=(1, 1),
                    padding=(2, 2))
        (1): GroupNorm(groups=8, channels=8, eps=1e-05, affine=True)
      )
      (conv_can): Sequential(
        (0): Conv2d(in=6, out=4, kernel_size=(5, 5),
                    stride=(1, 1),
                    padding=(2, 2))
        (1): GroupNorm(groups=4, channels=4, eps=1e-05, affine=True)
      )
    )
    (stage2): Sequential(
      (conv2_leaky_1): Conv2d(in=4, out=4, kernel_size=(3, 3),
                              stride=(2, 2),
                              padding=(1, 1))
      (leaky_conv2_leaky_1): LeakyReLU(negative_slope=0.2, inplace=True)
    )
    (rnn2): CGRU_cell(
      (dropout): Dropout2d(p=0.1, inplace=False)
      (conv_gates): Sequential(
        (0): Conv2d(in=8, out=8, kernel_size=(5, 5),
                    stride=(1, 1),
                    padding=(2, 2))
        (1): GroupNorm(groups=8, channels=8, eps=1e-05, affine=True)
      )
      (conv_can): Sequential(
        (0): Conv2d(in=8, out=4, kernel_size=(5, 5),
                    stride=(1, 1),
                    padding=(2, 2))
        (1): GroupNorm(groups=4, channels=4, eps=1e-05, affine=True)
      )
    )
    (stage3): Sequential(
      (conv3_leaky_1): Conv2d(in=4, out=4, kernel_size=(3, 3),
                              stride=(2, 2),
                              padding=(1, 1))
      (leaky_conv3_leaky_1): LeakyReLU(negative_slope=0.2, inplace=True)
```

```
  )
  (rnn3): CGRU_cell(
    (dropout): Dropout2d(p=0.1, inplace=False)
    (conv_gates): Sequential(
      (0): Conv2d(in=8, out=8, kernel_size=(5, 5),
                  stride=(1, 1),
                  padding=(2, 2))
      (1): GroupNorm(groups=8, channels=8, eps=1e-05, affine=True)
    )
    (conv_can): Sequential(
      (0): Conv2d(in=8, out=4, kernel_size=(5, 5),
                  stride=(1, 1),
                  padding=(2, 2))
      (1): GroupNorm(groups=4, channels=4, eps=1e-05, affine=True)
    )
  )
  (stage4): Sequential(
    (conv4_leaky_1): Conv2d(in=4, out=4, kernel_size=(3, 3),
                            stride=(2, 2),
                            padding=(1, 1))
    (leaky_conv4_leaky_1): LeakyReLU(negative_slope=0.2, inplace=True)
  )
  (rnn4): CGRU_cell(
    (dropout): Dropout2d(p=0.1, inplace=False)
    (conv_gates): Sequential(
      (0): Conv2d(in=8, out=8, kernel_size=(5, 5),
                  stride=(1, 1),
                  padding=(2, 2))
      (1): GroupNorm(groups=8, channels=8, eps=1e-05, affine=True)
    )
    (conv_can): Sequential(
      (0): Conv2d(in=8, out=4, kernel_size=(5, 5),
                  stride=(1, 1),
                  padding=(2, 2))
      (1): GroupNorm(groups=4, channels=4, eps=1e-05, affine=True)
    )
  )
)
(decoder): Decoder(
  (rnn4): CGRU_cell(
    (dropout): Dropout2d(p=0.1, inplace=False)
    (conv_gates): Sequential(
      (0): Conv2d(in=8, out=8, kernel_size=(5, 5),
                  stride=(1, 1),
                  padding=(2, 2))
      (1): GroupNorm(groups=8, channels=8, eps=1e-05, affine=True)
    )
    (conv_can): Sequential(
      (0): Conv2d(in=8, out=4, kernel_size=(5, 5),
                  stride=(1, 1),
                  padding=(2, 2))
      (1): GroupNorm(groups=4, channels=4, eps=1e-05, affine=True)
```

```
    )
  )
  (stage4): Sequential(
    (deconv0_leaky_1): ConvTranspose2d(in=4, out=4, kernel_size=(4, 4),
                                       stride=(2, 2),
                                       padding=(1, 1))
    (leaky_deconv0_leaky_1): LeakyReLU(negative_slope=0.2, inplace=True)
  )
  (rnn3): CGRU_cell(
    (dropout): Dropout2d(p=0.1, inplace=False)
    (conv_gates): Sequential(
      (0): Conv2d(in=8, out=8, kernel_size=(5, 5),
                  stride=(1, 1),
                  padding=(2, 2))
      (1): GroupNorm(groups=8, channels=8, eps=1e-05, affine=True)
    )
    (conv_can): Sequential(
      (0): Conv2d(in=8, out=4, kernel_size=(5, 5),
                  stride=(1, 1),
                  padding=(2, 2))
      (1): GroupNorm(groups=4, channels=4, eps=1e-05, affine=True)
    )
  )
  (stage3): Sequential(
    (deconv1_leaky_1): ConvTranspose2d(in=4, out=4, kernel_size=(4, 4),
                                       stride=(2, 2),
                                       padding=(1, 1))
    (leaky_deconv1_leaky_1): LeakyReLU(negative_slope=0.2, inplace=True)
  )
  (rnn2): CGRU_cell(
    (dropout): Dropout2d(p=0.1, inplace=False)
    (conv_gates): Sequential(
      (0): Conv2d(in=8, out=8, kernel_size=(5, 5),
                  stride=(1, 1),
                  padding=(2, 2))
      (1): GroupNorm(groups=8, channels=8, eps=1e-05, affine=True)
    )
    (conv_can): Sequential(
      (0): Conv2d(in=8, out=4, kernel_size=(5, 5),
                  stride=(1, 1),
                  padding=(2, 2))
      (1): GroupNorm(groups=4, channels=4, eps=1e-05, affine=True)
    )
  )
  (stage2): Sequential(
    (deconv2_leaky_1): ConvTranspose2d(in=4, out=4, kernel_size=(4, 4),
                                       stride=(2, 2),
                                       padding=(1, 1))
    (leaky_deconv2_leaky_1): LeakyReLU(negative_slope=0.2, inplace=True)
  )
  (rnn1): CGRU_cell(
    (dropout): Dropout2d(p=0.1, inplace=False)
```

```
    (conv_gates): Sequential(
      (0): Conv2d(in=8, out=8, kernel_size=(5, 5),
                  stride=(1, 1),
                  padding=(2, 2))
      (1): GroupNorm(groups=8, channels=8, eps=1e-05, affine=True)
    )
    (conv_can): Sequential(
      (0): Conv2d(in=8, out=4, kernel_size=(5, 5),
                  stride=(1, 1),
                  padding=(2, 2))
      (1): GroupNorm(groups=4, channels=4, eps=1e-05, affine=True)
    )
  )
  (stage1): Sequential(
    (conv3_leaky_1): Conv2d(in=4, out=2, kernel_size=(3, 3),
                            stride=(1, 1),
                            padding=(1, 1))
    (leaky_conv3_leaky_1): LeakyReLU(negative_slope=0.2, inplace=True)
    (conv4_leaky_1): Conv2d(in=2, out=1, kernel_size=(1, 1),
                            stride=(1, 1),
                            padding=(0, 0))
    (leaky_conv4_leaky_1): LeakyReLU(negative_slope=0.2, inplace=True)
  )
 )
)
```

# Appendix D

## Glossary

**AD**    anomaly detection

**AE**    autoencoder

**AML**   average model loss

**ANN**   artificial neural network

**ATU**   average transport unit

**CNN**   convolutional neural network

**CPS**   cyber-physical system

**DCS**   distributed control system

**DPI**   deep packet inspection

**DR**    dimension reduction

**DoS**   Denial of Service

**F1**    $F_1$-score

**FN**    False Negatives

**FP**    False Positives

**FPR**   False Positive Rate

**GRU**   gated recurrent unit

**HIDS**   host-based IDS

**ICS**   industrial control system

**IF**    isolation forest

**IDS**   intrusion detection system

**IOC**   indicators of compromise

**IPS**   intrusion prevention system

**LOF**   local outlier factor

**LR**    learning rate

**LRP**   layer-wise relevance propagation

**LSTM**   Long short-term memory

**MLP**   multi-layer perceptron

**MSE**   mean square error

**MitM**   Man-in-the-Middle

**MTU**   maximum transport unit

**NIDS**   network-based IDS

**NLP**   natural language processing

**OCC**   one-class classification

**OCSVM**   one-class support vector machine

**OOS**   out-of-sample

**PCA**   principal component analysis

**PCAP**   packet capture

**PLC**   programmable logic controller

**PR**    Precision

**RC**    Recall

**RL**    representation learning

**RNN**   recurrent neural network

**ReLU**   rectified linear unit

**SVM**   support vector machine

**SWaT**   Secure Water Treatment

**TCP**   transmission control protocol

**TN**    True Negatives

**TP**    True Positives

**UDP**   user datagram protocol

**XAI**   explainable artificial intelligence

# Bibliography

[1]  *Netherlands Enterprise Agency. US and Washington DC Cyber Market Analysis 2021.* 2021.

[2]  R. I. Ogie. *Cyber Security Incidents on Critical Infrastructure and Industrial Networks.* In: *Proceedings of the 9th International Conference on Computer and Automation Engineering.* ICCAE '17. Sydney, Australia: Association for Computing Machinery, 2017, pp. 254–258.

[3]  L. Bilge and T. Dumitraş. *Before we knew it: an empirical study of zero-day attacks in the real world.* In: *Proceedings of the 2012 ACM conference on Computer and communications security.* 2012, pp. 833–844.

[4]  B. für Sicherheit in der Informationstechnik (BSI). *IT-Sicherheitsgesetz Broschüre.* bsi.bund.de/Publikationen/Broschueren/IT-Sicherheitsgesetz.html. 2019.

[5]  H. Hadeli, R. Schierholz et al. *Leveraging determinism in industrial control systems for advanced anomaly detection and reliable security configuration.* In: *2009 IEEE Conference on Emerging Technologies & Factory Automation.* IEEE. 2009, pp. 1–8.

[6]  M. Niedermaier, T. Hanka et al. *Efficient Passive ICS Device Discovery and Identification by MAC Address Correlation.* In: *CoRR* abs/1904.04271 (2019). arXiv: 1904.04271.

[7]  S. Ghosh and S. Sampalli. *A survey of security in SCADA networks: Current issues and future challenges.* In: *IEEE Access* 7 (2019), pp. 135812–135831.

[8]  K. E. Hemsley and D. R. E. Fisher. *History of Industrial Control System Cyber Incidents.* In: (12/2018).

[9]  I. Zografopoulos, J. Ospina et al. *Cyber-Physical Energy Systems Security: Threat Modeling, Risk Assessment, Resources, Metrics, and Case Studies.* 2021. arXiv: 2101.10198 [cs.CR].

[10]  Y. Luo, Y. Xiao et al. *Deep Learning-Based Anomaly Detection in Cyber-Physical Systems: Progress and Opportunities.* 2021. arXiv: 2003.13213 [cs.CR].

[11]  C. Parsons. *Deep Packet Inspection in Perspective: Tracing its lineage and surveillance potentials.* Citeseer, 2011.

[12]  B. Jabiyev, O. Mirzaei et al. *Preventing Server-Side Request Forgery Attacks.* In: *Proceedings of the 36th Annual ACM Symposium on Applied Computing.* SAC '21. Virtual Event, Republic of Korea: Association for Computing Machinery, 2021, pp. 1626–1635.

[13]  V.-T. Pham, M. Böhme and A. Roychoudhury. *AFLNET: A Greybox Fuzzer for Network Protocols.* In: *2020 IEEE 13th International Conference on Software Testing, Validation and Verification (ICST).* 2020, pp. 460–465.

[14]  mitre.org. *MITRE ATT&CK for Industrial Control Systems.* collaborate.mitre.org/attackics/index.php/Main_Page. 2021.

[15] L. Portnoy, E. Eskin and S. Stolfo. *Intrusion detection with unlabeled data using clustering*. In: *In Proceedings of ACM CSS Workshop on Data Mining Applied to Security (DMSA-2001*. 2001, pp. 5–8.

[16] D. E. Denning. *An Intrusion-Detection Model*. In: *Proceedings of the 1986 IEEE Symposium on Security and Privacy, Oakland, California, USA, April 7-9, 1986*. 1986, pp. 118–133.

[17] S. Axelsson. *The Base-Rate Fallacy and the Difficulty of Intrusion Detection*. In: *ACM Trans. Inf. Syst. Secur.* 3.3 (08/2000), pp. 186–205.

[18] D. Arp, E. Quiring et al. *Dos and Don'ts of Machine Learning in Computer Security*. 2020. arXiv: `2010.09470 [cs.CR]`.

[19] R. Sommer and V. Paxson. *Outside the closed world: On using machine learning for network intrusion detection*. In: *2010 IEEE symposium on security and privacy*. IEEE. 2010, pp. 305–316.

[20] E. Rahm and H. H. Do. *Data cleaning: Problems and current approaches*. In: *IEEE Data Eng. Bull.* 23.4 (2000), pp. 3–13.

[21] E. Tuv, A. Borisov et al. *Feature Selection with Ensembles, Artificial Variables, and Redundancy Elimination*. In: *J. Mach. Learn. Res.* 10 (12/2009), pp. 1341–1366.

[22] I. Guyon, S. Gunn et al. *Feature extraction: foundations and applications*. Vol. 207. Springer, 2008.

[23] Y. Bengio, A. Courville and P. Vincent. *Representation learning: A review and new perspectives*. In: *IEEE transactions on pattern analysis and machine intelligence* 35.8 (2013), pp. 1798–1828.

[24] L. Portnoy. *Intrusion detection with unlabeled data using clustering*. PhD thesis. Columbia University, 2000.

[25] T. Kohonen. *Correlation matrix memories*. In: *IEEE transactions on computers* 100.4 (1972), pp. 353–359.

[26] W. S. McCulloch and W. Pitts. *A logical calculus of the ideas immanent in nervous activity*. In: *The bulletin of mathematical biophysics* 5.4 (1943), pp. 115–133.

[27] V. Nair and G. E. Hinton. *Rectified linear units improve restricted boltzmann machines*. In: *ICML*. 2010.

[28] K. Hornik, M. Stinchcombe, H. White et al. *Multilayer feedforward networks are universal approximators*. In: *Neural networks* 2.5 (1989), pp. 359–366.

[29] H. T. Siegelmann and E. D. Sontag. *On the computational power of neural nets*. In: *Journal of computer and system sciences* 50.1 (1995), pp. 132–150.

[30] S. Linnainmaa. *The representation of the cumulative rounding error of an algorithm as a Taylor expansion of the local rounding errors*. In: *Master's Thesis (in Finnish), Univ. Helsinki* (1970), pp. 6–7.

[31] M. Andrychowicz, M. Denil et al. *Learning to learn by gradient descent by gradient descent.* In: *Advances in neural information processing systems.* 2016, pp. 3981–3989.

[32] S. Raschka. *Gradient Descent and Stochastic Gradient Descent.* `rasbt.github.io/mlxtend/`. 2019.

[33] B. Hanin. *Which Neural Net Architectures Give Rise to Exploding and Vanishing Gradients?* In: *Advances in Neural Information Processing Systems 31: Annual Conference on Neural Information Processing Systems 2018, NeurIPS 2018, December 3-8, 2018, Montréal, Canada.* Ed. by S. Bengio, H. M. Wallach et al. 2018, pp. 580–589.

[34] D. P. Kingma and J. Ba. *Adam: A Method for Stochastic Optimization.* 2017. arXiv: `1412.6980 [cs.LG]`.

[35] I. Loshchilov and F. Hutter. *Decoupled Weight Decay Regularization.* 2019. arXiv: `1711.05101 [cs.LG]`.

[36] D. O. Hebb. *The organization of behavior: A neuropsychological theory.* Psychology Press, 2005.

[37] L. N. Smith and N. Topin. *Super-Convergence: Very Fast Training of Neural Networks Using Large Learning Rates.* 2018. arXiv: `1708.07120 [cs.LG]`.

[38] N. Srivastava, G. Hinton et al. *Dropout: a simple way to prevent neural networks from overfitting.* In: *The journal of machine learning research* 15.1 (2014), pp. 1929–1958.

[39] vdumoulin. *A technical report on convolution arithmetic in the context of deep learning.* `github.com/vdumoulin/conv_arithmetic`. 2018.

[40] D. E. Rumelhart and J. L. McClelland. *An interactive activation model of context effects in letter perception: II. The contextual enhancement effect and some tests and extensions of the model.* In: *Psychological review* 89.1 (1982), p. 60.

[41] K. Cho, B. van Merrienboer et al. *Learning Phrase Representations using RNN Encoder-Decoder for Statistical Machine Translation.* In: *Proceedings of the 2014 Conference on Empirical Methods in Natural Language Processing, EMNLP 2014.* Ed. by A. Moschitti, B. Pang and W. Daelemans. ACL, 2014, pp. 1724–1734.

[42] M. Phi. *Illustrated Guide to LSTM's and GRU's: A step by step explanation.* 2018.

[43] D. H. Ballard. *Modular Learning in Neural Networks.* In: *AAAI.* 1987, pp. 279–284.

[44] W.-J. Jia and Y.-D. Zhang. *Survey on theories and methods of autoencoder.* In: *Computer Systems & Applications* 5 (2018).

[45] E. Plaut. *From Principal Subspaces to Principal Components with Linear Autoencoders.* In: *CoRR* abs/1804.10253 (2018). arXiv: `1804.10253`.

[46] D. P. Kingma and M. Welling. *Auto-Encoding Variational Bayes.* 2014. arXiv: `1312.6114 [stat.ML]`.

[47] S. S. Khan and M. G. Madden. *One-class classification: taxonomy of study and review of techniques.* In: *The Knowledge Engineering Review* 29.3 (2014), pp. 345–374.

[48] V. Chandola, A. Banerjee and V. Kumar. *Anomaly detection: A survey.* In: *ACM computing surveys (CSUR)* 41.3 (2009), pp. 1–58.

[49] M. A. Hearst, S. T. Dumais et al. *Support vector machines.* In: *IEEE Intelligent Systems and their Applications* 13.4 (1998), pp. 18–28.

[50] B. Schölkopf, R. C. Williamson et al. *Support vector method for novelty detection.* In: *Advances in neural information processing systems.* 2000, pp. 582–588.

[51] J. Mourão-Miranda, D. Hardoon et al. *Patient classification as an outlier detection problem: An application of the One-Class Support Vector Machine.* In: *NeuroImage* 58 (06/2011), pp. 793–804.

[52] F. T. Liu, K. M. Ting and Z.-H. Zhou. *Isolation forest.* In: *2008 Eighth IEEE International Conference on Data Mining.* IEEE. 2008, pp. 413–422.

[53] M. M. Breunig, H.-P. Kriegel et al. *LOF: Identifying Density-Based Local Outliers.* In: *SIGMOD Rec.* 29.2 (05/2000), pp. 93–104.

[54] W. Samek, G. Montavon et al. *Explaining Deep Neural Networks and Beyond: A Review of Methods and Applications.* In: *Proceedings of the IEEE* 109.3 (2021), pp. 247–278.

[55] L. H. Gilpin, D. Bau et al. *Explaining Explanations: An Overview of Interpretability of Machine Learning.* 2019. arXiv: 1806.00069 [cs.AI].

[56] A. Adadi and M. Berrada. *Peeking inside the black-box: a survey on explainable artificial intelligence (XAI).* In: *IEEE access* 6 (2018), pp. 52138–52160.

[57] G. Montavon, W. Samek and K.-R. Müller. *Methods for interpreting and understanding deep neural networks.* In: *Digital Signal Processing* 73 (02/2018), pp. 1–15.

[58] W. Samek, G. Montavon et al. *Explainable AI: interpreting, explaining and visualizing deep learning.* Vol. 11700. Springer Nature, 2019.

[59] S. Bach, A. Binder et al. *On pixel-wise explanations for non-linear classifier decisions by layer-wise relevance propagation.* In: *PloS one* 10.7 (2015), e0130140.

[60] D. Gunning. *Broad Agency Announcement Explainable Artificial Intelligence (XAI).* Tech. rep. Technical report, 2016.

[61] A. Binder, G. Montavon et al. *Layer-wise relevance propagation for neural networks with local renormalization layers.* In: *International Conference on Artificial Neural Networks.* Springer. 2016, pp. 63–71.

[62] G. Montavon, A. Binder et al. *Layer-wise relevance propagation: an overview.* In: *Explainable AI: interpreting, explaining and visualizing deep learning* (2019), pp. 193–209.

[63] A. Warnecke, D. Arp et al. *Evaluating Explanation Methods for Deep Learning in Security.* 2020. arXiv: 1906.02108 [cs.LG].

[64] L. Ruff, R. Vandermeulen et al. *Deep One-Class Classification.* In: *Proceedings of the 35th International Conference on Machine Learning.* Ed. by J. Dy and A. Krause.

Vol. 80. Proceedings of Machine Learning Research. PMLR, 10–15 Jul/2018, pp. 4393–4402.

[65] C. Olah, A. Mordvintsev and L. Schubert. *Feature Visualization*. In: *Distill* (2017). https://distill.pub/2017/feature-visualization.

[66] R. Boutaba, M. A. Salahuddin et al. *A comprehensive survey on machine learning for networking: evolution, applications and research opportunities*. In: *Journal of Internet Services and Applications* 9.1 (2018), pp. 1–99.

[67] A. Javaid, Q. Niyaz et al. *A Deep Learning Approach for Network Intrusion Detection System*. In: *SESA* 3.9 (05/2016).

[68] J. S. Bridle. *Training stochastic model recognition algorithms as networks can lead to maximum mutual information estimation of parameters*. In: *Advances in neural information processing systems*. 1990, pp. 211–217.

[69] W. Liu, Y. Wen et al. *Large-Margin Softmax Loss for Convolutional Neural Networks*. In: *Proceedings of the 33nd International Conference on Machine Learning, ICML 2016, New York City, NY, USA, June 19-24, 2016*. Ed. by M. Balcan and K. Q. Weinberger. Vol. 48. JMLR Workshop and Conference Proceedings. JMLR.org, 2016, pp. 507–516.

[70] M. Tavallaee, E. Bagheri et al. *A detailed analysis of the KDD CUP 99 data set*. In: *2009 IEEE symposium on computational intelligence for security and defense applications*. IEEE. 2009, pp. 1–6.

[71] N. Shone, T. N. Ngoc et al. *A deep learning approach to network intrusion detection*. In: *IEEE transactions on emerging topics in computational intelligence* 2.1 (2018), pp. 41–50.

[72] L. Breiman. *Random forests*. In: *Machine learning* 45.1 (2001), pp. 5–32.

[73] R. Vinayakumar, K. Soman and P. Poornachandran. *Applying convolutional neural network for network intrusion detection*. In: *2017 International Conference on Advances in Computing, Communications and Informatics (ICACCI)*. IEEE. 2017, pp. 1222–1228.

[74] M. M. U. Chowdhury, F. Hammond et al. *A few-shot deep learning approach for improved intrusion detection*. In: *2017 IEEE 8th Annual Ubiquitous Computing, Electronics and Mobile Communication Conference (UEMCON)*. IEEE. 2017, pp. 456–462.

[75] S. Naseer, Y. Saleem et al. *Enhanced network anomaly detection based on deep neural networks*. In: *IEEE Access* 6 (2018), pp. 48231–48246.

[76] R. Vinayakumar, M. Alazab et al. *Deep learning approach for intelligent intrusion detection system*. In: *IEEE Access* 7 (2019), pp. 41525–41550.

[77] B. J. Radford, L. M. Apolonio et al. *Network traffic anomaly detection using recurrent neural networks*. In: *arXiv preprint arXiv:1803.10769* (2018).

[78] K. Babaei, Z. Chen and T. Maul. *AEGR: A simple approach to gradient reversal in autoencoders for network anomaly detection.* In: *CoRR* abs/1912.13387 (2019). arXiv: [1912.13387](1912.13387).

[79] Y. Mirsky, T. Doitshman et al. *Kitsune: an ensemble of autoencoders for online network intrusion detection.* In: *Appears in Network and Distributed Systems Security Symposium (NDSS) 2018* (2018).

[80] D. Hadžiosmanović, L. Simionato et al. *N-gram against the machine: On the feasibility of the n-gram network analysis for binary protocols.* In: *International Workshop on Recent Advances in Intrusion Detection.* Springer. 2012, pp. 354–373.

[81] P. Düssel, C. Gehl et al. *Cyber-Critical Infrastructure Protection Using Real-Time Payload-Based Anomaly Detection.* In: vol. 6027. 09/2009, pp. 85–97.

[82] C. Wressnegger, A. Kellner and K. Rieck. *Zoe: Content-based anomaly detection for industrial control systems.* In: *2018 48th Annual IEEE/IFIP International Conference on Dependable Systems and Networks (DSN).* IEEE. 2018, pp. 127–138.

[83] M. Caselli, E. Zambon and F. Kargl. *Sequence-aware intrusion detection in industrial control systems.* In: *Proceedings of the 1st ACM Workshop on Cyber-Physical System Security.* 2015, pp. 13–24.

[84] S. D. D. Anton, A. Hafner and H. D. Schotten. *Devil in the detail: Attack scenarios in industrial applications.* In: *2019 IEEE Security and Privacy Workshops (SPW).* IEEE. 2019, pp. 169–174.

[85] S. Kim, W. Jo and T. Shon. *APAD: Autoencoder-based Payload Anomaly Detection for industrial IoE.* In: *Applied Soft Computing* 88 (2020), p. 106017.

[86] A. P. Mathur and N. O. Tippenhauer. *SWaT: a water treatment testbed for research and training on ICS security.* In: *2016 International Workshop on Cyber-physical Systems for Smart Water Networks (CySWater).* 2016, pp. 31–36.

[87] S. D. D. Anton, D. Fraunholz and H. D. Schotten. *Using temporal and topological features for intrusion detection in operational networks.* In: *Proceedings of the 14th International Conference on Availability, Reliability and Security.* 2019, pp. 1–9.

[88] J. Inoue, Y. Yamagata et al. *Anomaly detection for a water treatment system using unsupervised machine learning.* In: *2017 IEEE international conference on data mining workshops (ICDMW).* IEEE. 2017, pp. 1058–1065.

[89] P. Schneider and K. Bottinger. *High-performance unsupervised anomaly detection for cyber-physical system networks.* In: *Proceedings of the 2018 Workshop on Cyber-Physical Systems Security and PrivaCy.* 2018, pp. 1–12.

[90] H. Yang, L. Cheng and M. C. Chuah. *Deep-learning-based network intrusion detection for SCADA systems.* In: *2019 IEEE Conference on Communications and Network Security (CNS).* IEEE. 2019, pp. 1–7.

[91] J. Gao, L. Gan et al. *Omni SCADA Intrusion Detection Using Deep Learning Algorithms.* In: *IEEE Internet of Things Journal* (2020), pp. 1–1.

[92] G. Marin, P. Casas and G. Capdehourat. *Rawpower: Deep learning based anomaly detection from raw network traffic measurements*. In: *Proceedings of the ACM SIGCOMM 2018 Conference on Posters and Demos*. 2018, pp. 75–77.

[93] G. Marin, P. Casas and G. Capdehourat. *Deep in the Dark-Deep Learning-Based Malware Traffic Detection Without Expert Knowledge*. In: *2019 IEEE Security and Privacy Workshops (SPW)*. IEEE. 2019, pp. 36–42.

[94] P. Casas, G. Marin et al. *MLSEC-Benchmarking Shallow and Deep Machine Learning Models for Network Security*. In: *2019 IEEE Security and Privacy Workshops (SPW)*. IEEE. 2019, pp. 230–235.

[95] G. Bernieri, M. Conti and F. Turrin. *KingFisher: An Industrial Security Framework Based on Variational Autoencoders*. In: *Proceedings of the 1st Workshop on Machine Learning on Edge in Sensor Systems*. SenSys-ML 2019. New York, NY, USA: Association for Computing Machinery, 2019, pp. 7–12.

[96] L. Zhao, L. Cai et al. *A Novel Network Traffic Classification Approach via Discriminative Feature Learning*. In: *Proceedings of the 35th Annual ACM Symposium on Applied Computing*. SAC '20. Brno, Czech Republic: Association for Computing Machinery, 2020, pp. 1026–1033.

[97] W. Zhang, J. Wang et al. *A Framework for Resource-Aware Online Traffic Classification Using CNN*. In: *Proceedings of the 14th International Conference on Future Internet Technologies*. CFI'19. Phuket, Thailand: Association for Computing Machinery, 2019.

[98] W. Wang, M. Zhu et al. *Malware traffic classification using convolutional neural network for representation learning*. In: *2017 International Conference on Information Networking (ICOIN)*. 2017, pp. 712–717.

[99] Y. Yu, J. Long and Z. Cai. *Session-based network intrusion detection using a deep learning architecture*. In: *International Conference on Modeling Decisions for Artificial Intelligence*. Springer. 2017, pp. 144–155.

[100] S. García, M. Grill et al. *An empirical comparison of botnet detection methods*. In: *Computers & Security* 45 (2014), pp. 100–123.

[101] M. Lotfollahi, M. J. Siavoshani et al. *Deep packet: A novel approach for encrypted traffic classification using deep learning*. In: *Soft Computing* 24.3 (2020), pp. 1999–2012.

[102] W. Wang, M. Zhu et al. *End-to-end encrypted traffic classification with one-dimensional convolution neural networks*. In: *2017 IEEE International Conference on Intelligence and Security Informatics (ISI)*. IEEE. 2017, pp. 43–48.

[103] G. Draper-Gil., A. H. Lashkari. et al. *Characterization of Encrypted and VPN Traffic using Time-related Features*. In: *Proceedings of the 2nd International Conference on Information Systems Security and Privacy - ICISSP,* INSTICC. SciTePress, 2016, pp. 407–414.

[104] W. Wang, M. Zhu et al. *Malware traffic classification using convolutional neural network for representation learning*. In: *2017 International Conference on Information Networking (ICOIN)*. IEEE. 2017, pp. 712–717.

[105] Y. LeCun, L. Bottou et al. *Gradient-based learning applied to document recognition.* In: *Proceedings of the IEEE* 86.11 (1998), pp. 2278–2324.

[106] H.-K. Lim, J.-B. Kim et al. *Packet-based network traffic classification using deep learning.* In: *2019 International Conference on Artificial Intelligence in Information and Communication (ICAIIC).* IEEE. 2019, pp. 046–051.

[107] K. He, X. Zhang et al. *Deep residual learning for image recognition.* In: *Proceedings of the IEEE conference on computer vision and pattern recognition.* 2016, pp. 770–778.

[108] P. Sperl, J. Schulze and K. Böttinger. *Activation Anomaly Analysis.* In: *Machine Learning and Knowledge Discovery in Databases - European Conference, ECML PKDD 2020, Ghent, Belgium, September 14-18, 2020, Proceedings, Part II.* Ed. by F. Hutter, K. Kersting et al. Vol. 12458. Lecture Notes in Computer Science. Springer, 2020, pp. 69–84.

[109] A. Shiravi, H. Shiravi et al. *Toward developing a systematic approach to generate benchmark datasets for intrusion detection.* In: *Computers & Security* 31.3 (2012), pp. 357–374.

[110] W. Wang, Y. Sheng et al. *HAST-IDS: Learning hierarchical spatial-temporal features using deep neural networks to improve intrusion detection.* In: *IEEE Access* 6 (2017), pp. 1792–1806.

[111] Z.-Q. Qin, X.-K. Ma and Y.-J. Wang. *Attentional payload anomaly detector for web applications.* In: *International Conference on Neural Information Processing.* Springer. 2018, pp. 588–599.

[112] A. Vaswani, N. Shazeer et al. *Attention is all you need.* In: *Advances in neural information processing systems.* 2017, pp. 5998–6008.

[113] I. Sharafaldin., A. Habibi Lashkari. and A. A. Ghorbani. *Toward Generating a New Intrusion Detection Dataset and Intrusion Traffic Characterization.* In: *ICISSP,* IN-STICC. SciTePress, 2018, pp. 108–116.

[114] H. D. C. 2010. *CSIC-dataset.* In:

[115] E. L. Goodman, C. Zimmerman and C. Hudson. *Packet2Vec: Utilizing Word2Vec for Feature Extraction in Packet Data.* In: (29/04/2020). arXiv: `2004.14477v1 [cs.LG]`.

[116] T. Mikolov, K. Chen et al. *Efficient Estimation of Word Representations in Vector Space.* In: (16/01/2013). arXiv: `1301.3781v3 [cs.CL]`.

[117] J. Liu, X. Song et al. *Deep Anomaly Detection in Packet Payload.* In: *CoRR* abs/1912.02549 (2019). arXiv: `1912.02549`.

[118] R.-H. Hwang, M.-C. Peng et al. *An LSTM-Based Deep Learning Approach for Classifying Malicious Traffic at the Packet Level.* In: *Applied Sciences* 9.16 (2019).

[119] R. Hwang, M. Peng et al. *An Unsupervised Deep Learning Model for Early Network Traffic Anomaly Detection.* In: *IEEE Access* 8 (2020), pp. 30387–30399.

[120] Z. Du, L. Ma et al. *Network Traffic Anomaly Detection Based on Wavelet Analysis*. In: *16th IEEE International Conference on Software Engineering Research, Management and Applications, SERA 2018, Kunming, China, June 13-15, 2018*. Ed. by S. Yao, Z. Jin et al. IEEE Computer Society, 2018, pp. 94–101.

[121] Z. Wang. *The applications of deep learning on traffic identification*. In: *BlackHat USA* 24.11 (2015), pp. 1–10.

[122] M. Zheng, H. Robbins et al. *Cybersecurity research datasets: taxonomy and empirical analysis*. In: *11th USENIX Workshop on Cyber Security Experimentation and Test CSET 18)*. 2018.

[123] A. P. Mathur and N. O. Tippenhauer. *SWaT: a water treatment testbed for research and training on ICS security*. In: *2016 International Workshop on Cyber-physical Systems for Smart Water Networks (CySWater)*. 2016, pp. 31–36.

[124] M. Tavallaee, N. Stakhanova and A. A. Ghorbani. *Toward Credible Evaluation of Anomaly-Based Intrusion-Detection Methods*. In: *IEEE Transactions on Systems, Man, and Cybernetics, Part C (Applications and Reviews)* 40.5 (2010), pp. 516–524.

[125] H. Ringberg, M. Roughan and J. Rexford. *The Need for Simulation in Evaluating Anomaly Detectors*. In: *SIGCOMM Comput. Commun. Rev.* 38.1 (01/2008), pp. 55–59.

[126] S. Abt and H. Baier. *A plea for utilising synthetic data when performing machine learning based cyber-security experiments*. In: *Proceedings of the 2014 Workshop on Artificial Intelligent and Security Workshop*. 2014, pp. 37–45.

[127] P. Schneider and A. Giehl. *Realistic Data Generation for Anomaly Detection in Industrial Settings Using Simulations*. In: *Computer Security*. Ed. by S. K. Katsikas, F. Cuppens et al. Cham: Springer International Publishing, 2019, pp. 119–134.

[128] R. Hannusch. *Synthetische Generierung von netzbasierten Angriffsdaten*. In: *Master's Thesis (in German), BTU Cottbus* (2018).

[129] M. Conti, D. Donadel and F. Turrin. *A Survey on Industrial Control System Testbeds and Datasets for Security Research*. In: *IEEE Communications Surveys & Tutorials* (2021), pp. 1–1.

[130] P. Cunningham. *Dimension reduction*. In: *Machine learning techniques for multimedia*. Springer, 2008, pp. 91–112.

[131] S. Jegelka and A. Gretton. *Brisk Kernel ICA*. In: *Large Scale Kernel Machines*. Neural Information Processing. Cambridge, MA, USA: MIT Press, 09/2007, pp. 225–250.

[132] S. T. Roweis and L. K. Saul. *Nonlinear dimensionality reduction by locally linear embedding*. In: *science* 290.5500 (2000), pp. 2323–2326.

[133] P. Orponen et al. *Computational complexity of neural networks: a survey*. In: *Nordic Journal of Computing* (1994).

[134] J. Saxe, R. Harang et al. *A deep learning approach to fast, format-agnostic detection of malicious web content*. In: *2018 IEEE Security and Privacy Workshops (SPW)*. IEEE. 2018, pp. 8–14.

[135]  X. Shi, Z. Chen et al. *Convolutional LSTM Network: A Machine Learning Approach for Precipitation Nowcasting.* In: vol. 28. 2015, pp. 802–810.

[136]  S. Hochreiter and J. Schmidhuber. *Long short-term memory.* In: *Neural computation* 9.8 (1997), pp. 1735–1780.

[137]  Y. Wang, M. Long et al. *Predrnn: Recurrent neural networks for predictive learning using spatiotemporal lstms.* In: *Proceedings of the 31st International Conference on Neural Information Processing Systems.* 2017, pp. 879–888.

[138]  D. Song. *DPKT.* https://github.com/kbandla/dpkt. 2006.

[139]  A. Paszke, S. Gross et al. *PyTorch: An Imperative Style, High-Performance Deep Learning Library.* In: *Advances in Neural Information Processing Systems 32.* Curran Associates, Inc., 2019, pp. 8024–8035.

[140]  F. Pedregosa, G. Varoquaux et al. *Scikit-learn: Machine Learning in Python.* In: *Journal of Machine Learning Research* 12 (2011), pp. 2825–2830.

[141]  B. Xu, A. Shirani et al. *Prediction of Relatedness in Stack Overflow: Deep Learning vs. SVM: A Reproducibility Study.* In: *Proceedings of the 12th ACM/IEEE International Symposium on Empirical Software Engineering and Measurement.* ESEM '18. Oulu, Finland: Association for Computing Machinery, 2018.

[142]  A. Collette. *Python and HDF5.* O'Reilly, 2013.

[143]  D. Erhan, A. Courville et al. *Why Does Unsupervised Pre-training Help Deep Learning?* In: *Proceedings of the Thirteenth International Conference on Artificial Intelligence and Statistics.* Vol. 9. Proceedings of Machine Learning Research. PMLR, 13–15 May/2010, pp. 201–208.

[144]  I. Sutskever, J. Martens et al. *On the importance of initialization and momentum in deep learning.* In: *Proceedings of the 30th International Conference on Machine Learning, ICML 2013, Atlanta, GA, USA, 16-21 June 2013.* Vol. 28. JMLR Workshop and Conference Proceedings. JMLR.org, 2013, pp. 1139–1147.

[145]  K. Mukherjee, A. Khare and A. Verma. *A Simple Dynamic Learning Rate Tuning Algorithm For Automated Training of DNNs.* In: *CoRR* abs/1910.11605 (2019). arXiv: 1910.11605.

[146]  A. Senior, G. Heigold et al. *An Empirical study of learning rates in deep neural networks for speech recognition.* In: *IEEE ICASSP.* Vancouver, CA, 2013.

[147]  H. P. Gavin. *The Levenberg-Marquardt algorithm for nonlinear least squares curve-fitting problems.* In: *Department of Civil and Environmental Engineering, Duke University* (2019), pp. 1–19.

[148]  Y. Zhou, X. Song et al. *Feature Encoding with AutoEncoders for Weakly-supervised Anomaly Detection.* In: *CoRR* abs/2105.10500 (2021). arXiv: 2105.10500.

[149]  L. Ruff, R. A. Vandermeulen et al. *Deep Semi-Supervised Anomaly Detection.* In: *CoRR* abs/1906.02694 (2019). arXiv: 1906.02694.